\documentclass{ieeetmlcn}
\usepackage{cite}
\usepackage{amsmath,amssymb,amsfonts}
\usepackage{graphicx,color}
\usepackage{textcomp}
\usepackage{xcolor}
\usepackage{hyperref}
\usepackage{algorithm}
\usepackage{algpseudocode}

\usepackage{graphics,hhline,bbm,array,psfrag,subfigure,stfloats}
\usepackage{acronym,tikz,xcolor}
\usepackage{epsfig}
\usepackage{soul}
\usepackage{amsthm}
\usepackage{subcaption}
\usepackage{booktabs}
\usepackage[markup=underlined]{changes}
\usepackage{multirow}

\usetikzlibrary{shapes,arrows}
\usepackage{verbatim}
\usepackage{winsnotation}

\def\BibTeX{{\rm B\kern-.05em{\sc i\kern-.025em b}\kern-.08em
    T\kern-.1667em\lower.7ex\hbox{E}\kern-.125emX}}
\AtBeginDocument{\definecolor{tmlcncolor}{cmyk}{0.93,0.59,0.15,0.02}\definecolor{NavyBlue}{RGB}{0,86,125}}

\newcommand{\cyan}[1] {{\color{black}{#1}}}
\newcommand{\blue}[1] {{\color{black}{#1}}}

\newcommand{\bluee}[1] {{\color{black}{#1}}}

\newcommand{\R}{\mathbb{R}}

\renewcommand{\IEEEQED}{\IEEEQEDopen}

\newtheorem{definition}{Definition}

\newtheorem{remark}{Remark}

\newtheorem{proposition}{Proposition}

\acrodef{PDF}{probability density function}
\acrodef{PMF}{probability mass function}
\acrodef{PP}{point process}
\acrodef{RV}{random variable}
\acrodef{IID}{independent identically distributed}
\acrodef{LTI}{linear time invariant}
\acrodef{MCvD}{molecular communication via diffusion}
\acrodef{MISO}{multiple input single output}
\acrodef{FA}{fully absorbing}
\acrodef{PS}{passive}
\acrodef{SNR}{signal-to-noise ratio}
\acrodef{SIR}{signal-to-interference ratio}
\acrodef{SINR}{signal-to-interference-noise ratio}
\acrodef{PPP}{Poisson point process}
\acrodef{ISI}{inter-symbol interference}
\acrodef{BEP}{bit error probability}
\acrodef{BER}{bit error rate}
\acrodef{AWGN}{additive white Gaussian noise}
\acrodef{OOK}{on-off keying}
\acrodef{FIM}{Fisher information matrix}
\acrodef{CRB}{Cram\'er-Rao bound}
\acrodef{JCS}{Joint communication and sensing}
\acrodef{ML}{maximum likelihood}
\acrodef{SI}{shift invariant}
\acrodef{WKS}{Whittaker-Kotelnikov-Shannon}
\acrodef{MSE}{mean square error}
\acrodef{PSD}{power spectral density}
\acrodef{TDA}{topological data analysis}
\acrodef{GENEO}{Group Equivariant Non-Expansive Operator}
\acrodef{RSRP}{reference signal received power}
\acrodef{RSS}{received signal strength}
\acrodef{WSN}{wireless sensor networks}
\acrodef{GP}{Gaussian process}
\acrodef{6G}{sixth generation}
\acrodef{ML}{machine learning}
\acrodef{CNN}{convolutional neural network}
\acrodef{CVAE}{conditional variational autoencoder}

\def\OJlogo{\vspace{-4pt}\includegraphics[height=18pt]{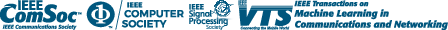}}
\def\seclogo{\vspace{10pt}\includegraphics[height=18pt]{new_logo}}

\def\authorrefmark#1{\ensuremath{^{\textbf{#1}}}}

\begin{document}


\title{Reconstruction of SINR Maps from Sparse Measurements using Group Equivariant Non-Expansive Operators}

\author{~Lorenzo~Mario~Amorosa\authorrefmark{1}, Member, IEEE, ~Francesco~Conti\authorrefmark{2}, ~Nicola~Quercioli\authorrefmark{1}, ~Flavio~Zabini\authorrefmark{1}, Member, IEEE, ~Tayebeh~Lotfi~Mahyari\authorrefmark{3}, ~Yiqun~Ge\authorrefmark{3}, and ~Patrizio~Frosini\authorrefmark{4}}
\affil{Department of Electrical, Electronic and Information Engineering (DEI), ``Guglielmo Marconi", University of Bologna \& WiLab - National Wireless Communication Laboratory (CNIT), Bologna, Italy.}
\affil{Institut national de recherche en sciences et technologies du numérique (Inria), Université Côte d’Azur, France.}
\affil{Huawei Technologies Canada, Ottawa, ON, Canada.}
\affil{Department of Computer Science, University of Pisa, Pisa, Italy.}
\corresp{Corresponding author: Flavio~Zabini (email: flavio.zabini2@unibo.it).}
\authornote{This work has been carried out in the framework of the CNIT National Laboratory WiLab and the WiLab-Huawei Joint Innovation Center.}

\begin{abstract}
As \ac{6G} wireless networks evolve, accurate \ac{SINR} maps are becoming increasingly critical for effective resource management and optimization. However, acquiring such maps at high resolution is often cost-prohibitive, creating a severe data scarcity challenge. This necessitates \ac{ML} approaches capable of robustly reconstructing the full map from extremely sparse measurements. To address this, we introduce a novel reconstruction framework based on Group Equivariant Non-Expansive Operators (GENEOs). Unlike data-hungry \ac{ML} models, GENEOs are low-complexity operators that embed domain-specific geometric priors, such as translation \cyan{and rotational equivariance}, directly into their structure. This provides a strong inductive bias, enabling effective reconstruction from very few samples. Our key insight is that for network management, preserving the topological structure of the \ac{SINR} map, such as the geometry of coverage holes and interference patterns, is often more critical than minimizing pixel-wise error. We validate our approach on realistic ray-tracing-based urban scenarios, evaluating performance with both traditional statistical metrics (mean squared error (MSE)) and, crucially, a topological metric (1-Wasserstein distance). \blue{Results show that our method consistently achieves superior statistical and topological accuracy across diverse urban scenarios. Compared to the best-performing baselines, GENEO reduces MSE up to 45\% and decreases the 1-Wasserstein distance up to 54\%. Crucially, these substantial performance gains are maintained even under the most extreme tested conditions, such as a 1\% sampling rate with a 30\% measurement error, and when measurements are highly spatially biased.} This demonstrates the practical advantage of GENEOs for creating structurally accurate \ac{SINR} maps that are more reliable for downstream network optimization tasks.
\end{abstract}


\maketitle

\section{INTRODUCTION}
\IEEEPARstart{I}{n current} and future communication systems such as \acf{6G} wireless networks, key functionalities like resource management and service provisioning rely heavily on the knowledge of spatially-varying quantities \cite{wang2023}, including \acf{SINR} maps, which indicate the signal quality experienced by users in a wireless network. 
However, acquiring large volumes of such data is often prohibitively expensive in terms of resource consumption, including power, spectrum, and time \cite{dataScarcity}. Therefore, it becomes crucial to reconstruct these quantities from a significantly smaller set of measurements.
\blue{In this work, we consider \ac{SINR} map reconstruction problems under extremely sparse sampling conditions. Given a geographic area represented as a 2D spatial grid where only a minor fraction of points contain measured \ac{SINR} values and the vast majority are unknown, the objective is to accurately predict the \ac{SINR} values for all missing points to recover the complete 2D target signal. 
}
\blue{In urban environments, radio propagation is heavily influenced by the complex geometry of buildings and streets, leading to highly complex \ac{SINR} distributions. Reconstructing these maps from extremely sparse measurements (e.g., sampling rates as low as $1\%$) and without large training datasets poses a difficult problem. Under such constrained conditions, both traditional spatial interpolation techniques (e.g., Kriging \cite{kriging2018}) and established \acf{ML} models (e.g., \acp{CNN}-based models \cite{radiounet}) face fundamental limitations.}

\blue{While \ac{ML} is increasingly leveraged in next-generation wireless networks for tasks such as radio resource scheduling, network traffic forecasting, or beamforming \cite{use_cases_6G}, the heavy parameterization of traditional neural networks makes them ill-suited for such data-scarce scenarios.} 
\blue{Overcoming these limitations requires a paradigm shift towards lightweight approaches capable of bypassing intensive training phases by naturally incorporating prior geometric constraints of the urban environment.}
\blue{To address this, we propose a completely novel approach for signal reconstruction based on the framework of \acp{GENEO}} \cite{Bocchi2025, bergomi2019towards,BoFeFr25,BocchiFMPPGGLBF24,colombini2025mathematical} \cyan{and apply it to the reconstruction of \ac{SINR} maps for wireless systems in urban environments.}
\acp{GENEO} were originally developed within topological data analysis (TDA) and are central to the theory of group equivariant operators (GEOs) \cite{bergomi2019towards}.
GEOs are of particular interest because they provide a mathematical formalization of the concept of an observer as a functional agent and, in particular, enable a topological-geometric representation of neural networks, \blue{while not requiring computational overhead due to training}.
Within the GEO framework, GENEOs play a crucial role, as they allow for the definition of a distance between GEOs without requiring the operators to share the same domains and codomains \cite{colombini2025mathematical}. Moreover, they exhibit a useful approximation property via finite sets of operators.

Our main contributions are summarized as follows:
\begin{enumerate}
    \item \blue{We address the problem of \ac{SINR} map reconstruction from extremely sparse samples, proposing a novel GENEO-based framework.}
    \item \blue{We integrate \ac{TDA} metrics, specifically the 1-Wasserstein distance, alongside standard statistical metrics (\ac{MSE}) to comprehensively evaluate both geometric fidelity and pixel-wise accuracy.}
    \item We provide a comprehensive empirical evaluation using realistic wireless scenarios generated with the Sionna RT ray tracing tool \cite{sionna-rt}. \blue{Our results demonstrate that GENEOs consistently achieve superior topological and statistical metrics than state-of-the-art statistical and \ac{ML} baselines across all considered scenarios under extreme data sparsity, while varying data sampling schemes, and measurement noise.}
\end{enumerate}
%
\blue{The remainder of the paper is organized as follows. In Sec.~\ref{sec:sota}, we introduce theoretical concepts about TDA and mathematical foundations of GENEOs.
Sec.~\ref{sec:related_work} reports related works. 
In Sec.~\ref{sec:geneo_rec}, we develop our proposed GENEO for signal reconstruction. In  Sec.~\ref{sec:numerical}, we numerically evaluate our approach in urban scenarios leveraging ray-tracing simulations in Sionna RT and comparing with state-of-the-art statistical and \ac{ML} baselines. Finally, Sec.~\ref{sec:conclusion} provides the conclusions of the paper.}

\section{BACKGROUND}
\label{sec:sota}

\subsection{TOPOLOGICAL DATA ANALYSIS}

TDA \cite{carlsson2009topology} is a branch of applied mathematics that studies the shape of digital data and leverages this information across various fields of artificial intelligence. Unlike traditional statistical methods, TDA applies concepts from topology to analyze data, enabling the investigation of global structures and properties that remain invariant under continuous deformations.

In practice, TDA analyzes connected components, holes, and higher-dimensional cavities. The mathematical foundation of TDA is persistent homology \cite{edelsbrunner2008persistent}, with its precursor being the size function \cite{frosini1992measuring}. Persistent homology not only identifies topological features in the data\blue{, such as the connected components and cavities mentioned above,} but also quantifies their persistence across multiple scales. A common assumption is that features with longer persistence are more relevant for characterizing the underlying shape of the data.
TDA has achieved remarkable success in real-world applications, spanning medicine  \cite{nicolau2011topology, carlsson2021topological, petri2014homological}, materials science \cite{lee2017quantifying, hiraoka2016hierarchical}, finance \cite{gidea2018topological, gidea2020topological}, computer vision \cite{carlsson2008local}, and more. Recently, TDA has also been applied to wireless networks to study coverage properties \cite{de2007coverage, ramazani2016topological}.

Since TDA focuses on the shape of data, it is well-suited for analyzing wireless networks, which often exhibit geometric patterns in signal propagation. For this reason, a metric from the TDA framework\blue{, i.e., the 1-Wasserstein distance (1-W) defined in Definition \ref{def:p-w},} is employed alongside the \ac{MSE}, to evaluate how well the original signal is preserved during reconstruction. While MSE captures differences in signal intensity, the 1-Wasserstein distance assesses the preservation of the signal's shape.

The concept of GENEO was originally introduced to reduce the invariance of topological descriptors in the context of classical TDA. Such invariance, in fact, holds for any possible 
homeomorphism applied to the data domain and results in a notion of shape equivalence that is too generic for many applications. The use of GENEOs makes it possible to restrict this invariance to the action of transformation groups chosen arbitrarily \cite{FrJa16}. Subsequently, the theory of GENEOs has been applied in the field of \ac{ML} as a geometric-topological technique for obtaining an interpretable approximation of functional operators and neural networks \cite{bergomi2019towards,BocchiFMPPGGLBF24,BoFeFr25,colombini2025mathematical}.

In the remainder of this section, we introduce the notion of \textit{persistence diagram} 
to capture the concept of shape, briefly recalling fundamental definitions and concepts of TDA. We refer the interested reader to \cite{carlsson2009topology} and \cite{edelsbrunner2008persistent} for further details.
Formally, given a topological space or simplicial complex $X$, a \textit{filtration} is a finite nested sequence of subspaces or subcomplexes $\emptyset = X_0 \subseteq X_1 \subseteq \dots \subseteq X_n = X$. For each $i=0, \dots, n$ we can compute the homology $H(X_i)$ and keep track of how the respecting features persist during the filtration process. In particular, we say that a topological feature \textit{borns} at time $i$ if it appears in $H(X_i)$ but not in $H(X_{i-1})$. Similarly, a topological feature \textit{dies} at time $j$ if it appears in $H(X_j)$ but not in $H(X_{j+1})$. Informally, we can track the evolution of components and holes in the sublevels of the filtration, keeping record of their creation and disappearance. We collect the pairs (birth time, death time) of components and holes in the so-called persistence diagram (PD). Mathematically, a PD is a multiset, i.e. a set where each point has a multiplicity. In literature, all points of the diagonal are added to the PD with infinite multiplicity for technical reasons. 

\begin{figure}[t]
\setkeys{Gin}{width=1\linewidth}
\centering
\begin{minipage}[t]{0.205\textwidth}
\label{fig:1a}
\includegraphics{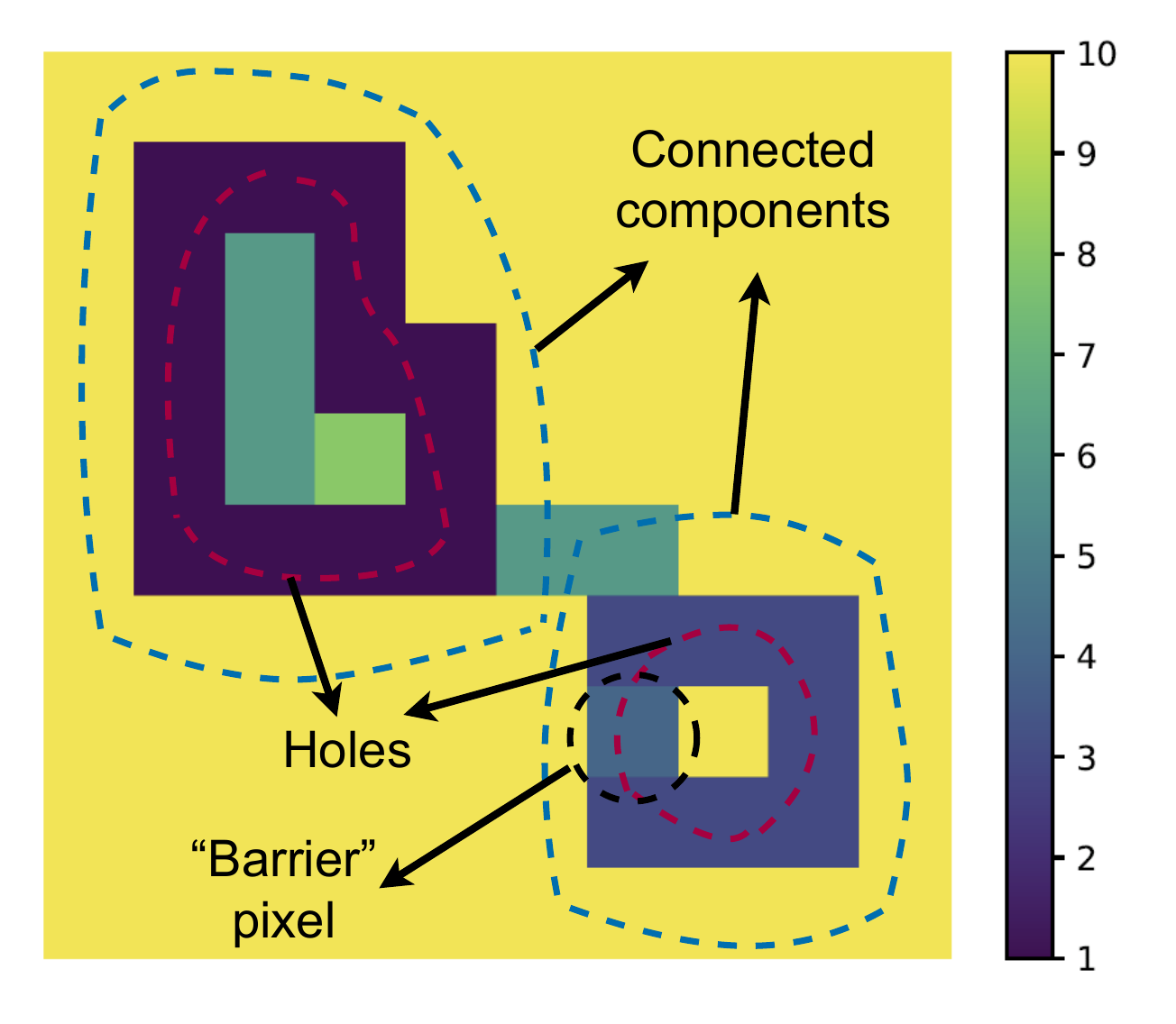}
\centering{(a)}
\end{minipage}
\begin{minipage}[t]{0.275\textwidth}
\label{fig:1b}
\includegraphics{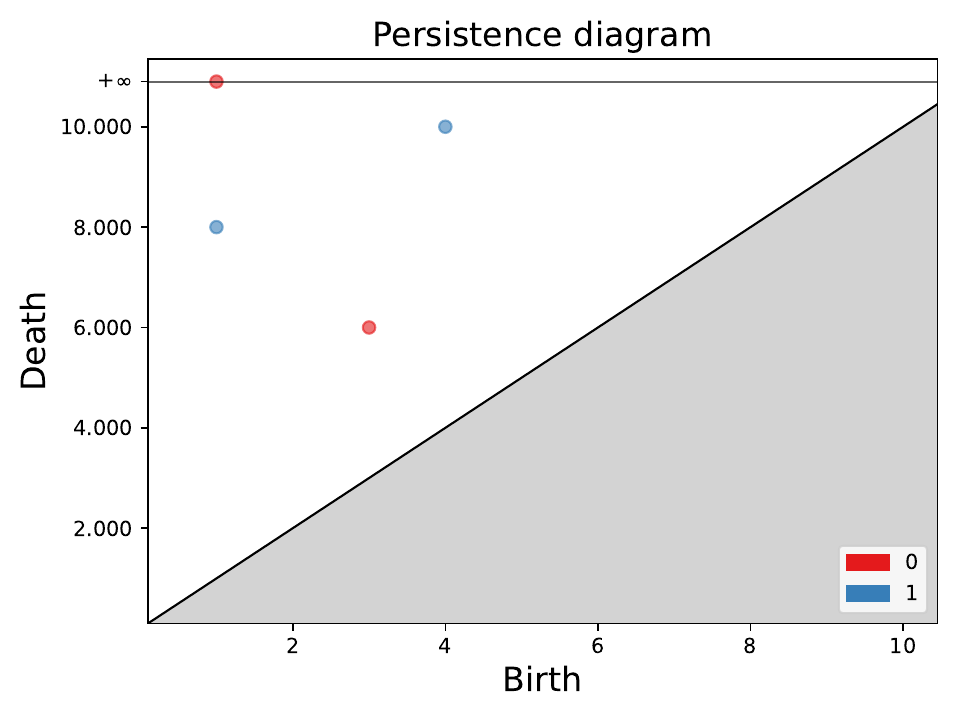}
\centering{(b)}
\end{minipage}
\caption{Example of a 2D signal and its corresponding persistence diagram, with
(a) a synthetic 2D signal with pixel intensities varying from 1 to 10 and (b) its persistence diagram. In the persistence diagram, red points indicate the birth and death of connected components, while blue points indicate the birth and death of one-dimensional holes.
}
\label{fig:PD}
\end{figure}

In this work, the data considered are 2D signals, which can be naturally interpreted as real-valued functions $f$ defined on a topological space or a simplicial complex.
The filtration is constructed considering sets of the form $X_t = f^{-1}((-\infty, t])$, where $t \text{ ranges in } \R$ and $f^{-1}((-\infty, t])$ is the inverse image of the interval $(-\infty, t]$. We refer to $X_t$ as a sublevel set. 
As illustrated in Fig.~\ref{fig:PD}, TDA provides a powerful framework for quantifying structural features in 2D signals through persistent homology. In Fig.~\ref{fig:PD}(a) we visualize an example of synthetic 2D signal with pixel intensities varying from 1 to 10 and in Fig.~\ref{fig:PD}(b) its persistence diagram. 
As the sublevel threshold $t$ increases, dark regions (intensity $\le t$) appear first and eventually merge or fill in. Two connected components are highlighted by light-blue dashed outlines, and two holes (enclosed bright regions) are marked by red dashed outlines. At $t=1$, the first connected component (red dot in Fig.~\ref{fig:PD}(b)) appears, and a hole (blue dot in Fig.~\ref{fig:PD}(b)) appears when its surrounding boundary closes. A second connected component is born at $t=3$ but cannot form a hole until the intensity-4 “barrier” pixel (circled in black) is added at $t=4$. Once this barrier pixel activates, a new hole emerges. One connected component (the lower-right blob) merges into the first component at $t=6$, causing its death in the $H_0$ diagram, while the two holes disappear at $t=8$ and $t=10$ as the last boundary pixels fill. Note that one component (the initial dark region) persists for all thresholds (an infinite-lifetime feature). Fig~\ref{fig:PD}(b) shows the persistence diagram in (birth, death) coordinates: red and blue markers correspond to the $H_0$ and $H_1$ events described above. Features farther from the diagonal represent longer-lived topological structures.

\begin{figure}[t]
    \centering
    \includegraphics[width=.27\textwidth]{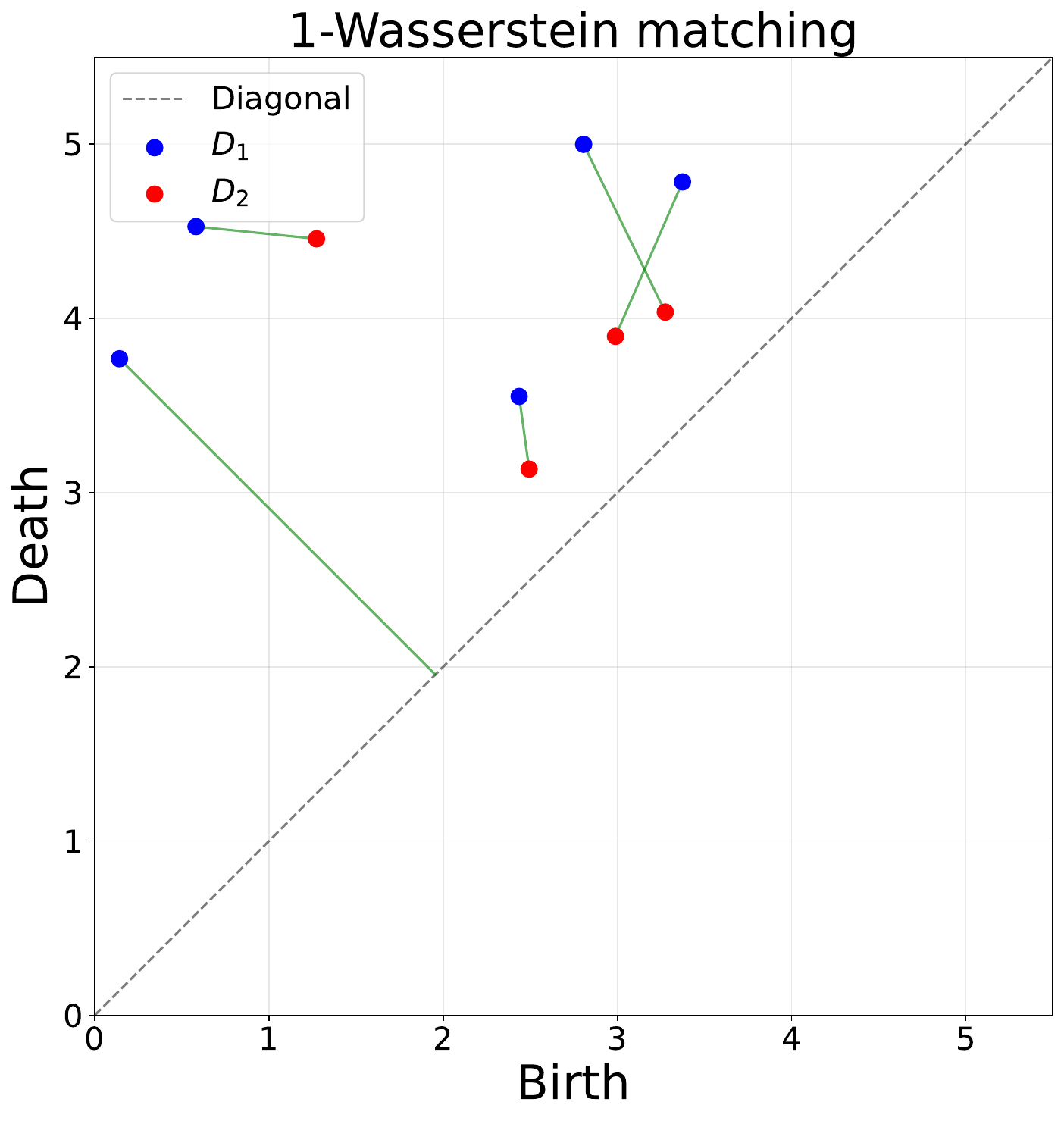}
    \caption{The persistence diagrams $\mathcal{D}_1$ and $\mathcal{D}_2$ are matched through their optimal $1$-Wasserstein matching.}
    \label{fig:1W}
\end{figure}

We can endow the space of persistence diagrams with a metric that aims to quantify the topological differences.
\begin{definition}
\label{def:p-w}
    Given two persistence diagrams $\mathcal{D}_1, \mathcal{D}_2$ and $p \in \mathbb{N}$, the \emph{p-Wasserstein metric} is
    \begin{equation}
    W_p(\mathcal{D}_1, \mathcal{D}_2) = \inf_{\gamma} \left(  \sum_{x \in \mathcal{D}_1} \|x - \gamma(x)\|_\infty^p \right)^{1/p},
    \end{equation}
    where $\gamma$ ranges over all bijections between $\mathcal{D}_1$ and $\mathcal{D}_2$.
\end{definition}
We considered the $1$-Wasserstein metric since it is particularly suitable for our analysis as it weights all features proportionally to their persistence, and maintains stability against small perturbations of input data. We show in Figure~\ref{fig:1W} the optimal matching between two persistence diagrams $\mathcal{D}_1$ and $\mathcal{D}_2$. 

\subsection{THEORETICAL FOUNDATIONS OF GENEOS}

We define an agent as a functional operator that transforms data preserving important invariances and symmetries and we formalize it as a GENEO. 
In our theoretical framework (see \cite{bergomi2019towards}), a data set is   given by a  set of bounded real-valued measurements on some suitable domain $X$:
\begin{equation}
    \Phi=\{\varphi\colon X\to \mathbb{R}\} \subseteq \mathbb{R}^X_b~,
\end{equation}
where $\mathbb{R}^X_b$ is the set of all bounded real-valued functions on the set $X$.
We can think of $X$ as the space where one makes measurements, and of $\Phi$ as the set of admissible measurements. For example, a 
wireless signal can be represented as a function $\varphi$ from the real plane $X$ to the real numbers. 

The concept of symmetry is modeled as a group action. In our context, we consider the group action of planar rigid motions on signals. The consecutive application of two rigid motions is equivalent to applying their composition, and the action of the identity rigid motion leaves the signal unchanged.
A data set $\Phi$ is naturally equipped with a group action:
\begin{equation}
    \rho\colon\Phi\times \mathrm{Aut}_{\Phi}(X)\to \Phi,\ \ \ \ (\varphi,g)\mapsto \varphi g~,
\end{equation}
where $\varphi g$ is the usual function composition and $\mathrm{Aut}_{\Phi}(X)$ is the group of all bijections $g$ such that $ \varphi g, \varphi g^{-1}\in \Phi.$
Thus $\Phi$ is not just a set, but a set with an action of the group $\mathrm{Aut}_{\Phi}(X)$. 
To encode the symmetries of $\Phi$ 
induced by this action, we  consider its perception pairs.
\begin{definition}
	A \emph{perception pair} is a pair $(\Phi,G)$ with $\Phi \subseteq \mathbb{R}^X_b$ and $G\subseteq\mathrm{Aut}_{\Phi}(X)$.
\end{definition}
The choice of $G$ encodes certain  symmetries of $\Phi$. GENEOs enable us to transform datasets (or perception pairs) while preserving symmetries and not increasing distances. In other words, data transformations are modelled as sets of GENEOs in our framework.

\begin{definition} 
	Consider two perception pairs $(\Phi,G)$, $(\Psi,H)$. A map $(F,T):(\Phi,G) \to(\Psi,H)$  is said to be a \emph{GEO} from $(\Phi,G)$ to $(\Psi,H)$ if $F$ is $T$-equivariant (i.e., $F(\varphi g)=F(\varphi) T(g)$ for every $\varphi\in\Phi, \ g\in G$) and $T$ is a group homomorphism. Moreover, if $F$ is non-expansive (i.e., $\Vert F(\varphi_1) - F(\varphi_2) \Vert_\infty \le \Vert \varphi_1 - \varphi_2 \Vert_\infty $ for every $\varphi_1, \varphi_2 \in\Phi$), then we say $(F,T)$ is a \emph{GENEO}.

\end{definition}

\cyan{A novel GENEO specifically designed for signal reconstruction tasks under conditions of extremely sparse sampling is presented in Sec. \ref{sec:geneo_rec}.}


\section{RELATED WORK}
\label{sec:related_work}

Traditional wireless signal reconstruction techniques often rely on spatial interpolation and model-based recovery. Classical methods such as inverse‐distance weighting (IDW) or radial basis functions (RBF) interpolate radio maps from sparse samples, and geostatistical kriging has been widely used to fuse measurements into smooth coverage fields \cite{Cressie1993}\blue{, including specific adaptations tailored for robust radio map reconstruction \cite{kriging2018}}. Compressive sensing and variational approaches have also been applied to channel reconstruction, exploiting sparsity or smoothness to recover signals from limited measurements \cite{Donoho2006}. For example, graph signal processing interprets the radio map as a signal on a graph and performs spectral interpolation; this can outperform local methods like IDW or RBF interpolation \cite{Redondi2018}. However, these approaches require strong prior assumptions (e.g. smoothness, known propagation models) and often oversmooth sharp features. In practice, traditional kriging or interpolation can be computationally expensive in high dimensions and may struggle to model complex multi-path or interference environments.

In recent years, data-driven neural methods have emerged as powerful alternatives. CNNs have been used to predict radio propagation or coverage maps from environment inputs, essentially treating the problem as image-to-image regression \cite{Imai2019}. Recurrent models (e.g. long-short term memory (LSTM)) capture temporal correlations in dynamic scenarios: for instance, an LSTM-based architecture was shown to reconstruct time-varying road-side radio environment maps (REMs) for V2X vehicular networks with high accuracy \cite{Roger2023}. Graph neural networks (GNNs) have been applied to REM completion by leveraging spatial connectivity: multi-source GNNs learn to fuse sparse spectra from monitoring stations into full \ac{RSRP} or \ac{SINR} maps, improving prediction in heterogeneous domains \cite{Wen2024}. Self-supervised techniques have also been explored. For example, deep image prior treats the radio map as an image and fits an untrained neural network directly to the observed measurements, thereby reconstructing missing values without any external training data \cite{Ulyanov2017}. Neural architecture search (NAS) and semi-supervised learning can further improve performance with little data: a recent \ac{RSS}-mapping system uses NAS combined with self-training on unlabeled points, achieving lower error than traditional RBF, Kriging or manually designed nets under extreme sparsity \cite{Malkova2021}. These deep and graph-based approaches generally achieve higher fidelity than classical methods, especially with abundant data, but they tend to require large training sets, careful tuning, and may still produce unrealistic artifacts when data is very limited \cite{Liu2025}.

Complementary approaches model uncertainty and incorporate generative modeling. Bayesian and probabilistic deep networks have been used to quantify confidence in reconstructed maps, which is crucial for network optimization. Gaussian process regression, for example, provides uncertainty bounds in interpolating coverage maps, though it scales poorly with data size. More recently, conditional generative models treat map reconstruction as a data generation task. \blue{For instance, \acp{CVAE} have been successfully employed to reconstruct radio maps by learning the conditional distribution of the spatial signals \cite{cvaeReconstruction}. Similarly, approaches like} \cite{Skocaj2024} introduce a Bayesian framework for uncertainty-aware generation of mobile data. GUMBLE can produce synthetic signal measurements conditioned on partial real samples, effectively augmenting sparse crowdsourced datasets. This line of work highlights the interplay between reconstruction and data augmentation: a generative model that fills in missing radio map data can be seen as performing reconstruction while also creating diverse samples for downstream learning. However, generative methods often lack hard constraints on physical consistency and may suffer from mode collapse or bias if the training data is unbalanced.

Another emerging trend is leveraging geometric and topological priors. TDA techniques have been applied to wireless signals to capture global structure: for instance, persistent homology and persistence images of radio frequency time-series have been used for modulation classification and anomaly detection \cite{Myers2023}. These methods show that capturing the topology of signal features can improve interpretability. More broadly, group-equivariant learning has proven effective in vision and signal tasks: Cohen et al. \cite{Cohen2016} showed that neural networks designed to respect transformations (rotations, translations, scaling) can dramatically reduce sample complexity. Building on these ideas, GENEOs have been proposed as a theoretical framework for embedding symmetries and stability into learning \cite{bergomi2019towards, Frosini2023}. A GENEO is an operator commuting with a group action (e.g. shifting coordinates) and satisfying a 1-Lipschitz property. In practice, GENEO-based models (e.g. GENEOnet) have demonstrated that injecting equivariance and topological constraints yields compact, convex model spaces and good performance even with small training sets. In wireless signal mapping, equivariance could encode invariance of wireless signals under certain domain transformations (for example, translations of the coordinate frame) or known physical symmetries, while non-expansiveness enforces robustness to noise. 
By contrast, conventional CNNs or GNNs cannot incorporate by design diverse symmetries and therefore must rely on substantially larger datasets to achieve comparable generalization.

\blue{In summary, while existing literature offers a broad spectrum of techniques for radio map reconstruction, a common set of critical limitations persists. Traditional spatial methods (e.g., Kriging) struggle to maintain accuracy under extreme data scarcity. On the other hand, advanced data-driven and generative methods (e.g., CNNs, CVAEs) heavily depend on massive training datasets and often fail to preserve the topological integrity of the signal when measurements are highly sparse or corrupted. This reveals the need for a robust and lightweight reconstruction framework that does not rely on vast amounts of training data. By inherently embedding geometric symmetries (e.g., translation equivariance), our proposed GENEO-based approach directly addresses these shared shortcomings. It offers a mathematically grounded solution aiming at preserving critical signal topology and ensuring robust reconstruction even under severely constrained sampling and high-noise conditions.}

\section{GENEOS FOR SIGNAL RECONSTRUCTION}

\label{sec:geneo_rec}

\begin{figure}[t]
\setkeys{Gin}{width=1\linewidth}
\centering
\begin{minipage}[t]{0.24\textwidth}
\label{fig:3a}
\includegraphics{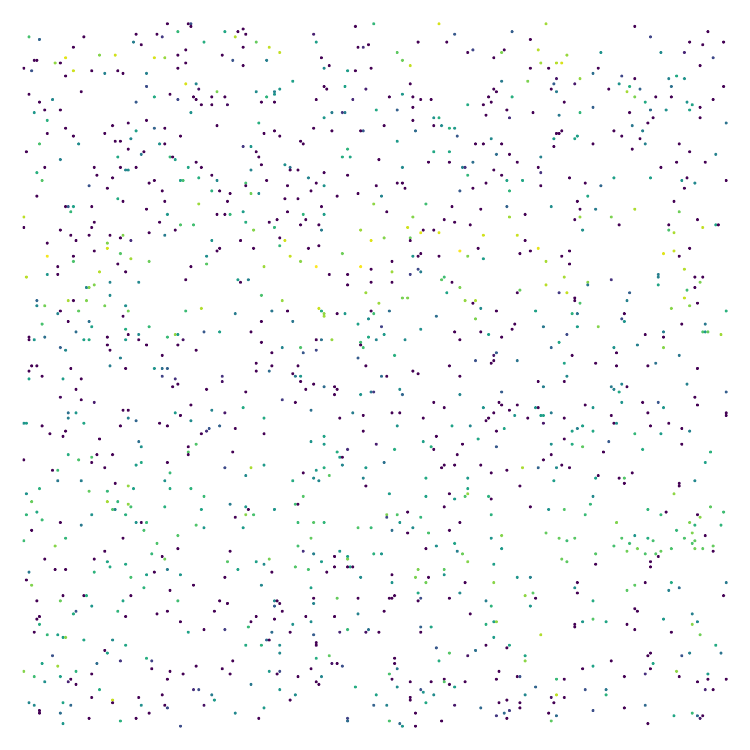}
\centering{(a)}
\end{minipage}
\begin{minipage}[t]{0.24\textwidth}
\label{fig:3b}
\includegraphics{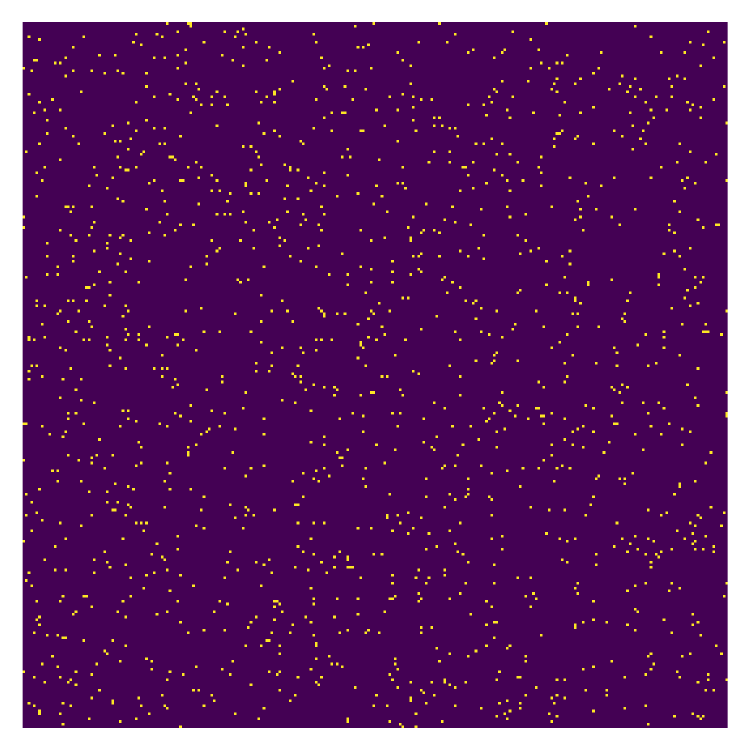}
\centering{(b)}
\end{minipage}
\caption{Visualization of (a) $\hat{\varphi}$ values and (b) $\hat{\psi}$ values. In (a), each point $p$ such that $\hat\psi(p) = 1$ is colored based on $\hat\varphi(p) \in [0, 1]$; remaining areas are shown in white. In (b), the color of each point $p$ represents the intensity of $\hat\psi(p)$ which takes only binary values in this example. The color purple indicates the zero value, while the yellow indicates the ones.}
\label{fig:phi_psi}
\end{figure}

We describe the signal to be reconstructed as a function $\varphi:\mathbb{R}^2\to [0,1]$. The function $\varphi$ represents the intensity of the signal over a 2D grid, and is often referred to as ground truth or GT.
We also consider the function $\psi:\mathbb{R}^2\to [0,1]$
taking each point $p$ to the reliability $\psi(p)$
of the value $\varphi(p)$ measured at the point $p$\footnote{In our work, we define a point as $p = (x,y) \in \mathbb{R}^2$.}. 
We aim at reconstructing $\varphi$ by starting from a very poor sampling of $\varphi$, by using GENEOs. In literature, GENEOs accept as input only continuous functions that share the same domain, but samplings are functions that do not possess these properties. An analogous statement holds for the corresponding sampling of $\psi$. To obtain $\hat{\varphi}$ and $\hat{\psi}$, namely the continuous functional representations with the same domain extracted from $\varphi$ and $\psi$, let us assume that we know the intensity $\varphi(p_i)$ of the ground truth signal at $r$ points $p_1,\ldots,p_r$ and the reliability $\psi(p_i)$ of each value $\varphi(p_i)$. 
We can now define the functions $\hat\varphi,\ \hat\psi:\mathbb{R}^2\to[0,1]$ by setting 
\begin{align}
\begin{split}
    \hat\varphi(p):=&\sum_{i=1}^{r} \varphi(p_i) e^{-\frac{\|p-p_i\|^2}{2\sigma^2}}\ ,
\mathrm{\ \ }\\
\hat\psi(p):=&\sum_{i=1}^{r} \psi(p_i) e^{-\frac{\|p-p_i\|^2}{2\sigma^2}}
\end{split}
\end{align}
for a small value of $\sigma$. 
In our implementation, $\sigma$ is chosen sufficiently small such that the continuous functions $\hat{\varphi}$ and $\hat{\psi}$ can be approximated by two arrays consisting entirely of zero-valued pixels, except for a number less or equal to $r$ of pixels with nonzero values.
We refer to Fig.~\ref{fig:phi_psi} for an example of $\hat{\varphi}$ (Fig.~\ref{fig:phi_psi}(a)) and $\hat{\psi}$ (Fig.~\ref{fig:phi_psi}(b)). The color scheme for $\hat{\varphi}$ is as follows: the image takes the value $\varphi(p) \in [0, 1]$ if $\hat{\psi}(p) = 1$, and is white otherwise.

Now we present how we built GENEOs for the reconstruction of $\varphi$. The main idea is to look for meaningful patterns in the signal.
In our model, a pattern is a pair of functions $P=(h,\chi_h)$, where:
\begin{itemize}
    \item $h  \colon \mathbb{R}^2\to [0,1]$ represents a shape we are looking for in the signals;
    \item $\chi_h \colon \mathbb{R}^2 \to \{0,1\}$ establishes where the values expressed by $h$ are reliable.
\end{itemize}
Furthermore, we assume that the support $D_h$ of $\chi_h$ is compact, i.e., closed and bounded. This assumption implies that the effect of each pattern is local.

Let us now consider a pair $S=(f,\psi_f)$ of functions from $\mathbb{R}^2$ to $[0,1]$. In this paper, $f = \hat \varphi$ and $\psi_f = \hat \psi$, but we want to emphasize that our model is applicable to broader contexts. After choosing the pair $S$ representing the data and the pair $P=(h,\chi_h)$ representing a pattern, we can define \cyan{the following non-linear operator}:
\begin{align}
\begin{split}
\label{eq:s}
    \mathcal{S}_{S,P}(x,y)=\int_{\mathbb{R}^2}\lvert &{f}(x+\xi,y+\eta) -h(\xi,\eta) \rvert \\
    &\cdot \psi_f(x+\xi,y+\eta) \chi_h(\xi,\eta) \ d\mu(\xi,\eta)~,
\end{split}
\end{align}
which tells us how much $S$ differs from $P$ in a neighborhood of the point $(x,y)$. Also, we define
\begin{align}
\begin{split}
\label{eq:a}
    \mathcal{A}_{S,P}(x,y)=\int_{\mathbb{R}^2} \psi_f(x+\xi,y+\eta)\chi_h(\xi,\eta)\ d\mu(\xi,\eta)~,
\end{split}
\end{align}
which quantifies the reliability of our data in the region where we can perform the comparison between $S$ and $P$. In Eq.\eqref{eq:s} and  Eq.\eqref{eq:a}, defining $\bar\mu$ as the standard Lebesgue measure on $\mathbb{R}^2$, we set $\mu(A) = \bar\mu(A) /\bar\mu(D_h) $ for each measurable $A \subseteq \mathbb{R}^2$, where $D_h$ is assumed to have finite positive measure. Hence, we can consider the following GENEO $F$ that takes the function $f$ to the function $\hat c_{S,P}$ defined by setting:

\begin{equation}
    \hat c_{S,P}(x,y) = \mathcal{A}_{S,P}(x,y) - \mathcal{S}_{S,P}(x,y)~.
\end{equation}
The value $\hat c_{S,P}(x,y)$ measures the similarity between the pattern described by $h$ and the structure of $f$ in the neighborhood of $(x,y)$. 

\begin{proposition}\label{prophatC}
$0 \le \hat c_{S,P}(x,y) \le \mathcal{A}_{S,P}(x,y) \le 1$ for every $(x,y) \in \mathbb{R}^2$.
\end{proposition}
\begin{proof}
The proof is provided in Appendix A.
\end{proof}

\begin{proposition}\label{propA}
Assume that two signals $S_1=(f_1,\psi_{f_1})$, $S_2=(f_2,\psi_{f_2})$ and a pattern $P=(h,\chi_h)$ are given. If $\psi_{f_1}\equiv\psi_{f_2}$, it holds that $\mathcal{A}:=\mathcal{A}_{S_1,P}=\mathcal{A}_{S_2,P}$ and 
\begin{equation}
    \lVert \hat c_{S_1,P} - \hat c_{S_2,P} \rVert_\infty \le \lVert \mathcal{A} \rVert_\infty \lVert f_1 - f_2 \rVert_\infty.
\end{equation}
\end{proposition}
\begin{proof}
The proof is provided in Appendix A.
\end{proof}

\cyan{In this work, we assume that the ground truth $\varphi$ is reliable everywhere (i.e., $\psi \equiv 1$). We recall that the ground truth is the true signal we aim to reconstruct from an extremely sparse sample $\hat{\varphi}$.} Hence, the resulting sampling reliability $\hat \psi$ takes value in $\{0, 1\}$, where $\hat \psi(x,y)=1$ if $(x,y)$ is a known sample, otherwise $\hat \psi(x,y)=0$. \cyan{In addition, we assume that $\chi_h \equiv 1$ in the support for every pattern $P$.}
\begin{proposition}\label{propA2}
The map $F$ taking $S=(f,\psi_f)$ to the function $\hat c_{S,P}$ is a GENEO with respect to the group of planar translations.
\end{proposition}
\begin{proof}
From Proposition \ref{propA} and the translation invariance of the Lebesgue measure, the statement immediately follows.
\end{proof}

\begin{remark}
Depending on the geometry of $D_h$ and the choice of patterns, we can allow $F$ to be equivariant w.r.t. a larger group, i.e. the group of isometries of the plane.
\end{remark}

Given a sampled signal $\hat S=(\hat \varphi,\hat \psi)$,  we consider a \textit{library} of $N$ patterns to reconstruct $\varphi$, that is
\begin{equation}
    \{P_1=(h_1,\chi_{h_1}), \dots, P_N=(h_N,\chi_{h_N})\}~.
\end{equation}
Therefore, we can compute $\{\hat c_{\hat S,P_i}\}_{i = 1}^N$.  For any index $i$ we define the similarity coefficient of the pattern $P_i$ at the point $p$, when $P_i$ is centered at another point $q\in\mathbb{R}^2$, as

\begin{equation}
\mathrm{sim}(p,P_i,q) = \hat c_{\hat S,P_i}(q)\chi_{h_i}(p-q)~.
\end{equation}

\begin{figure}[t]
\setkeys{Gin}{width=1\linewidth}
\centering
\begin{minipage}[t]{0.075\textwidth}
\label{fig:4a}
\includegraphics{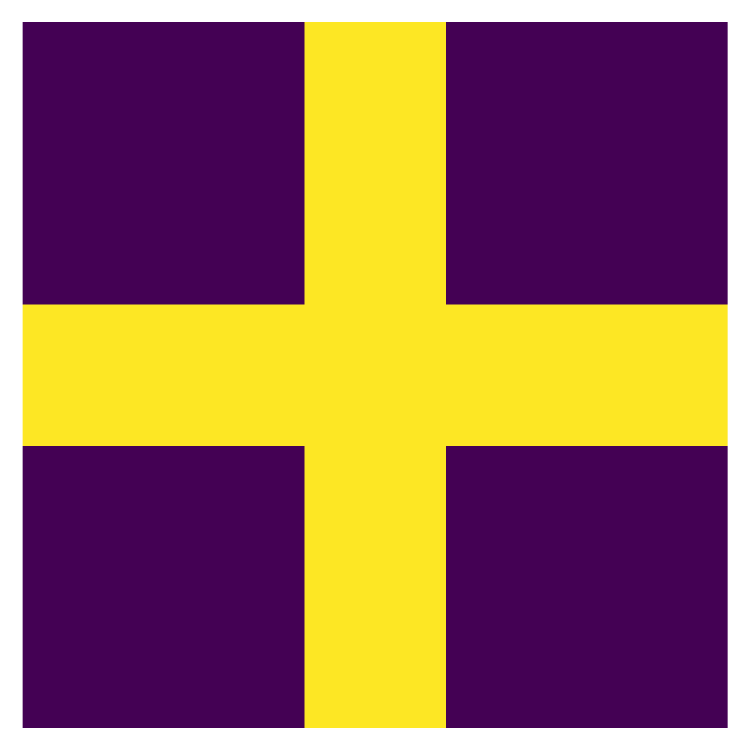}
\centering{(a)}
\end{minipage}\hfill
\begin{minipage}[t]{0.075\textwidth}
\label{fig:4b}
\includegraphics{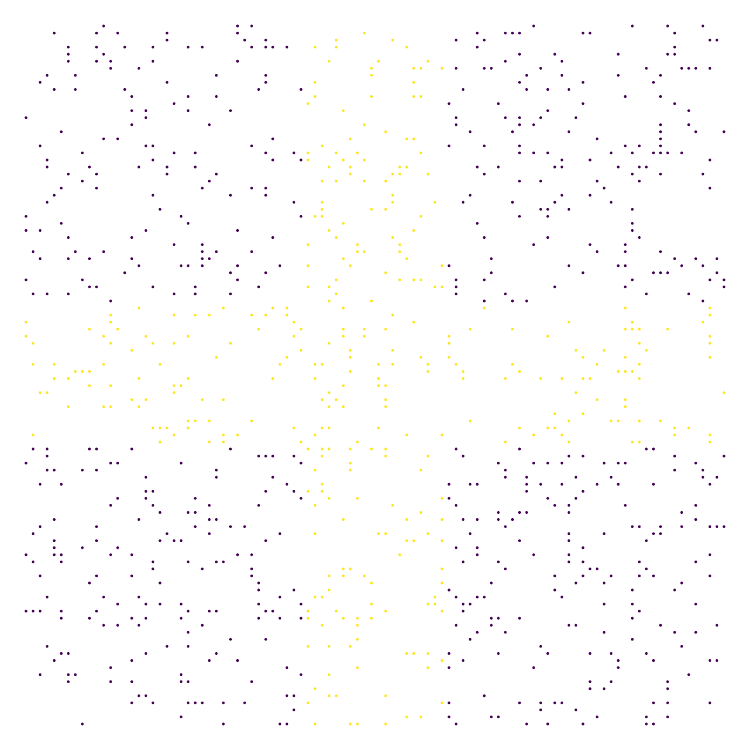}
\centering{(b)}
\end{minipage}\hfill
\begin{minipage}[t]{0.075\textwidth}
\label{fig:4c}
\includegraphics{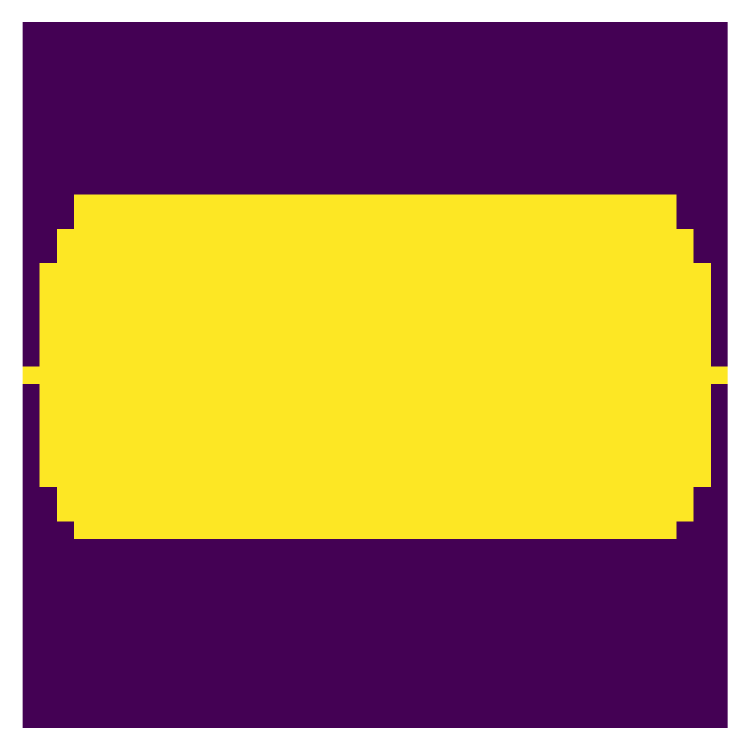}
\centering{(c)}
\end{minipage}\hfill
\begin{minipage}[t]{0.075\textwidth}
\label{fig:4d}
\includegraphics{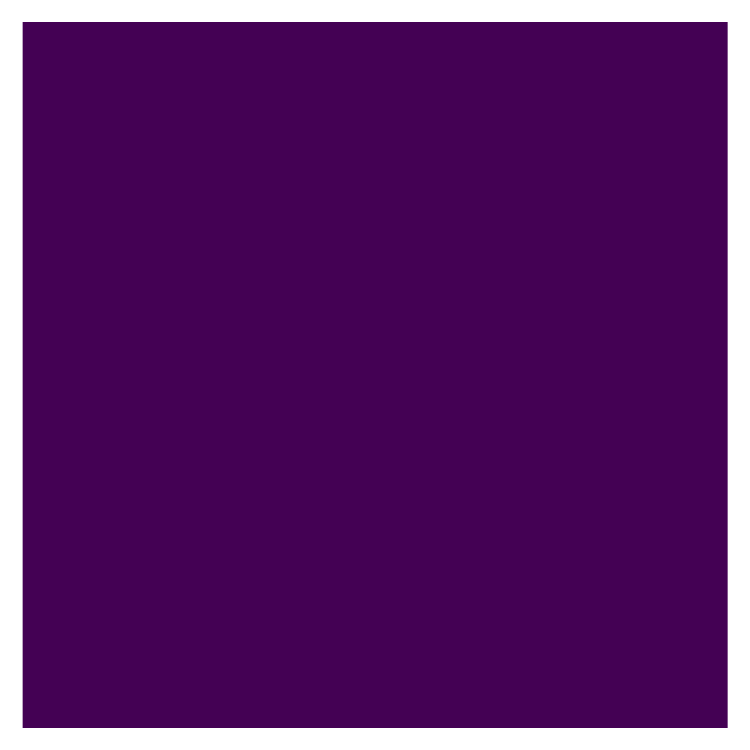}
\centering{(d)}
\end{minipage}
\begin{minipage}[t]{0.075\textwidth}
\label{fig:4e}
\includegraphics{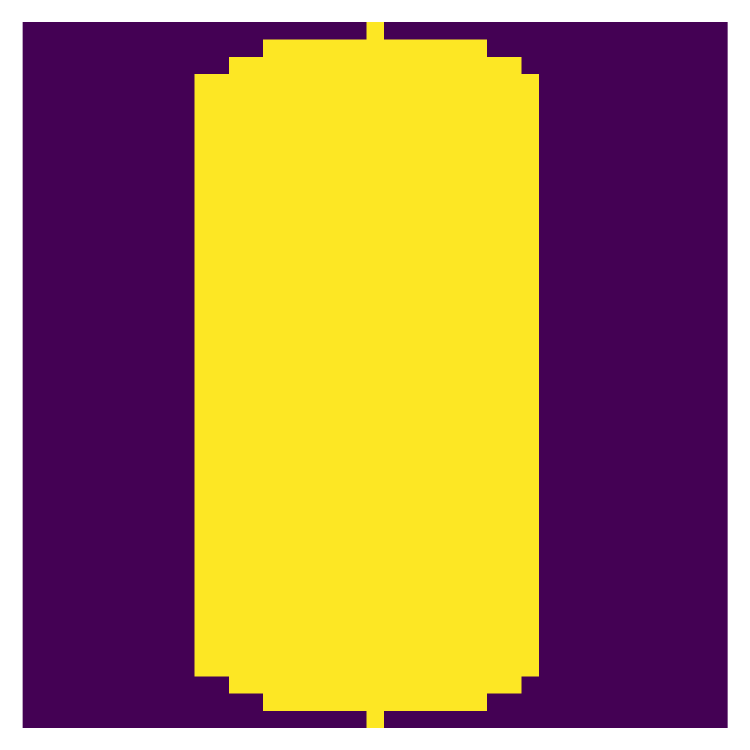}
\centering{(e)}
\end{minipage}
\begin{minipage}[t]{0.075\textwidth}
\label{fig:4f}
\includegraphics{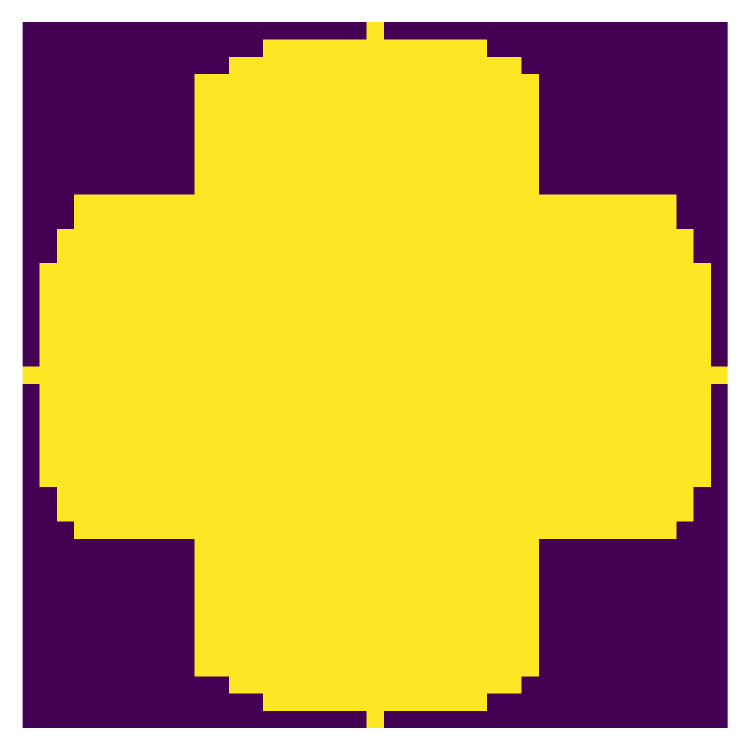}
\centering{(f)}
\end{minipage}
\caption{Illustrative toy example, where we present (a) a ground truth $\varphi$, (b) its sampled version $\hat{\varphi}$, and (c)-(f) four patterns $h_1\chi, \dots, h_4\chi$.}
\label{fig:toy}
\end{figure}

\begin{figure}[t]
\setkeys{Gin}{width=1\linewidth}
\centering
\begin{minipage}[t]{0.1125\textwidth}
\label{fig:5a}
\includegraphics{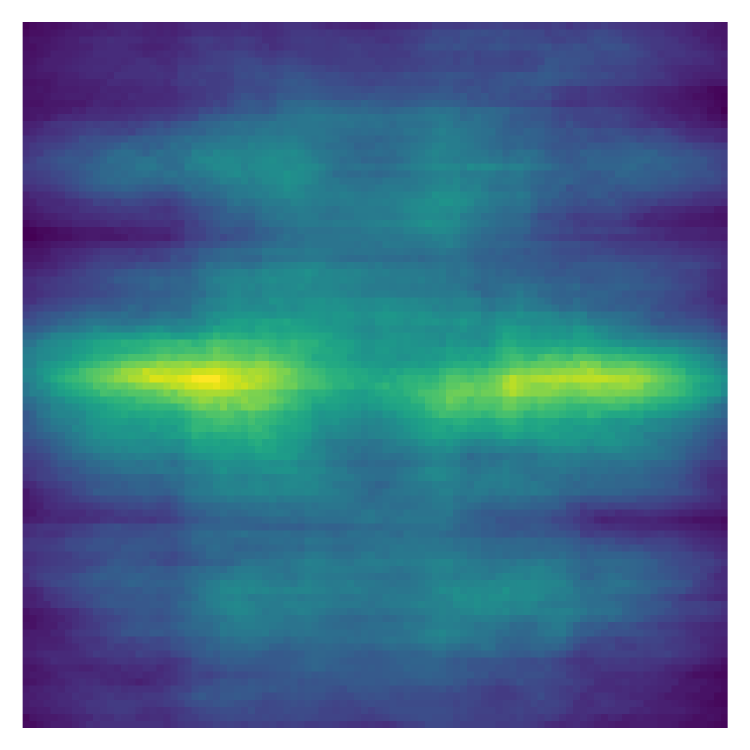}
\centering{(a)}
\end{minipage}
\begin{minipage}[t]{0.1125\textwidth}
\label{fig:5b}
\includegraphics{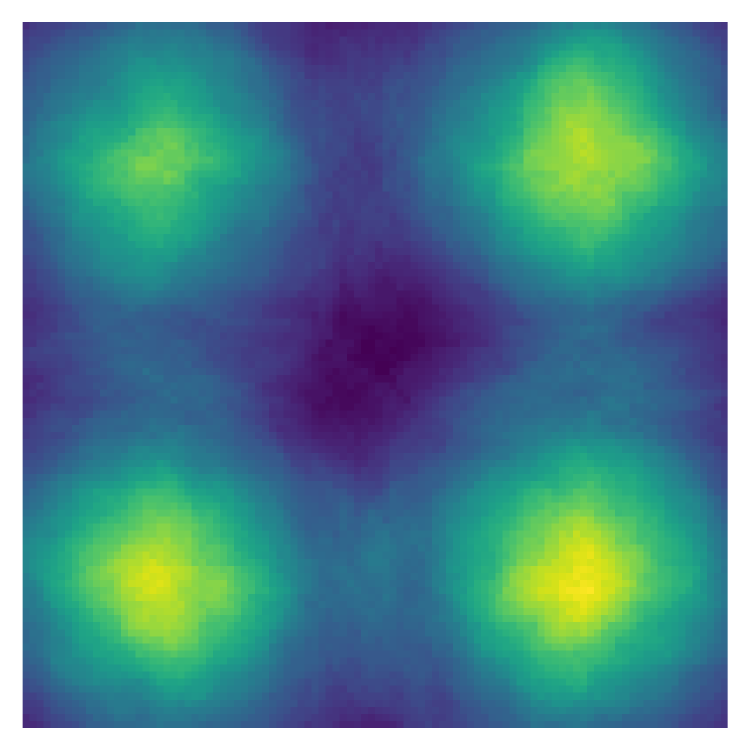}
\centering{(b)}
\end{minipage}
\begin{minipage}[t]{0.1125\textwidth}
\label{fig:5c}
\includegraphics{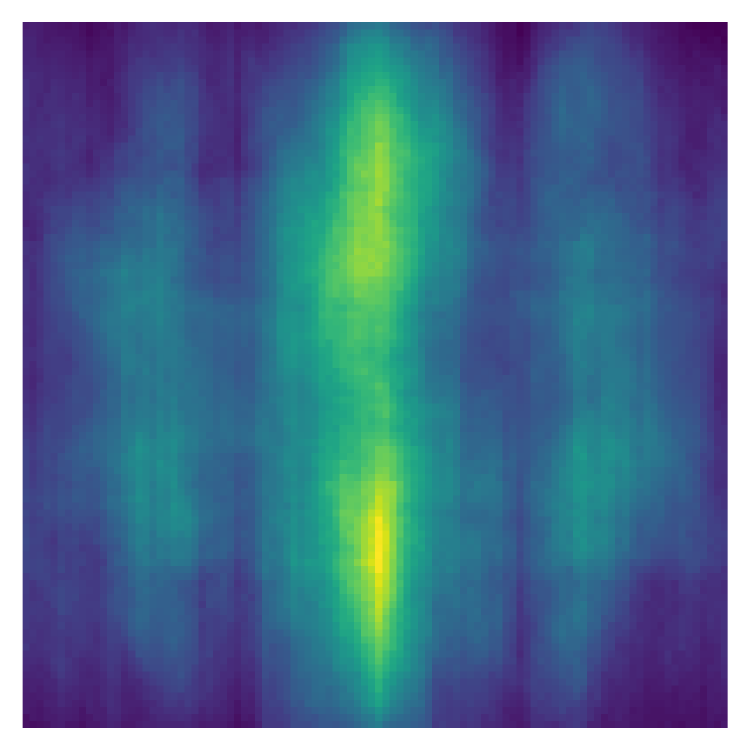}
\centering{(c)}
\end{minipage}
\begin{minipage}[t]{0.1125\textwidth}
\label{fig:5d}
\includegraphics{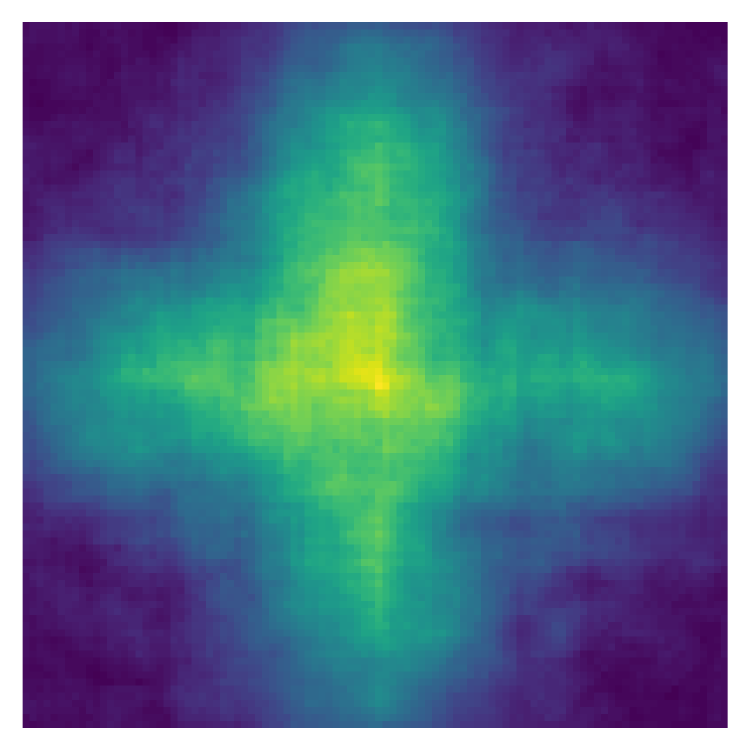}
\centering{(d)}
\end{minipage}
\caption{Heatmaps of the coefficients 
  \(\,\hat{c}_{S,P_i}\) for \(i=1,2,3,4\), 
  where 
  \(S = (\hat\varphi,\hat\psi)
    \quad\text{and}\quad
    P_i = (h_i,\chi)
    \,,
  \)
  with \(\hat\varphi,h_i,\chi\) as defined in Fig.~\ref{fig:toy}. 
  Sub-figures (a)–(d) correspond to \(i=1,2,3,4\), respectively.}
\label{fig:c_hats}
\end{figure}

In plain words, the value $\mathrm{sim}(p,P_i,q)$ quantifies how plausible it is that at the point $p$ the dominant pattern is $P_i$, centered at the point $q$. The term $\chi_{h_i}(p-q)$ is needed because we are interested only in the points of $P_i$ at which the pattern is reliable (i.e., $\chi_{h_i}(p-q)\neq 0$). Fig.~\ref{fig:toy} shows (from left to right) a toy example of $\varphi$, its sampled version \cyan{$\hat{\varphi} = \varphi \hat{\psi}$} with a $10\%$ sampling rate and four patterns $h_1, \dots, h_4$, displayed as $h_1 \chi, \dots, h_4\chi$, where $\chi$ is the characteristic function of the unit disk. Fig.~\ref{fig:c_hats} displays the corresponding $\hat{c}_{S,P_i}$, where $S = (\hat{\varphi}, \hat{\psi})$ and $P_i = (h_i, \chi)$ for $i=1, 2, 3, 4$. 

\begin{figure}[t]
\setkeys{Gin}{width=1\linewidth}
\centering
\begin{minipage}[t]{0.16\textwidth}
\label{fig:6a}
\includegraphics{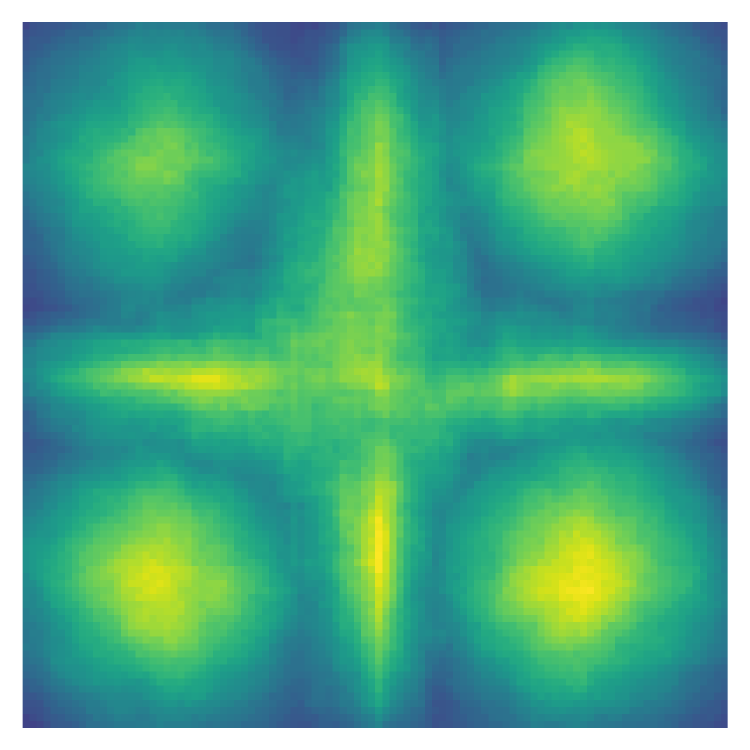}
\centering{(a)}
\end{minipage}\hfill
\begin{minipage}[t]{0.16\textwidth}
\label{fig:6b}
\includegraphics{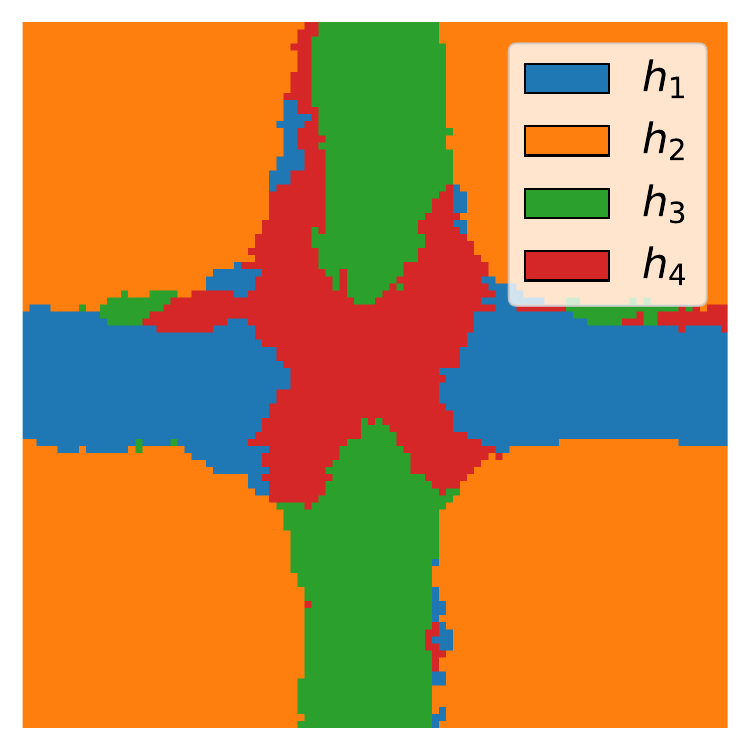}
\centering{(b)}
\end{minipage}\hfill
\begin{minipage}[t]{0.16\textwidth}
\label{fig:6c}
\includegraphics{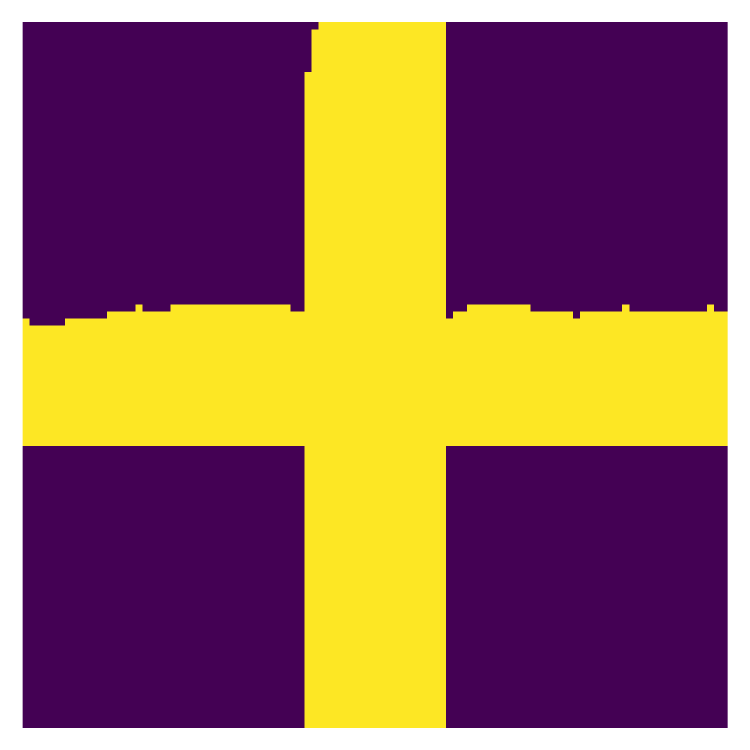}
\centering{(c)}
\end{minipage}\hfill
\caption{(a) Heatmap of the maximum confidence 
\(\displaystyle \max_{i} \hat{c}_{S,P_i}(p)\) 
at each point \(p\), indicating how well the best-matching pattern \(P_{\bar{\imath}}\) locally agrees with the original sample \(\varphi(p)\) displayed in Fig.~\ref{fig:toy}(a). 
(b) Map of indices 
\(\displaystyle \bar{\imath}(p) = \underset{i}{\mathrm{argmax}}\ \hat{c}_{S,P_i}(x,y)\), 
showing which pattern \(P_{\bar{\imath}}\) achieves the highest confidence at each point. 
(c) Reconstructed output 
\(\varphi_{\mathrm{rec}}(p)\) 
produced by the GENEO-based pattern matching.
}

\label{fig:c_best}
\end{figure}

The last step in our procedure consists of using the functions $\mathrm{sim}(\cdot,P_i,q)$, varying $P_i$ and $q$, to produce the reconstructed signal $\varphi_{\mathrm{rec}}\colon \mathbb{R}^2\to [0,1]$. 
\cyan{To do this, we first associate a weight to each similarity coefficient: $W\colon \mathrm{sim}(p,P_i,q) \mapsto w_{i, q}(p) \in \R$, which represents the weight of the $i$-th pattern centered at $q$ for the reconstruction of the signal at position $p$. The reconstructed signal is then defined as}
\begin{equation}
\label{eq:rec}
\varphi_{\mathrm{rec}}(p) := \sum_{i=1}^{N} \int_{\R^2} w_{i,q}(p) h_i(p - q) \, dq\,.
\end{equation}
%
\blue{The proposed framework is specifically designed for signal reconstruction tasks under conditions of extremely sparse sampling and is flexible enough to accommodate different weights $w_{i,q}(p)$. For example, to only account for the pattern achieving the highest similarity, it suffices to select $w_{\bar \imath,\bar q}(p)$~=~1} only for $(\,\bar q(p),\,\bar\imath(p)\,)$ such that 
\begin{equation}
    (\,\bar q(p),\,\bar\imath(p)\,)
\;=\;
\underset{\substack{i\in\{1,\dots,N\} \\ q\in \mathbb{R}^2}}{\mathrm{argmax}}\;\mathrm{sim}\bigl(p,\,P_i,\,q\bigr)\,,
\end{equation}
\blue{while selecting $w_{i,q}(p) = 0$ for all $(i,q) \neq (\bar q(p),\,\bar\imath(p))$}. 
\blue{In this case, the reconstruction of Equation~\eqref{eq:rec} reduces to}
\begin{equation}
\label{eq:recon_argmax}
   \varphi_{\mathrm{rec}}(p):= h_{\bar \imath (p)}\Big(p - \bar q(p)\Big)\,,
\end{equation}
where the reconstruction of the point $p$ occurs via the $\bar \imath (p)$-th pattern with a shift equal to $\bar q(p)$. \blue{In the remainder of the paper, we refer to this reconstruction strategy as \emph{argmax} reconstruction. }
\cyan{Alternatively, 
we can select 
the $K$ pairs $(i, q) \in \{1, \ldots, N\} \times \mathbb{R}^2$ with largest similarity coefficients and assign to each the weight}
\begin{equation}
    w_{i,q}(p) \;=\; \frac{\exp\left(\mathrm{sim}(p,\, P_{i},\, q)\right)}{\displaystyle\sum_{(j,\, q') \in \mathcal{T}_K(p)} \exp\left(\mathrm{sim}(p,\, P_{j},\, q')\right)} \,,
\end{equation}
\cyan{and $w_{i,q}(p) = 0$ for $(i,q) \notin \mathcal{T}_K(p)$, where $\mathcal{T}_K(p)$ denotes the set of such $K$ pairs. 
} 
\blue{In the remainder of the paper, we refer to this reconstruction strategy as \emph{softmax} reconstruction.}
Continuing the toy example \blue{and considering the argmax reconstruction strategy}, Fig.~\ref{fig:c_best} (a) displays the maximum among the $\hat c_{S,P_i}$ values for each point. 
This image essentially represents how confident the GENEO is about having found a pattern effectively explaining $\varphi$'s sampling at each point. Fig.~\ref{fig:c_best} (b) is colored according to the index of the pattern that achieves the optimal $\hat c_{S,P_{\bar \imath}}$. Finally, Fig.~\ref{fig:c_best} (c) shows the corresponding reconstruction $\varphi_\mathrm{rec}$ according to Eq.~\eqref{eq:rec}.
We observe that the reconstruction operator that associates $N$-tuple $(\hat c_{S,P_1},\ldots,\hat c_{S,P_N})$ with $\varphi_{\mathrm{rec}}$ is a GEO.
\blue{The pseudo-code detailing the core algorithms of our GENEO-based framework can be found in Appendix B.} 
\cyan{Overall, we emphasize that the proposed GENEO framework restricts the optimization variables exclusively to those required to combine the elements of the pattern library (i.e., to a set of cardinality $N$). The patterns themselves encode the prior knowledge injected into the framework, enabling optimization in a parameter space of significantly lower dimensionality than that typically required by standard deep learning approaches.} 

\section{NUMERICAL RESULTS}
\label{sec:numerical}

\subsection{SIGNAL RECONSTRUCTION}
\label{sec:signal_rec}

In our numerical experiments, we focus on reconstructing the \ac{SINR} over a two-dimensional area. We generate SINR measurements with the Sionna RT ray-tracing simulator \cite{sionna-rt}, using its built-in outdoor urban scenarios for Munich and Paris. Each scenario is discretized into a $L \times L$ grid of $1\,\mathrm{m}^2$ pixels $\{p_j\}_{j=1}^{L^2}$. 
\blue{To accurately simulate complex urban propagation, the ray-tracer is configured to compute up to a maximum depth of 3 ray reflections and explicitly includes diffraction modeling.}

The network contains $N_{\mathrm{tx}}=100$ transmitters, each operating at carrier frequency $f_c = 10\,$GHz with transmit power $P_{\mathrm{tx}} = 1\,$dBm and bandwidth $B_w = 50\,$MHz. Every transmitter is equipped with a planar array of 128 isotropic elements \blue{(configured as 16 rows and 4 columns per polarization) utilizing the 3GPP TR38.901 radiation pattern with vertical-horizontal (VH) polarization}. Similarly, at each receiver pixel $j$, we place a planar array of 16 isotropic elements (2 rows of 8). Denote by $P_{\mathrm{rx}}^{j,i}$ the received power at receiver $j$ from transmitter $i$.

We define the instantaneous SINR at receiver $j$ as
\begin{equation}
\gamma_j
= \frac{\displaystyle \max_{1 \le i \le N_{\mathrm{tx}}} P_{\mathrm{rx}}^{j,i}}
       {\displaystyle \sigma^2
        + \sum_{i=1}^{N_{\mathrm{tx}}} P_{\mathrm{rx}}^{j,i}
        - \max_{1 \le i \le N_{\mathrm{tx}}} P_{\mathrm{rx}}^{j,i}}~,
\end{equation}
where the noise power is
\begin{equation}
    \sigma^2 = k_B \, T_K \, B_w \, F_\sigma~,
\end{equation}
with $k_B = 1.38\times10^{-23}\,\mathrm{J/K}$ (Boltzmann’s constant), $T_K = 290\,$K (system temperature), and $F_\sigma = 5$ dB (noise figure), yielding $\sigma^2 = 10^{-12}$ W.
To obtain a bounded target to reconstruct, we normalize the SINR values across the grid:
\begin{equation}
\bar\gamma_j
= \frac{\max_{1 \le k \le L^2}\gamma_k - \gamma_j}
       {\max_{1 \le k \le L^2}\gamma_k - \min_{1 \le k \le L^2}\gamma_k}~.
\end{equation}
We then pose the reconstruction problem as reconstructing the mapping 
\begin{equation}
\hat\varphi : \{p_j\}_{j=1}^{L^2}
       \;\mapsto\;\{\bar\gamma_j\}_{j=1}^{L^2}~,
\end{equation}
i.e., from each 
pixel to its normalized SINR value.

\begin{figure*}
\setkeys{Gin}{width=1\linewidth}
\centering
\begin{minipage}[t]{0.16\textwidth}
\label{fig:munich_a}
\includegraphics{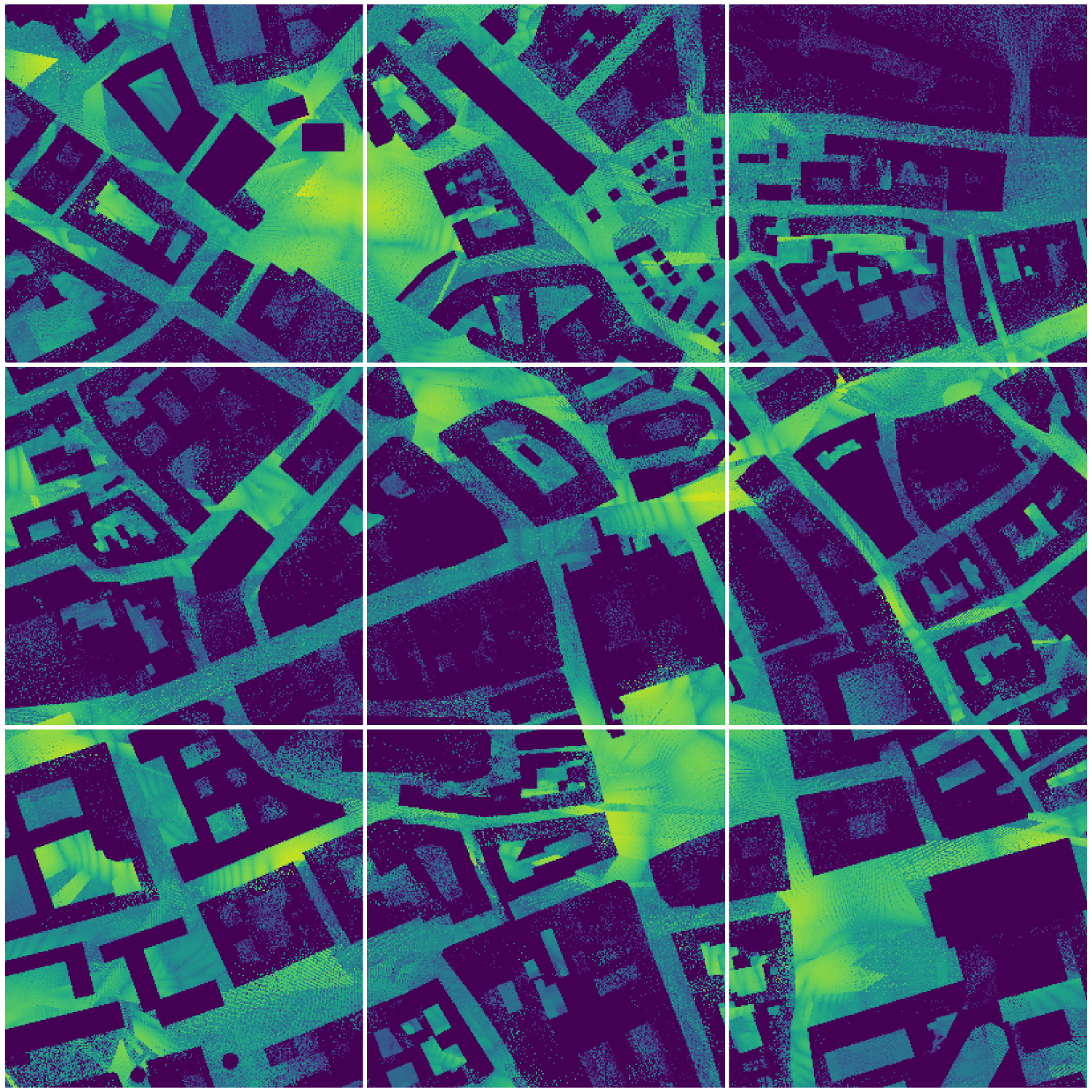}
\centering{(a)}
\end{minipage}\hfill
\begin{minipage}[t]{0.16\textwidth}
\label{fig:munich_b}
\includegraphics{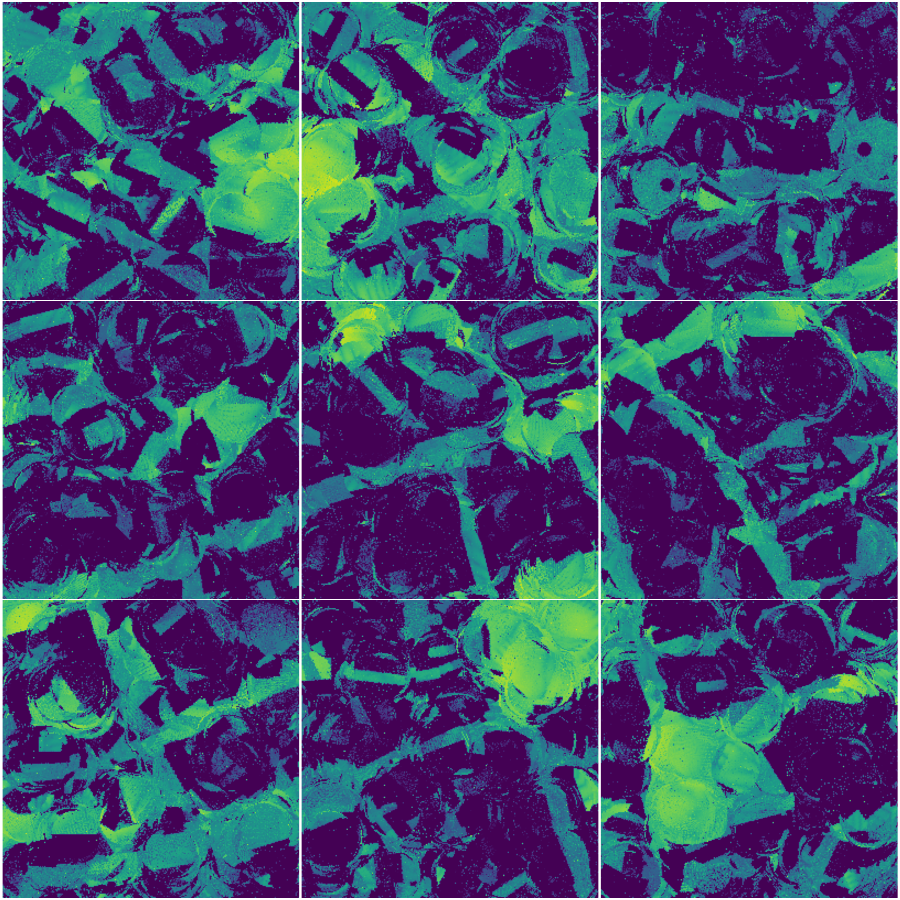}
\centering{(b)}
\end{minipage}\hfill
\begin{minipage}[t]{0.16\textwidth}
\label{fig:munich_c}
\includegraphics{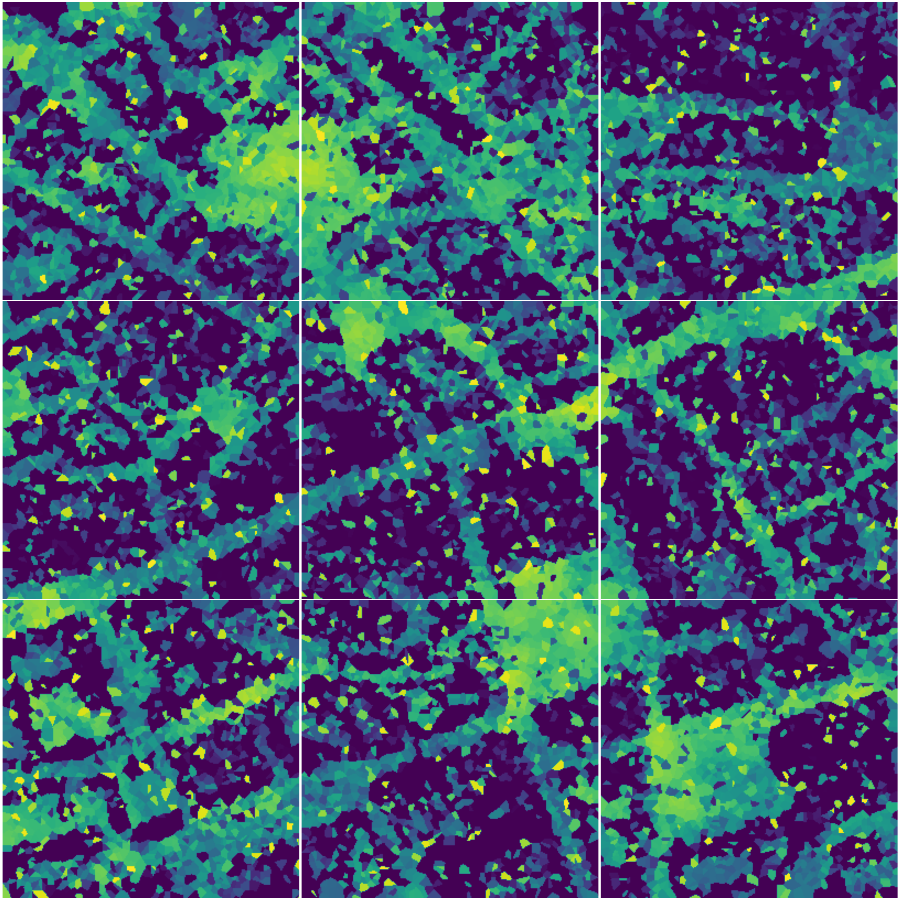}
\centering{(c)}
\end{minipage}\hfill
\begin{minipage}[t]{0.16\textwidth}
\label{fig:munich_d}
\includegraphics{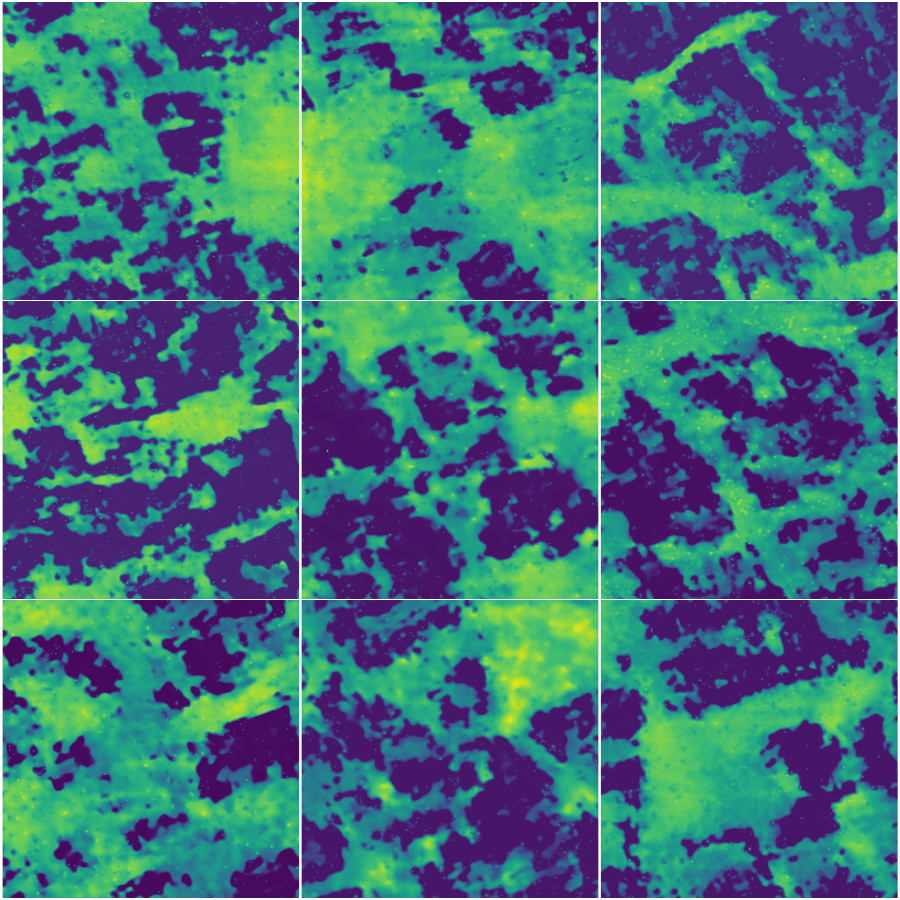}
\centering{(d)}
\end{minipage}
\begin{minipage}[t]{0.16\textwidth}
\label{fig:munich_e}
\includegraphics{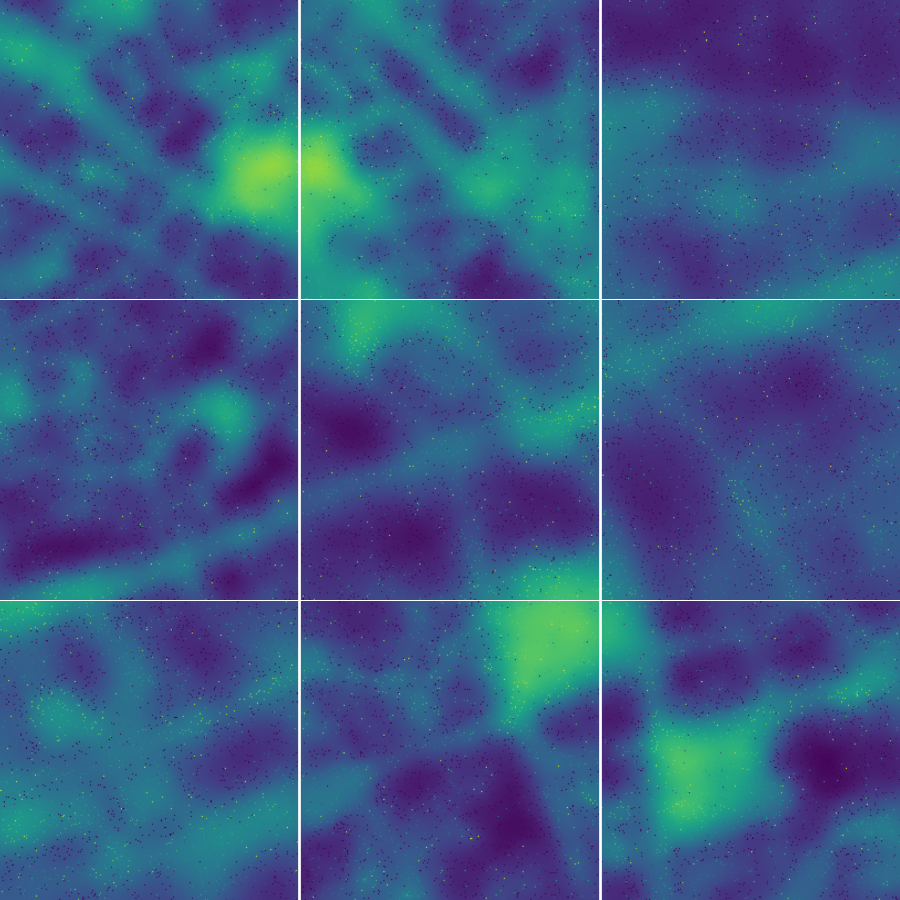}
\centering{(e)}
\end{minipage}
\begin{minipage}[t]{0.16\textwidth}
\label{fig:munich_f}
\includegraphics{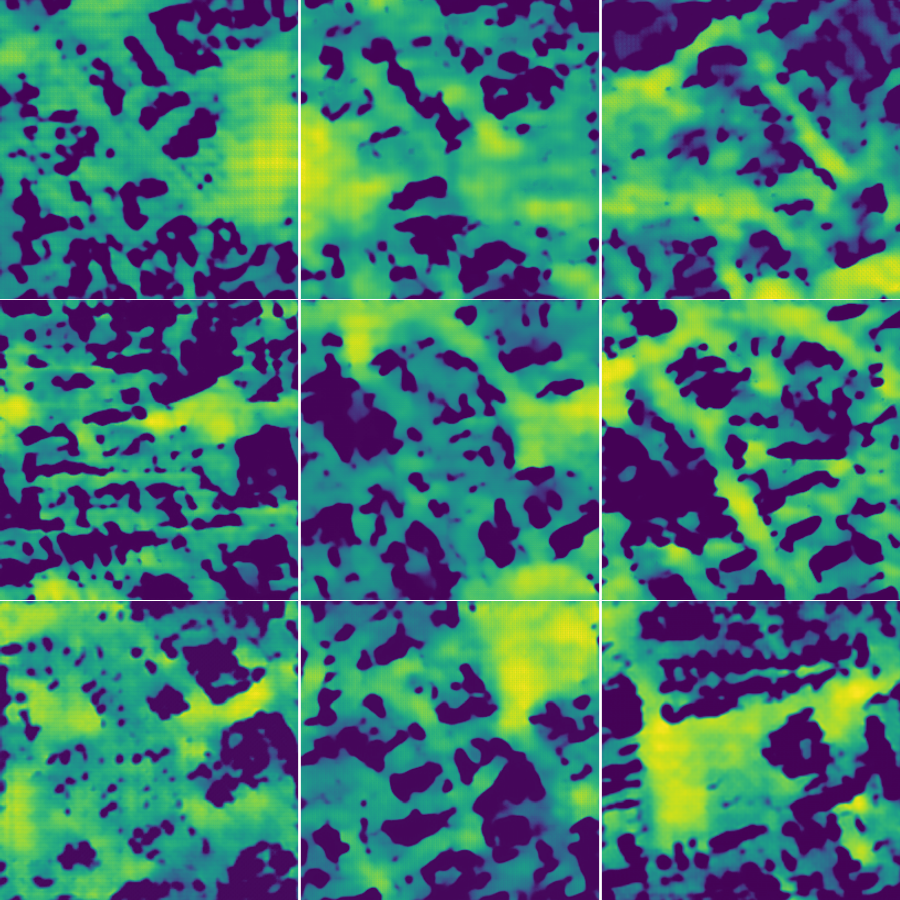}
\centering{(f)}
\end{minipage}
\caption{Comparison of reconstruction methods on the Munich scenario, with only $M=3\%$ of SINRs known, whose $Q=15\%$ of them is featuring errors. (a) Ground truth, (b) GENEO, (c) 1-KNN, (d) U-Net, \blue{(e) Kriging, and (f) CVAE.}}
\label{fig:munich_comparison}
\end{figure*}

\begin{figure*}
\setkeys{Gin}{width=1\linewidth}
\centering
\begin{minipage}[t]{0.16\textwidth}
\label{fig:paris_a}
\includegraphics{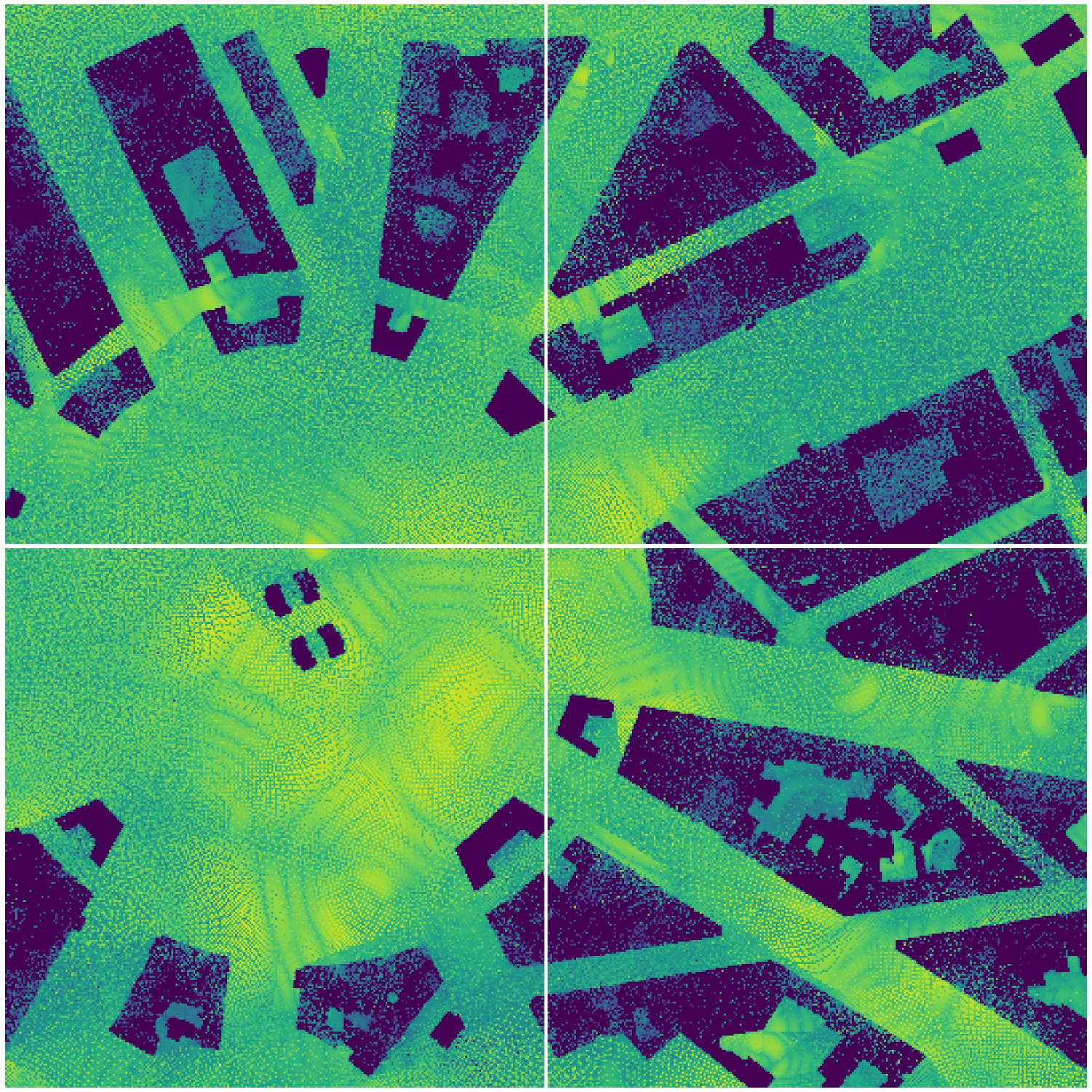}
\centering{(a)}
\end{minipage}\hfill
\begin{minipage}[t]{0.16\textwidth}
\label{fig:paris_b}
\includegraphics{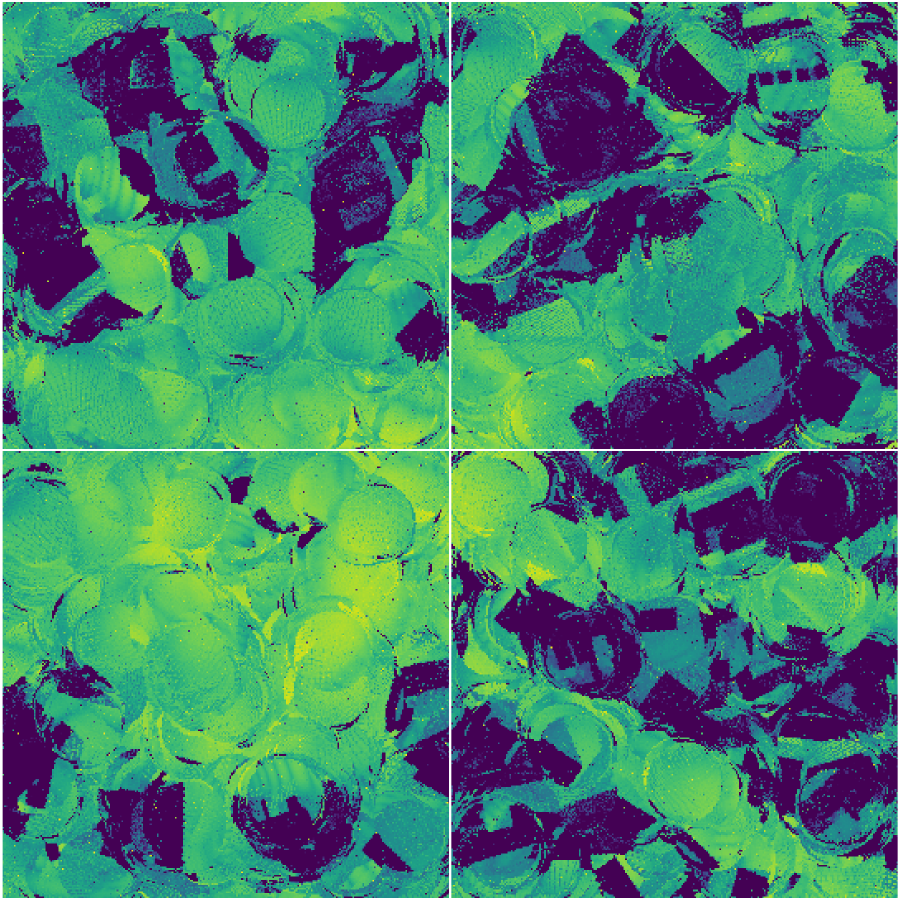}
\centering{(b)}
\end{minipage}\hfill
\begin{minipage}[t]{0.16\textwidth}
\label{fig:paris_c}
\includegraphics{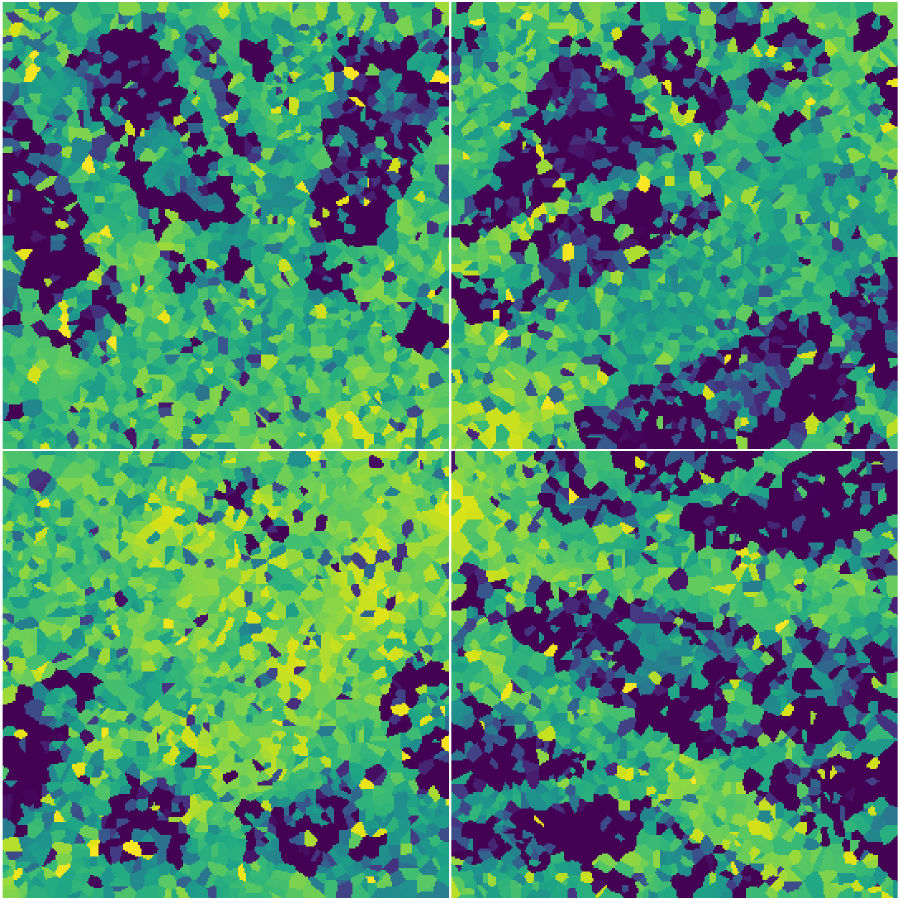}
\centering{(c)}
\end{minipage}\hfill
\begin{minipage}[t]{0.16\textwidth}
\label{fig:paris_d}
\includegraphics{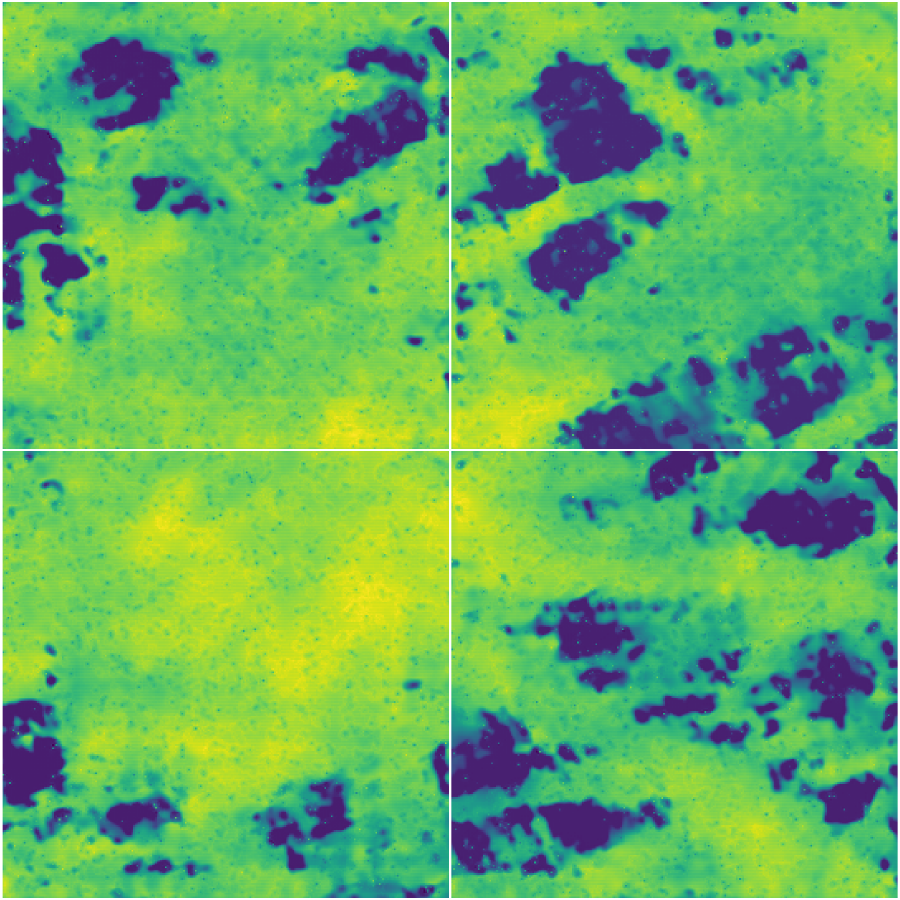}
\centering{(d)}
\end{minipage}
\begin{minipage}[t]{0.16\textwidth}
\label{fig:paris_e}
\includegraphics{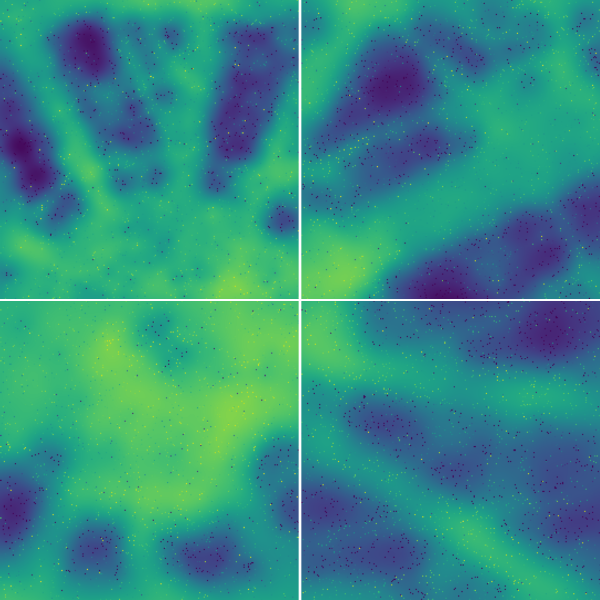}
\centering{(e)}
\end{minipage}
\begin{minipage}[t]{0.16\textwidth}
\label{fig:paris_f}
\includegraphics{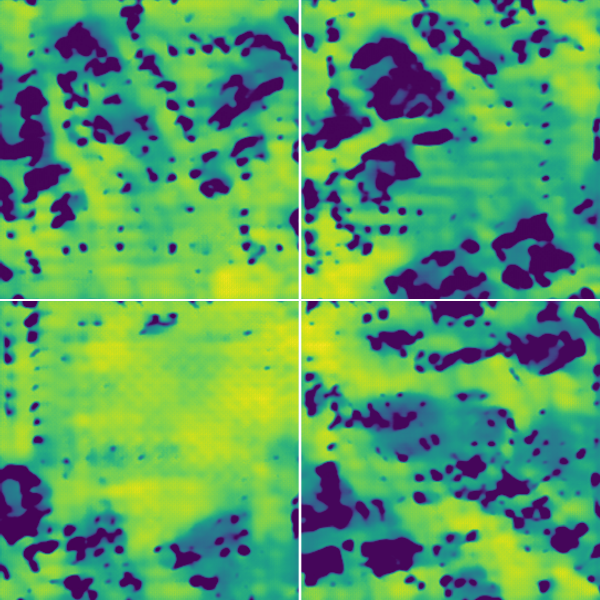}
\centering{(f)}
\end{minipage}
\caption{Comparison of reconstruction methods on the Paris scenario, with only $M=3\%$ of SINRs known, whose $Q=15\%$ of them is featuring errors. (a) Ground truth, (b) GENEO, (c) 1-KNN, (d) U-Net, \blue{(e) Kriging, and (f) CVAE.}}
\label{fig:paris_comparison}
\end{figure*}

\begin{figure}[t]
    \centering
    \includegraphics[width=.43\textwidth]{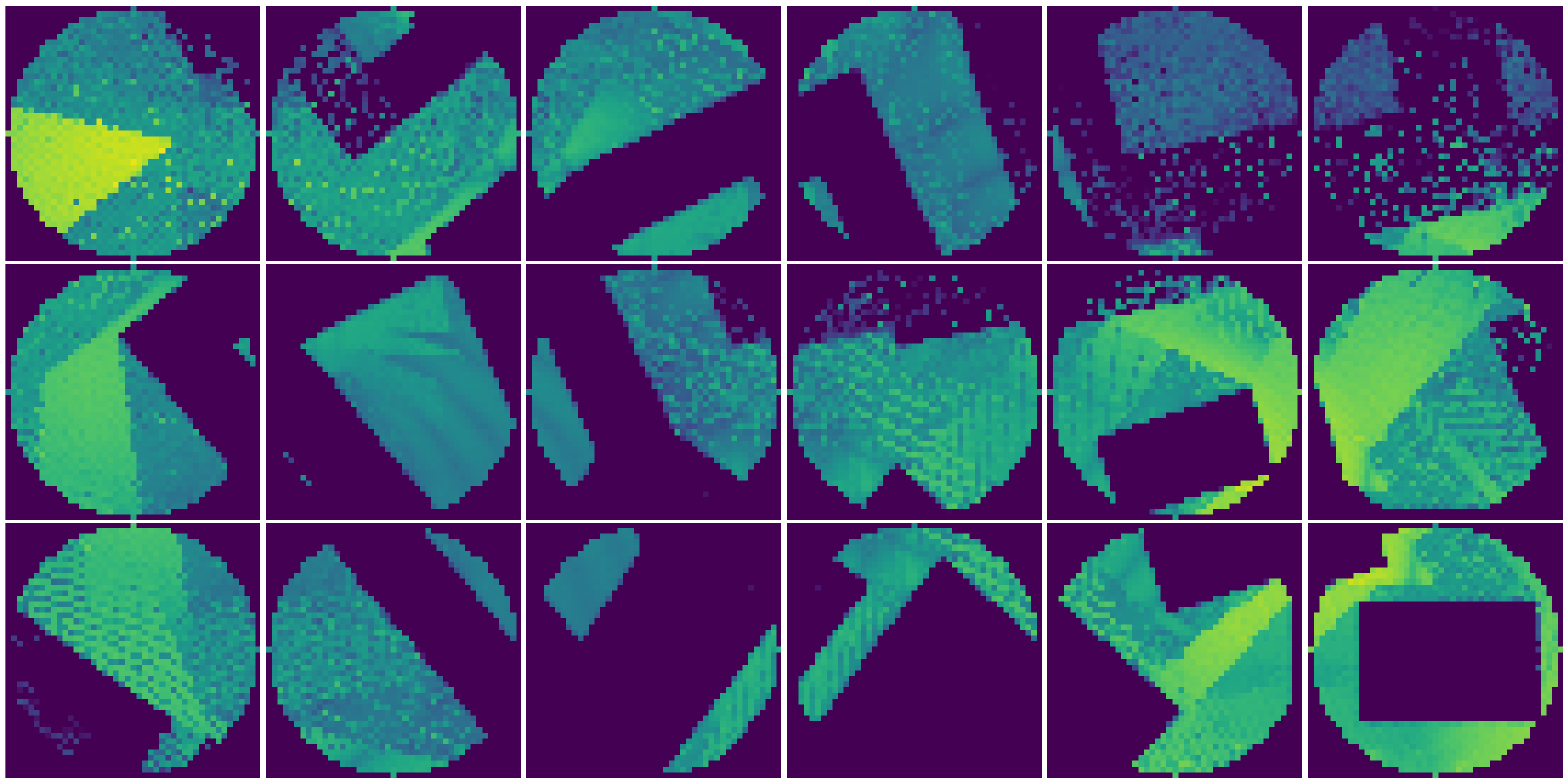}
    \caption{Example of patterns extracted from areas of Munich.}
    \label{fig:patterns}
\end{figure}

\subsection{SCENARIOS}
\label{sec:scenarios}

We generate target signals $\varphi$, samplings $\hat\varphi$, and reliability masks $\hat\psi$ from two outdoor ray‐tracing scenarios (Munich and Paris), each discretized into a set of $L\times L$ grids with $L=270$ (i.e., $270\times270$\,m\(^2\) per grid). For each scenario:
\begin{itemize}
  \item We sample only $M\%$ of the SINR values, selected uniformly at random, and mark those as “known.” 
  We define
  \begin{align}
  \begin{split}
     \hat\psi : &\{p_j\}_{j=1}^{L^2} \;\rightarrow\; \{0,1\}~, \\
    &\sum_{j}\hat\psi(p_j) = \frac{M}{100}\,L^2~, 
  \end{split}
  \end{align}
  where 
  \(\hat\psi(p_j)=1\) if the measurement \(\hat\varphi(p_j)\) is retained, and 0 otherwise.  We consider \(M\in\{1,2,3\}\)\blue{ and evaluate two distinct sampling schemes:
  \begin{itemize}
      \item \textbf{Uniform sampling.} The retained pixels are selected uniformly at random across the grid.
      \item \textbf{Non-uniform sampling.} We define a non-uniform sampling measure on the pixel grid via a softmax distribution:
      \begin{equation}
          \pi(p_j) = \frac{\exp(\varphi(p_j) / \tau)}{\sum_{k=1}^{L^2} \exp(\varphi(p_k) / \tau)} \,,
      \end{equation}
      where $\tau > 0$ is a temperature parameter controlling the sampling bias toward high-SINR regions. The sampled pixels are drawn without replacement from this distribution, with $\tau = 0.5$ in our experiments.
  \end{itemize}}
  
  \item To simulate noisy or corrupted measurements, we replace a further \(Q\%\) of those retained pixels with uniform random noise in \([0,1]\).  Concretely, out of the \(\tfrac{M}{100}\,L^2\) locations with \(\hat\psi=1\), we choose \(Q\%\) uniformly at random and for each such pixel \(\bar{p}_j\) set
  \begin{equation}
     \hat\varphi(\bar{p}_j) \;\leftarrow\; u \sim \mathcal{U}(0,1)~, 
  \end{equation}
  with \(Q\in\{15,30\}\).
\end{itemize}
This process yields incomplete and noisy observations of the true normalized SINR map, reflecting realistic measurement limitations. 
\blue{By coupling the two geographic layouts (Munich and Paris) with the two distinct sampling strategies (uniform and non-uniform), we effectively formulate four comprehensive evaluation scenarios with corruption of up to 30\% of the retained measurements. This ensures our models are tested against varying urban environments as well as varying spatial distributions of the available measurements.}
Although having substantially more data from multiple cities would be ideal, our experiments are based on the only two default urban scenarios provided by Sionna RT. While target signals $\varphi$ are fixed given a scenario, many different realizations of samplings $\hat{\varphi}$ and reliability masks $\hat{\psi}$ can be generated through the random selection process.

\subsubsection{Munich}
For the Munich scenario, we cover an overall area of \(810\times810\)\,m\(^2\), partitioned into nine non‐overlapping square subregions of \(270\times270\)\,m\(^2\) each (see Fig.~\ref{fig:munich_comparison}(a)). We evaluate GENEO and the baselines via leave-one-out cross‐validation over these nine areas: in each of nine trials, eight subregions provide the known/corrupted measurements (for pattern preparation and training, where applicable), while the ninth subregion's SINR values (masked by \(\hat\psi\)) form the reconstruction target \(\varphi\).

\subsubsection{Paris}
For the Paris scenario, we span \(540\times540\)\,m\(^2\), divided into four square subregions of \(270\times270\)\,m\(^2\) each (see Fig.~\ref{fig:paris_comparison}(a)). We assess generalization by training solely on all nine Munich subregions and then reconstructing each of the four Paris subregions in turn: the entire Munich collection provides known/corrupted measurements for GENEO and baseline training, and each Paris subregion's masked SINR values constitute the test targets. This experimental setup challenges the generalization capabilities of tested approaches, it requires training exclusively on data from one urban environment (Munich) and testing to a distinct and unseen city layout (Paris) without any retraining.

\subsection{GENEO IMPLEMENTATION}
\label{sec:GENEO_impl}

Each $L\times L$ SINR map $\varphi$ (as defined in Sec.~\ref{sec:signal_rec}) is partitioned using circular tiling \cyan{with centers arranged on a hexagonal lattice (the optimal packing for equal-radius circles \cite{fukshansky2011revisiting})}, covering the full $270\times270 \text{ m}^2$ area and keeping minimal overlap between circles. This process yields 67 patterns per area. In this paper, each $\chi_i$ is modelled as a fixed, rotation‐invariant mask with a support of diameter of 45 m:
\begin{equation}
    \chi(x,y) = 
    \begin{cases}
    1 & \text{if } x^2 + y^2 \le 22^2,\\
    0 & \text{otherwise}.
    \end{cases}
\end{equation}  
Each pattern is therefore a pair $P_{i} = \bigl(h_{i},\,\chi\bigr),\text{ with } i = 1, \ldots, 67$. \cyan{We emphasize that the geometry of $\chi$ leads to circular patterns.} To enrich the library, we generate 24 rotated variants of every pattern by using $15^\circ$ increments, such that pattern $h_{\tfrac{i\,  \theta }{15} +i}$ is equal to pattern $h_{i}$ rotated by $\theta$ degrees. \cyan{The number of rotations was chosen as a trade-off between computational cost and exact rotational equivariance.} Thus, each area contributes $67 \times 24 = 1608$ patterns. \cyan{The equivariance of the proposed GENEO with respect to translations and rotations ensures that the reconstruction depends on the local structure of the signal rather than on its absolute position or orientation in the map.} The construction of the pattern library differs between the two scenarios:

\begin{itemize}
  \item \textit{Munich:} In each leave‐one‐out trial, patterns are drawn from the eight ``training'' subregions (i.e., $8\times1608$ patterns) to set up the GENEO operator; the remaining subregion's masked SINR map, \blue{which is given only the known entries indicated by \(\hat\psi\)}, is reconstructed via GENEO operator. An example of patterns extracted is presented in Fig.~\ref{fig:patterns}.
  \item \textit{Paris:} All nine Munich subregions supply patterns (i.e., $9\times1608$ in total) for GENEO set up, and the four Paris subregions are subsequently reconstructed, one at a time, from their partial observations \(\hat\varphi\).
\end{itemize}
This setup ensures that GENEO leverages spatially equivariant operators from rich, minimally overlapping, local signal structures and applies them to reconstruct the full normalized SINR map \(\varphi\) under varying availability and noise conditions. \cyan{In what follows, we use the argmax reconstruction strategy for all reconstructions evaluated with respect to the topological metric, and softmax reconstruction strategy with $K=50$ for those evaluated using MSE.}

\subsection{BASELINES}
\label{sec:baselines}

\begin{figure*}[t]
    \centering
    \includegraphics[width=\textwidth]{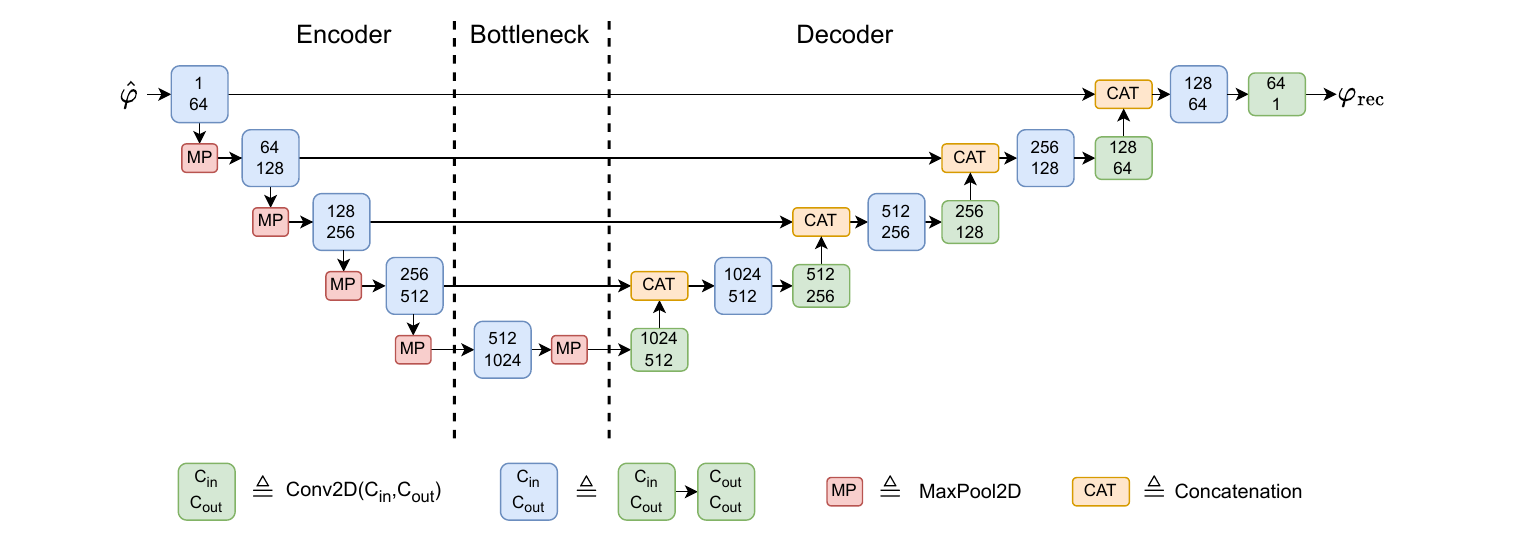}
    \caption{U-Net architecture.}
    \label{fig:unet}
\end{figure*}

\subsubsection{1-KNN}
As a simple, non-parametric reference, we implement a one-nearest-neighbor (1-KNN) interpolation over the spatial grid. The following steps break down the procedure: 

\begin{enumerate}
  \item Let \(\mathcal{K} = \{p_k\mid \hat\psi(p_k)=1\}\) be the set of pixels with available SINR measurements, and 
  \(\mathcal{P} = \{p_j\mid \hat\psi(p_j)=0\}\) the set of pixels to reconstruct.
  \item For any target \(p_j\in\mathcal{P}\), we compute the 2-norm to each known point \(p_k\in\mathcal{K}\)
  to identify the closest known pixel:
  \begin{equation}
    p^*
    = \underset{p_k\in\mathcal{K}}{\mathrm{argmin}}\ 
      \|p_j - p_k\|_2\,.
  \end{equation}
  \item Finally, we assign
  \begin{equation}
     \varphi_{\mathrm{rec}}(p_j)
    = \hat\varphi(p^*)\,, 
  \end{equation}
  i.e., we recover the SINR value of the nearest sampled pixel for reconstruction.
\end{enumerate}
This 1-KNN baseline exploits the spatial locality of SINR: each missing value is simply replaced by its nearest neighbor’s measurement in the two-dimensional plane, providing a simple yet effective benchmark.

\subsubsection{U-Net}
\blue{Convolutional neural networks \cite{ronneberger2015u} are advantageous for this task as they treat the reconstruction as an image-to-image regression problem. Specifically, the U-Net architecture can leverage fully supervised training to reconstruct missing information and correct noisy measurements \cite{radiounet}.}   
We employ a fully supervised U-Net model \cite{radiounet} to recover the complete SINR map $\varphi$ from its partially observed version $\hat\varphi$.
The network produces an output $\varphi_{\mathrm{rec}}\in\mathbb{R}^{L\times L}$, an estimate of the full reconstructed SINR.

Similarly to \cite{radiounet}, we consider a U-Net architecture comprising a four-level encoder, a central bottleneck, and a symmetric four-level decoder, connected via skip-connections. The considered U-Net architecture is presented in Fig.~\ref{fig:unet}. The 4-layers encoder features ReLU-activated $3\times3$ convolutional layers and $2\times2$ max-pooling, where the number of channels doubles at each depth. The bottleneck leverages two additional $3\times3$ convolutions with ReLU activation. The decoder successively performs upsampling by $2\times2$ transposed convolutions that halve the channel count. After each upsampling, the result is concatenated with the corresponding encoder feature map, and two $3\times3$ ReLU-activated convolutions restore spatial detail. A final $1\times1$ convolution produces the reconstructed map $\varphi_{\mathrm{rec}}$.

As supervised training loss, we train U-Net to minimize the mean squared error 
between $\varphi_{\mathrm{rec}}$ and $\varphi$ 
using the Adam optimizer with learning rate $10^{-4}$. The ground truth $\varphi$ is assumed fully available for supervision at training time, enabling U-Net to both reconstruct missing information and correct noisy measurements.

\blue{
\subsubsection{Kriging}
As a representative spatial interpolation benchmark, we implement Kriging, which is a geostatistical method that estimates unknown values based on the spatial covariance of the observed data \cite{kriging2018}. 
The procedure is formulated as follows:
\begin{enumerate}
    \item We extract the coordinates and corresponding measured \ac{SINR} values for all available points defined by the reliability mask, generating a set of known observations $\mathcal{K} = \{p_k \mid \hat\psi(p_k)=1\}$.
    \item We compute the empirical variogram from these observations and fit a theoretical variogram model to capture the spatial dependence of the \ac{SINR} field. 
    \item Using the fitted variogram, Kriging computes a weighted linear combination of the known values to predict the \ac{SINR} at every unmeasured coordinate $p_j \in \mathcal{P}$, generating the full reconstructed grid.
    \item Finally, to ensure strict consistency with the observations, the reconstructed map is explicitly forced to match the ground truth at the sampled locations: $\varphi_{\mathrm{rec}}(p_k) = \hat\varphi(p_k)$ for all $p_k \in \mathcal{K}$.
\end{enumerate}
}

\begin{figure*}
\setkeys{Gin}{width=1\linewidth}
\centering
\begin{minipage}[t]{0.49\textwidth}
\includegraphics{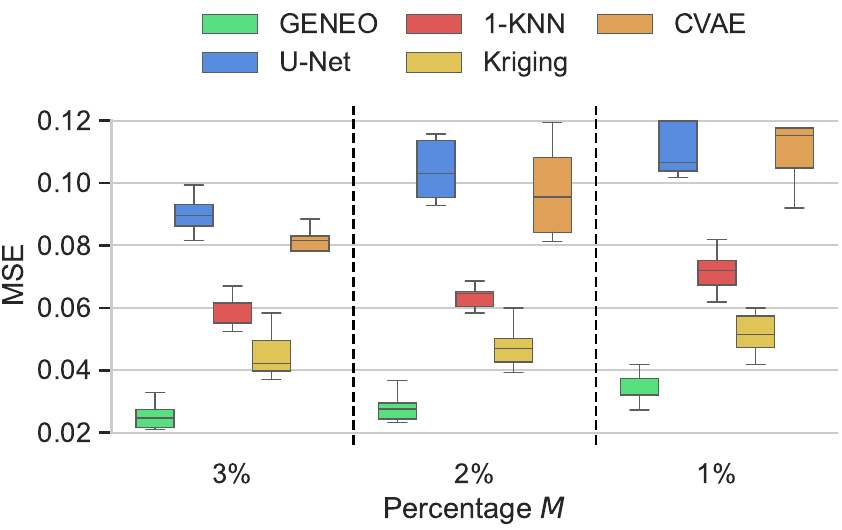}
\centering{(a)}
\end{minipage}\hfill
\begin{minipage}[t]{0.49\textwidth}
\includegraphics{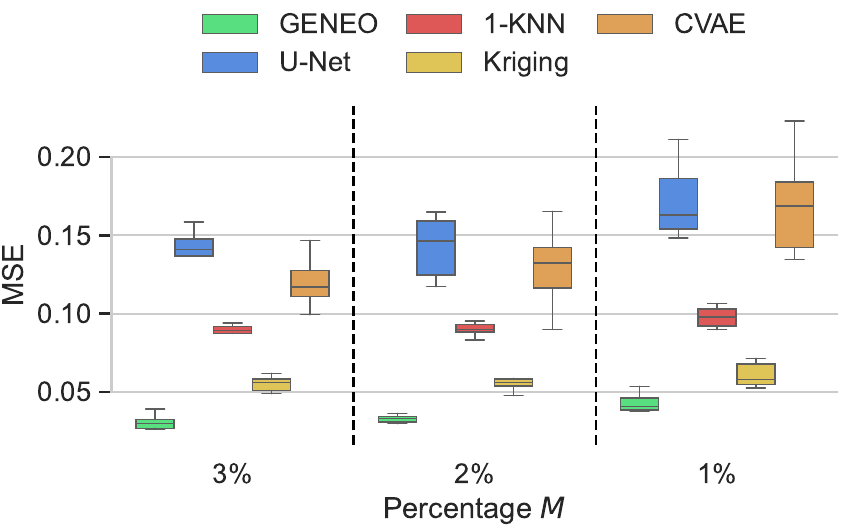}
\centering{(b)}
\end{minipage}
\caption{MSE achieved by GENEO, U-Net, 1-KNN, \blue{Kriging, and CVAE} for normalized SINR reconstruction in Munich scenario \blue{under uniform sampling}, where $M \in \{1,2,3\}$ with $Q=15$ in (a) and $Q=30$ in (b).}
\label{fig:munmse}
\end{figure*}

\begin{figure*}
\setkeys{Gin}{width=1\linewidth}
\centering
\begin{minipage}[t]{0.49\textwidth}
\includegraphics{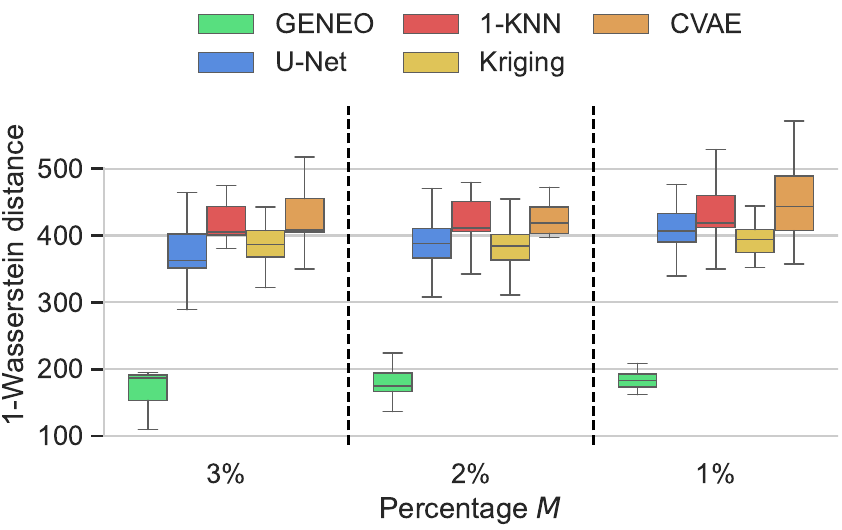}
\centering{(a)}
\end{minipage}\hfill
\begin{minipage}[t]{0.49\textwidth}
\includegraphics{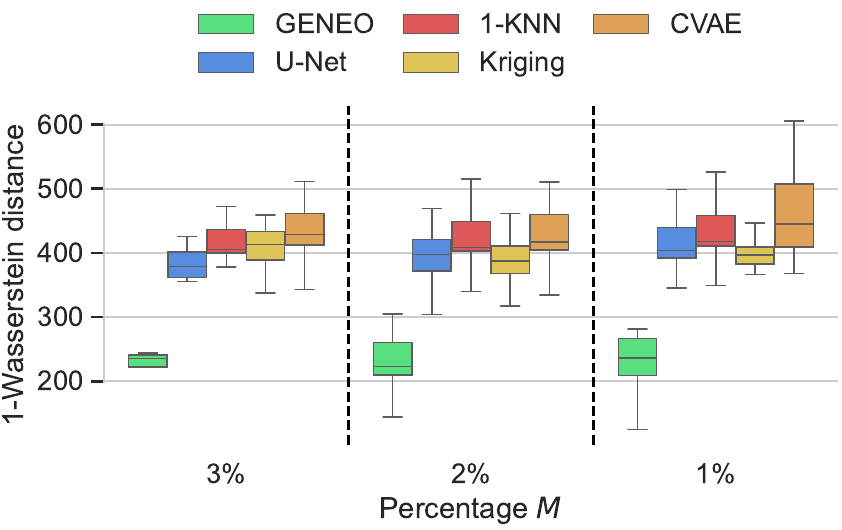}
\centering{(b)}
\end{minipage}
\caption{1-Wasserstein achieved by GENEO, U-Net, 1-KNN, \blue{Kriging, and CVAE} for normalized SINR reconstruction in Munich scenario \blue{under uniform sampling}, where $M \in \{1,2,3\}$ with $Q=15$ in (a) and $Q=30$ in (b).}
\label{fig:muntop}
\end{figure*}

\blue{
\subsubsection{\ac{CVAE}}
To benchmark against generative modeling \cite{cvaeReconstruction}, we introduce a \ac{CVAE}, which learns the conditional distribution of the complete \ac{SINR} maps given the sparse measurements.
The tested \ac{CVAE} architecture is composed of three main convolutional modules:
\begin{itemize}
    \item \textit{Posterior Encoder.} During training, this module takes the concatenation of the full ground truth map $\varphi$ and the sparse measurement map $\hat\varphi$. It processes them through four downsampling convolutional blocks (utilizing $3\times3$ kernels, batch normalization, ReLU activations, and $2\times2$ max-pooling) to output the parameters (mean $\mu$ and log-variance $\log\sigma^2$) of the latent space representation $z$.
    \item \textit{Condition Encoder.} This module processes only the sparse input map $\hat\varphi$ through a parallel four-block convolutional pathway to extract spatial condition features.
    \item \textit{Decoder.} The sampled latent vector $z$ (obtained via the reparameterization trick) is concatenated with the condition features. This joint representation is passed through a bottleneck and four transposed convolutional blocks ($2\times2$ stride) for upsampling. A final $1\times1$ convolution yields the reconstructed map $\varphi_{\mathrm{rec}}$.
\end{itemize}
The \ac{CVAE} is trained to minimize a combined loss function consisting of an $L^1$ reconstruction loss (to encourage pixel-wise fidelity) and a Kullback-Leibler (KL) divergence loss (to regularize the latent space). The model is optimized using the Adam optimizer with a learning rate of $10^{-3}$. During inference, since the ground truth $\varphi$ is unavailable, the latent vector $z$ is sampled directly from a standard normal prior $\mathcal{N}(0, I)$ and fed into the decoder alongside the condition features extracted from $\hat\varphi$ to generate the final prediction.
}

\subsection{RESULTS ON MUNICH SCENARIO \blue{WITH UNIFORM SAMPLING}}
\label{sec:results_munich}

We report leave-one-out cross-validation results on the Munich areas for both MSE and 1-Wasserstein distance, under two noise levels, with \(Q\in\{15,30\}\), and three sampling ratios, with \(M\in\{1,2,3\}\). Each plot summarizes nine trials, where in each trial eight subregions provide the known (and corrupted) SINR measurements and the remaining subregion is reconstructed. An example of reconstructions using the \blue{five} approaches is presented in Fig.~\ref{fig:munich_comparison}(b)-(d).

\paragraph{MSE Performance}  
Fig.~\ref{fig:munmse}(a) shows the MSE for \(Q=15\):
GENEO (green) consistently attains better performance \blue{across all sampling densities when compared to 1-KNN, U-Net, Kriging, and CVAE. Specifically, considering $Q=15$, GENEO achieves an MSE reduction ranging from 31\% ($M=1$) up to 43\% ($M=3$) against the best-performing baseline.} For the higher corruption level \(Q=30\), (Fig.~\ref{fig:munmse}(b)), GENEO again outperforms both all baselines, with its advantage being \blue{highly pronounced (between 29\% and 45\% improvement)}. U-Net and CVAE exhibit the largest errors in all settings.

\paragraph{1-Wasserstein Performance}  
Fig.~\ref{fig:muntop}(a) reports the 1-Wasserstein distance for \(Q=15\),
GENEO delivers dramatically lower topological discrepancy than \blue{the four baselines} across all \(M\), \blue{achieving a massive and stable improvement of approximately 54\% regardless of the sampling density.} At \(Q=30\) (Fig.~\ref{fig:muntop}(b)), GENEO maintains its superiority, achieving the smallest 1-Wasserstein distances in every case \blue{(improving from 41\% to 43\%)}. The baselines remain clustered at substantially higher values, showing minimal relative improvement as sampling density increases. 

\begin{figure*}
\setkeys{Gin}{width=1\linewidth}
\centering
\begin{minipage}[t]{0.49\textwidth}
\includegraphics{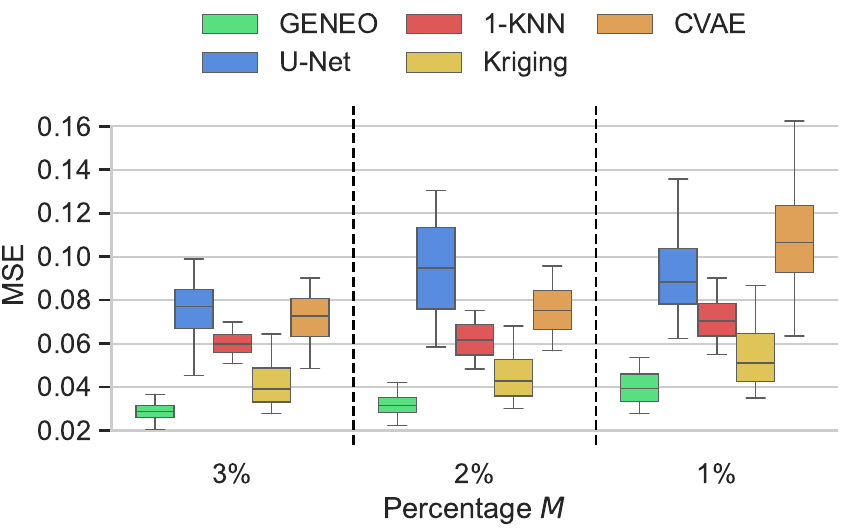}
\centering{(a)}
\end{minipage}\hfill
\begin{minipage}[t]{0.49\textwidth}
\includegraphics{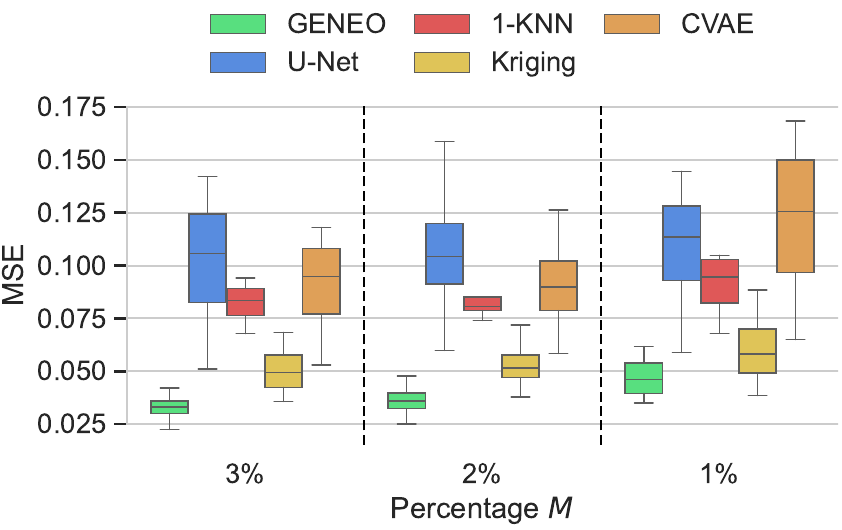}
\centering{(b)}
\end{minipage}
\caption{MSE achieved by GENEO, U-Net, 1-KNN, \blue{Kriging, and CVAE} for normalized SINR reconstruction in Paris scenario \blue{under uniform sampling}, where $M \in \{1,2,3\}$ with $Q=15$ in (a) and $Q=30$ in (b).}
\label{fig:parmse}
\end{figure*}

\begin{figure*}
\setkeys{Gin}{width=1\linewidth}
\centering
\begin{minipage}[t]{0.49\textwidth}
\includegraphics{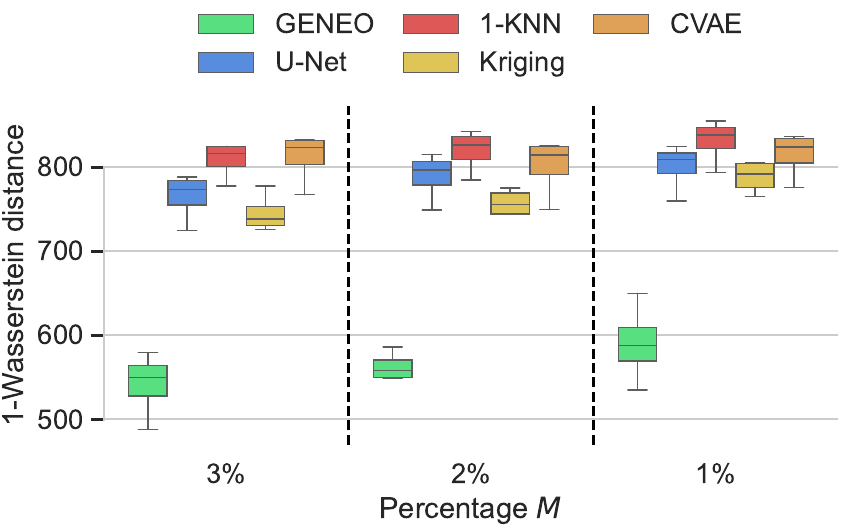}
\centering{(a)}
\end{minipage}\hfill
\begin{minipage}[t]{0.49\textwidth}
\includegraphics{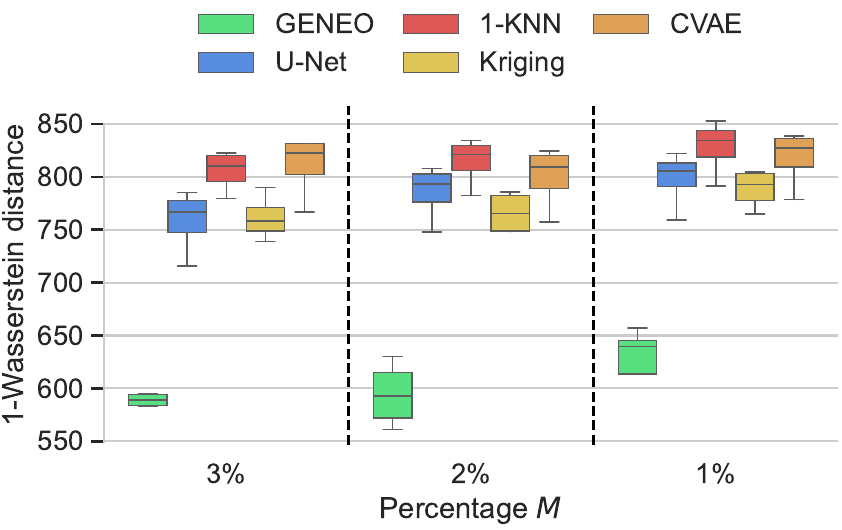}
\centering{(b)}
\end{minipage}
\caption{1-Wasserstein achieved by GENEO, U-Net, 1-KNN, \blue{Kriging, and CVAE} for normalized SINR reconstruction in Paris scenario \blue{under uniform sampling}, where $M \in \{1,2,3\}$ with $Q=15$ in (a) and $Q=30$ in (b).}
\label{fig:partop}
\end{figure*}

\subsection{RESULTS ON PARIS SCENARIO \blue{WITH UNIFORM SAMPLING}}
\label{sec:results_paris}

We evaluate zero‐shot generalization on the Paris areas using models trained exclusively on the nine Munich subregions, under two noise levels, with \(Q\in\{15,30\}\), and three sampling ratios, with \(M\in\{1,2,3\}\). Each boxplot aggregates four trials, one per Paris subregion. 
Results are collected for both MSE and 1‐Wasserstein distance. 
An example of reconstructions using the three approaches is presented in Fig.~\ref{fig:paris_comparison}(b)-(d).

\paragraph{MSE Performance}  
Fig.~\ref{fig:parmse}(a) (\(Q=15\)) reveals that GENEO (green) achieves the lowest reconstruction error in terms of MSE across all sampling ratios \(M\) \blue{when compared against 1-KNN, U-Net, Kriging, and CVAE. The generalization gap is well handled by our approach, which delivers an MSE improvement ranging from 28\% at $M=1$ to 32\% at $M=3$ over the best-performing baseline.}    
Under heavier corruption \(Q=30\) (Fig.~\ref{fig:parmse}(b)), GENEO's advantage persists: it consistently outperforms \blue{all baselines, securing an error reduction between 22\% ($M=1$) and 35\% ($M=3$). This confirms that the translation-equivariant filters learned in Munich scenario are highly transferable and robust to noise.}

\paragraph{1‐Wasserstein Performance}  
Fig.~\ref{fig:partop}(a) shows the 1‐Wasserstein distance for \(Q=15\):
GENEO again yields substantially lower topological error, \blue{improving upon the strongest baseline by a consistent margin of 25\% to 27\% across all values of $M$.} For \(Q=30\) (Fig.~\ref{fig:partop}(b)), GENEO maintains its lead in topological fidelity \blue{(yielding a 21\% to 22\% improvement)}, while \blue{all} baselines exhibit worse performance in terms of signal reconstruction. \blue{The ability to preserve spatial geometry in an unseen city under heavy noise strongly validates the structural bias embedded in GENEO.}

\begin{figure*}
\setkeys{Gin}{width=1\linewidth}
\centering
\begin{minipage}[t]{0.49\textwidth}
\includegraphics{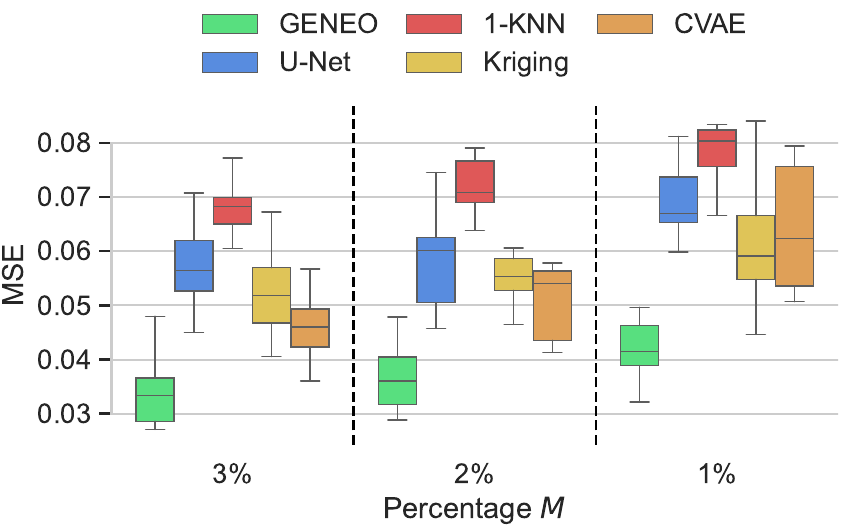}
\centering{(a)}
\end{minipage}\hfill
\begin{minipage}[t]{0.49\textwidth}
\includegraphics{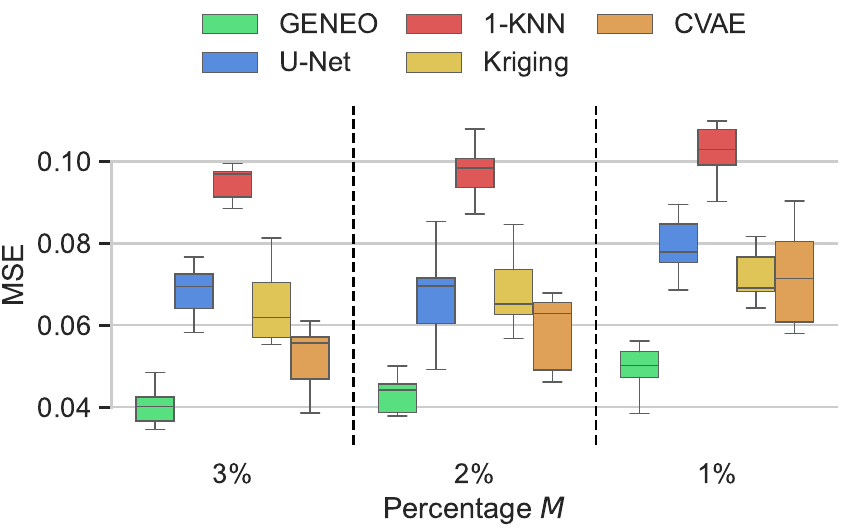}
\centering{(b)}
\end{minipage}
\caption{\blue{MSE achieved by GENEO, U-Net, 1-KNN, Kriging, and CVAE for normalized SINR reconstruction in Munich scenario under non-uniform sampling, where $M \in \{1,2,3\}$ with $Q=15$ in (a) and $Q=30$ in (b).}}
\label{fig:nonuniform_munmse}
\end{figure*}

\begin{figure*}
\setkeys{Gin}{width=1\linewidth}
\centering
\begin{minipage}[t]{0.49\textwidth}
\includegraphics{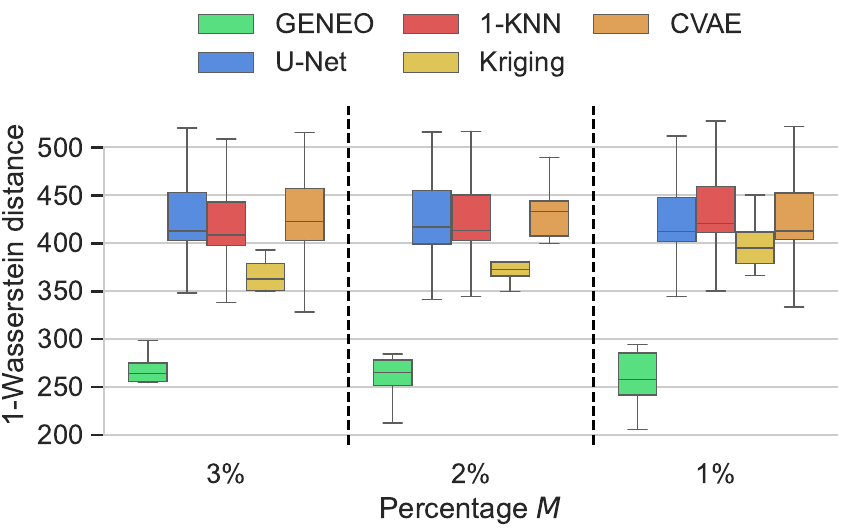}
\centering{(a)}
\end{minipage}\hfill
\begin{minipage}[t]{0.49\textwidth}
\includegraphics{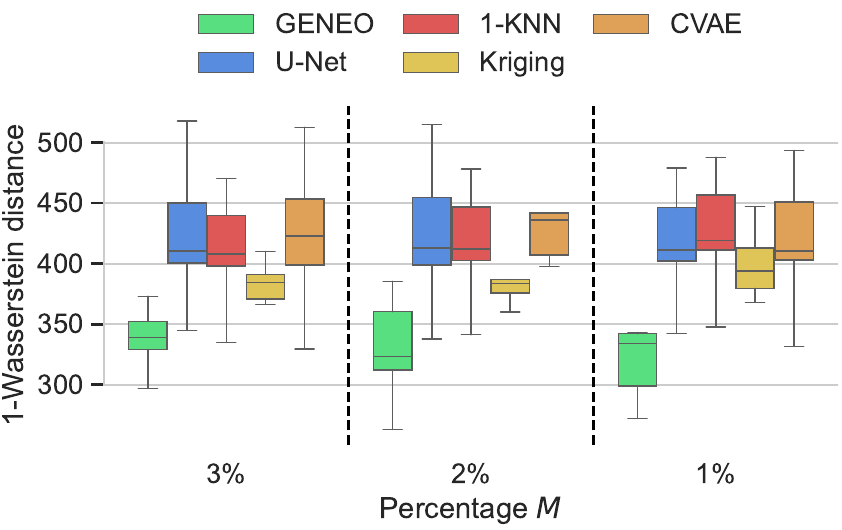}
\centering{(b)}
\end{minipage}
\caption{\blue{1-Wasserstein achieved by GENEO, U-Net, 1-KNN, Kriging, and CVAE for normalized SINR reconstruction in Munich scenario under non-uniform sampling, where $M \in \{1,2,3\}$ with $Q=15$ in (a) and $Q=30$ in (b).}}
\label{fig:nonuniform_muntop}
\end{figure*}

\subsection{\blue{RESULTS ON MUNICH SCENARIO WITH NON-UNIFORM SAMPLING}}
\label{sec:results_munich_nonuniform}

\blue{To evaluate robustness against spatial sampling bias, we test the models under the non-uniform sampling scheme. We maintain the same cross-validation setup, noise levels ($Q\in\{15,30\}$), and sampling ratios ($M\in\{1,2,3\}$).}

\paragraph{\blue{MSE Performance}}  
\blue{Fig.~\ref{fig:nonuniform_munmse}(a) shows the MSE for $Q=15$. Even with this alternatively sampling strategy, GENEO consistently outperforms all baselines. Interestingly, the relative improvement is most significant at extreme sparsity, yielding a 31\% reduction in MSE at $M=1$, which gradually lowers to 25\% at $M=3$ as the baselines receive more data. 
Under severe noise $Q=30$ (Fig.~\ref{fig:nonuniform_munmse}(b)), GENEO confirms its strong robustness, securing an MSE improvement between 20\% ($M=3$) and 30\% ($M=1$) over the best competing method.}

\paragraph{\blue{1-Wasserstein Performance}}  
\blue{Fig.~\ref{fig:nonuniform_muntop}(a) demonstrates that GENEO preserves topological fidelity far better than the baselines even when sampling is spatially biased. For $Q=15$, GENEO reduces the 1-Wasserstein distance by 33\% at $M=1$ and 27\% at $M=3$. 
When the corrupted samples reach $Q=30$ (Fig.~\ref{fig:nonuniform_muntop}(b)), the overall topological error increases for all models due to the combined challenge of high noise and biased sampling. Nevertheless, GENEO strictly maintains its lead, outperforming the best baseline by 19\% at $M=1$ and 10\% at $M=3$, proving that its geometric priors provide a structural advantage even in highly adverse conditions.}

\begin{figure*}
\setkeys{Gin}{width=1\linewidth}
\centering
\begin{minipage}[t]{0.49\textwidth}
\includegraphics{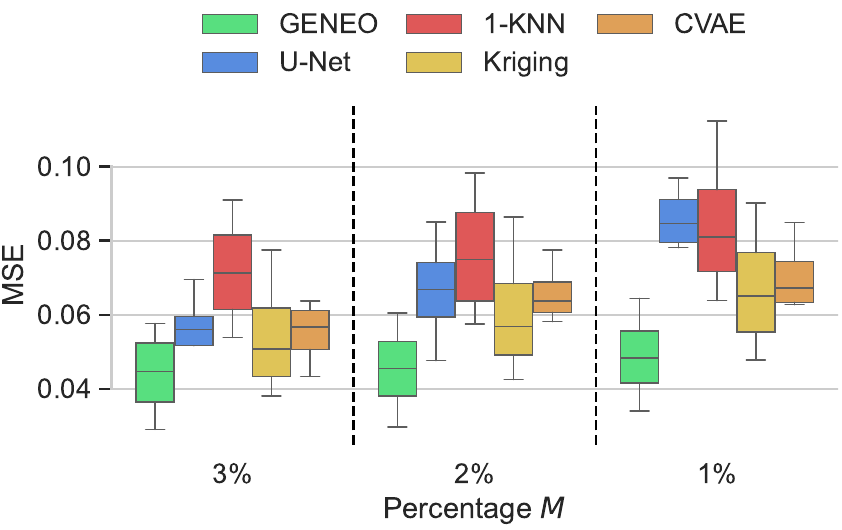}
\centering{(a)}
\end{minipage}\hfill
\begin{minipage}[t]{0.49\textwidth}
\includegraphics{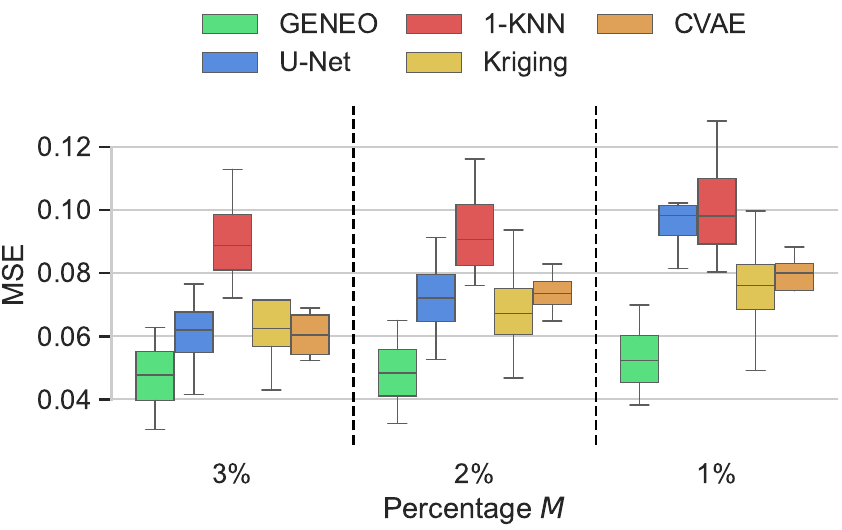}
\centering{(b)}
\end{minipage}
\caption{\blue{MSE achieved by GENEO, U-Net, 1-KNN, Kriging, and CVAE for normalized SINR reconstruction in Paris scenario under non-uniform sampling, where $M \in \{1,2,3\}$ with $Q=15$ in (a) and $Q=30$ in (b).}}
\label{fig:nonuniform_parmse}
\end{figure*}

\begin{figure*}
\setkeys{Gin}{width=1\linewidth}
\centering
\begin{minipage}[t]{0.49\textwidth}
\includegraphics{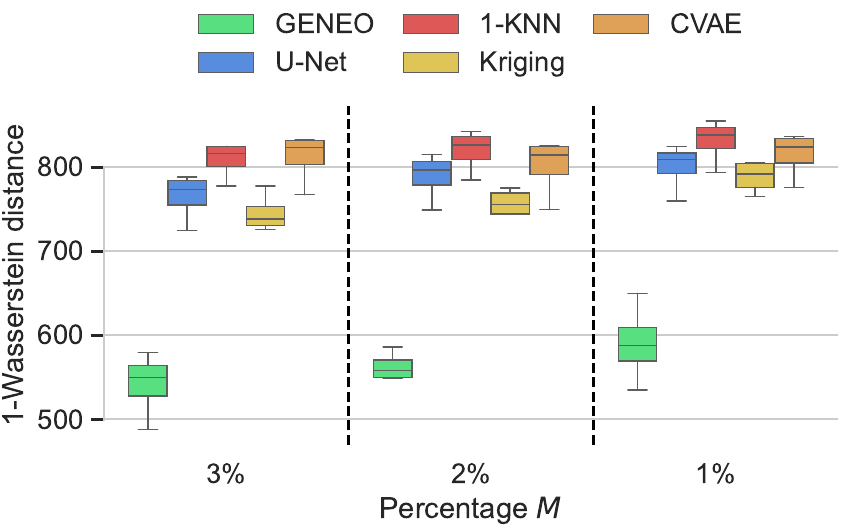}
\centering{(a)}
\end{minipage}\hfill
\begin{minipage}[t]{0.49\textwidth}
\includegraphics{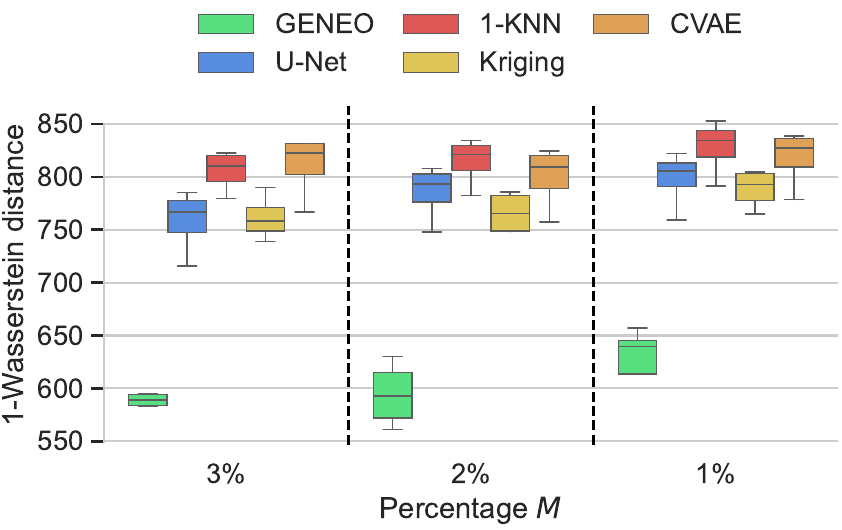}
\centering{(b)}
\end{minipage}
\caption{\blue{1-Wasserstein achieved by GENEO, U-Net, 1-KNN, Kriging, and CVAE for normalized SINR reconstruction in Paris scenario under non-uniform sampling, where $M \in \{1,2,3\}$ with $Q=15$ in (a) and $Q=30$ in (b).}}
\label{fig:nonuniform_partop}
\end{figure*}

\subsection{\blue{RESULTS ON PARIS SCENARIO WITH NON-UNIFORM SAMPLING}}
\label{sec:results_paris_nonuniform}

\blue{To thoroughly test the zero-shot generalization, we evaluate the Paris scenario under the non-uniform sampling scheme. Here, the models must reconstruct an unseen urban layout using measurements that are both highly sparse and spatially biased. We maintain the cross-validation setup, noise levels ($Q\in\{15,30\}$), and sampling ratios ($M\in\{1,2,3\}$).}

\paragraph{\blue{MSE Performance}}  
\blue{Fig.~\ref{fig:nonuniform_parmse}(a) reports the MSE for $Q=15$. In this settings, GENEO successfully generalizes, outperforming all baselines. Mirroring the results of the Munich scenario, the relative improvement is highest under extreme sparsity, achieving a 27\% reduction in MSE at $M=1$, and stabilizing at 18\% at $M=3$. 
Under the severe noise condition $Q=30$ (Fig.~\ref{fig:nonuniform_parmse}(b)), GENEO maintains excellent robustness, improving the MSE by up to 29\% ($M=1, 2$) and 21\% ($M=3$) against the most competitive baseline.}

\paragraph{\blue{1-Wasserstein Performance}}  
\blue{Fig.~\ref{fig:nonuniform_partop}(a) demonstrates GENEO's topological superiority under non-uniform sampling in an unseen domain. For $Q=15$, it decreases the 1-Wasserstein distance by 23\% at $M=1$, tapering to 20\% at $M=3$ as the data scarcity constraint slightly relaxes. 
When the measurement corruption is increased to $Q=30$ (Fig.~\ref{fig:nonuniform_partop}(b)), GENEO still consistently outperforms the established ML and spatial baselines, yielding topological improvements between 16\% ($M=3$) and 19\% ($M=1$). These zero-shot results clearly demonstrate that GENEO does not rely on a uniform sampling distribution or the specific geometry of the training city, but it generalizes and outperfoms all baselines consistently.}

\subsection{\blue{MODEL COMPLEXITY ANALYSIS AND DISCUSSION}}
\label{sec:model_complexity}

\begin{table*}[htbp]
\centering
\resizebox{\textwidth}{!}{
\begin{tabular}{llcc|cc|cc|cc|cc|cc}
\hline
\multirow{2}{*}{Scenario} & \multirow{2}{*}{Sampling} & \multirow{2}{*}{$M$(\%)} & \multirow{2}{*}{$Q$(\%)} & \multicolumn{2}{c|}{GENEO} & \multicolumn{2}{c|}{1-KNN} & \multicolumn{2}{c|}{U-Net} & \multicolumn{2}{c|}{Kriging} & \multicolumn{2}{c}{CVAE} \\
& & & & 1W & MSE ($\times 10^{2}$) & 1W & MSE ($\times 10^{2}$) & 1W & MSE ($\times 10^{2}$) & 1W & MSE ($\times 10^{2}$) & 1W & MSE ($\times 10^{2}$) \\
\hline
Munich & Uniform & 1 & 15 & $\mathbf{179 \pm 24}$ & $\mathbf{3.50 \pm 0.38}$ & $436 \pm 34$ & $7.15 \pm 0.39$ & $417 \pm 33$ & $12.14 \pm 1.88$ & $395 \pm 32$ & $5.11 \pm 0.42$ & $451 \pm 44$ & $12.35 \pm 2.16$ \\
Munich & Uniform & 1 & 30 & $\mathbf{226 \pm 35}$ & $\mathbf{4.30 \pm 0.38}$ & $435 \pm 34$ & $9.75 \pm 0.43$ & $419 \pm 33$ & $17.15 \pm 1.48$ & $398 \pm 32$ & $6.11 \pm 0.50$ & $463 \pm 49$ & $16.85 \pm 1.86$ \\
Munich & Uniform & 2 & 15 & $\mathbf{175 \pm 20}$ & $\mathbf{2.79 \pm 0.28}$ & $428 \pm 34$ & $6.39 \pm 0.23$ & $392 \pm 31$ & $11.01 \pm 1.57$ & $383 \pm 27$ & $4.70 \pm 0.41$ & $424 \pm 33$ & $9.83 \pm 0.91$ \\
Munich & Uniform & 2 & 30 & $\mathbf{226 \pm 31}$ & $\mathbf{3.37 \pm 0.28}$ & $425 \pm 34$ & $8.97 \pm 0.28$ & $395 \pm 32$ & $14.92 \pm 2.28$ & $392 \pm 27$ & $5.59 \pm 0.34$ & $429 \pm 33$ & $12.90 \pm 1.40$ \\
Munich & Uniform & 3 & 15 & $\mathbf{172 \pm 18}$ & $\mathbf{2.52 \pm 0.26}$ & $420 \pm 34$ & $5.91 \pm 0.32$ & $376 \pm 33$ & $8.88 \pm 0.49$ & $386 \pm 23$ & $4.45 \pm 0.48$ & $431 \pm 34$ & $8.11 \pm 0.54$ \\
Munich & Uniform & 3 & 30 & $\mathbf{223 \pm 29}$ & $\mathbf{3.02 \pm 0.27}$ & $417 \pm 33$ & $8.76 \pm 0.39$ & $383 \pm 33$ & $14.20 \pm 1.13$ & $407 \pm 25$ & $5.51 \pm 0.29$ & $436 \pm 33$ & $11.99 \pm 1.05$ \\
\hline
Munich & Non-unif. & 1 & 15 & $\mathbf{267 \pm 30}$ & $\mathbf{4.21 \pm 0.36}$ & $436 \pm 34$ & $8.00 \pm 0.58$ & $426 \pm 33$ & $7.05 \pm 0.64$ & $401 \pm 32$ & $6.12 \pm 0.77$ & $430 \pm 37$ & $6.36 \pm 0.74$ \\
Munich & Non-unif. & 1 & 30 & $\mathbf{322 \pm 37}$ & $\mathbf{4.98 \pm 0.37}$ & $435 \pm 34$ & $10.38 \pm 0.68$ & $425 \pm 33$ & $8.02 \pm 0.63$ & $400 \pm 33$ & $7.24 \pm 0.70$ & $429 \pm 37$ & $7.11 \pm 0.80$ \\
Munich & Non-unif. & 2 & 15 & $\mathbf{266 \pm 20}$ & $\mathbf{3.62 \pm 0.40}$ & $428 \pm 33$ & $7.20 \pm 0.35$ & $427 \pm 33$ & $5.84 \pm 0.58$ & $375 \pm 29$ & $5.55 \pm 0.56$ & $430 \pm 33$ & $5.07 \pm 0.45$ \\
Munich & Non-unif. & 2 & 30 & $\mathbf{328 \pm 26}$ & $\mathbf{4.42 \pm 0.40}$ & $426 \pm 33$ & $9.81 \pm 0.41$ & $425 \pm 34$ & $6.74 \pm 0.73$ & $383 \pm 28$ & $6.77 \pm 0.55$ & $430 \pm 34$ & $5.89 \pm 0.57$ \\
Munich & Non-unif. & 3 & 15 & $\mathbf{267 \pm 20}$ & $\mathbf{3.38 \pm 0.44}$ & $421 \pm 33$ & $6.73 \pm 0.39$ & $429 \pm 34$ & $5.67 \pm 0.52$ & $366 \pm 26$ & $5.20 \pm 0.55$ & $429 \pm 37$ & $4.53 \pm 0.44$ \\
Munich & Non-unif. & 3 & 30 & $\mathbf{340 \pm 24}$ & $\mathbf{4.17 \pm 0.49}$ & $420 \pm 32$ & $9.33 \pm 0.46$ & $426 \pm 34$ & $6.71 \pm 0.54$ & $381 \pm 26$ & $6.43 \pm 0.56$ & $427 \pm 36$ & $5.22 \pm 0.52$ \\
\hline
Paris & Uniform & 1 & 15 & $\mathbf{590 \pm 46}$ & $\mathbf{4.00 \pm 1.09}$ & $831 \pm 26$ & $7.15 \pm 1.45$ & $801 \pm 28$ & $9.37 \pm 3.02$ & $789 \pm 19$ & $5.59 \pm 2.19$ & $815 \pm 27$ & $10.98 \pm 3.98$ \\
Paris & Uniform & 1 & 30 & $\mathbf{620 \pm 51}$ & $\mathbf{4.73 \pm 1.15}$ & $828 \pm 26$ & $9.05 \pm 1.66$ & $798 \pm 27$ & $10.76 \pm 3.57$ & $789 \pm 19$ & $6.09 \pm 2.07$ & $818 \pm 27$ & $12.11 \pm 4.42$ \\
Paris & Uniform & 2 & 15 & $\mathbf{563 \pm 17}$ & $\mathbf{3.19 \pm 0.81}$ & $820 \pm 25$ & $6.18 \pm 1.14$ & $789 \pm 28$ & $9.46 \pm 3.06$ & $758 \pm 16$ & $4.59 \pm 1.61$ & $801 \pm 35$ & $7.58 \pm 1.62$ \\
Paris & Uniform & 2 & 30 & $\mathbf{594 \pm 31}$ & $\mathbf{3.61 \pm 0.91}$ & $815 \pm 23$ & $8.33 \pm 1.01$ & $786 \pm 26$ & $10.68 \pm 3.97$ & $766 \pm 20$ & $5.32 \pm 1.38$ & $800 \pm 30$ & $9.11 \pm 2.75$ \\
Paris & Uniform & 3 & 15 & $\mathbf{542 \pm 38}$ & $\mathbf{2.87 \pm 0.65}$ & $809 \pm 22$ & $6.01 \pm 0.79$ & $765 \pm 28$ & $7.47 \pm 2.18$ & $745 \pm 23$ & $4.26 \pm 1.56$ & $812 \pm 30$ & $7.11 \pm 1.72$ \\
Paris & Uniform & 3 & 30 & $\mathbf{589 \pm 6}$ & $\mathbf{3.27 \pm 0.78}$ & $806 \pm 19$ & $8.22 \pm 1.11$ & $759 \pm 30$ & $10.12 \pm 3.81$ & $761 \pm 21$ & $5.07 \pm 1.37$ & $811 \pm 30$ & $9.02 \pm 2.77$ \\
\hline
Paris & Non-unif. & 1 & 15 & $\mathbf{615 \pm 37}$ & $\mathbf{4.89 \pm 1.26}$ & $831 \pm 25$ & $8.45 \pm 2.05$ & $819 \pm 30$ & $8.61 \pm 0.85$ & $802 \pm 23$ & $6.71 \pm 1.80$ & $819 \pm 30$ & $7.06 \pm 1.00$ \\
Paris & Non-unif. & 1 & 30 & $\mathbf{642 \pm 37}$ & $\mathbf{5.32 \pm 1.32}$ & $828 \pm 26$ & $10.12 \pm 2.00$ & $820 \pm 29$ & $9.50 \pm 0.94$ & $796 \pm 23$ & $7.52 \pm 2.02$ & $820 \pm 29$ & $7.75 \pm 1.11$ \\
Paris & Non-unif. & 2 & 15 & $\mathbf{600 \pm 11}$ & $\mathbf{4.54 \pm 1.29}$ & $820 \pm 24$ & $7.65 \pm 1.79$ & $815 \pm 28$ & $6.66 \pm 1.52$ & $769 \pm 20$ & $6.07 \pm 1.86$ & $823 \pm 28$ & $6.58 \pm 0.82$ \\
Paris & Non-unif. & 2 & 30 & $\mathbf{622 \pm 17}$ & $\mathbf{4.86 \pm 1.35}$ & $818 \pm 24$ & $9.34 \pm 1.71$ & $814 \pm 30$ & $7.21 \pm 1.57$ & $770 \pm 23$ & $6.87 \pm 1.89$ & $823 \pm 28$ & $7.37 \pm 0.74$ \\
Paris & Non-unif. & 3 & 15 & $\mathbf{594 \pm 29}$ & $\mathbf{4.41 \pm 1.24}$ & $812 \pm 23$ & $7.19 \pm 1.59$ & $819 \pm 26$ & $5.52 \pm 1.23$ & $746 \pm 19$ & $5.44 \pm 1.69$ & $808 \pm 45$ & $5.52 \pm 0.88$ \\
Paris & Non-unif. & 3 & 30 & $\mathbf{630 \pm 22}$ & $\mathbf{4.72 \pm 1.35}$ & $812 \pm 20$ & $9.07 \pm 1.69$ & $816 \pm 26$ & $6.06 \pm 1.42$ & $756 \pm 27$ & $6.56 \pm 2.09$ & $812 \pm 45$ & $6.05 \pm 0.80$ \\
\hline
\end{tabular}
}
\caption{\blue{Performance of GENEO versus baselines (1-Wasserstein distance (1W) and MSE) on Munich and Paris scenarios across varying sampling strategies, densities, and noise levels. Results are reported as the mean $\pm$ 95\% confidence interval.}}
\label{tab:metrics_summary}
\end{table*}

A key advantage of the GENEO framework is its ability to naturally incorporate geometric priors, which drastically reduces the dimensionality of the optimization space compared to purely data-driven deep learning models. Moreover, deep learning approaches tend to require large training sets and may produce unrealistic artifacts when data is very limited \cite{Basheer2025}. Here, we quantitatively evaluate the model complexity of the tested approaches.
The considered supervised U-Net baseline implemented in our evaluation consists of an extensive encoder-bottleneck-decoder structure comprising approximately $3.1~\times~10^7$ parameters. Similarly, the generative \ac{CVAE} framework, which requires a dual-encoder architecture to map both the posterior and condition distributions alongside a deep decoder, utilizes approximately $4.2~\times~10^7$ trainable parameters.

In contrast, the proposed GENEO requires only $1.2~\times~10^4$ parameters (i.e., patterns) to model the Munich scenario and $1.4~\times~10^4$ parameters for the Paris scenario. This represents a parameter reduction of three orders of magnitude. While traditional spatial interpolation techniques such as 1-KNN and Kriging are inherently parameter-free, as demonstrated in the preceding performance evaluations (see Table~\ref{tab:metrics_summary}), they exhibit worse performance than GENEO. Indeed, GENEO delivers the best perfomance while achieving high computational compactness and parameter efficiency.

\bluee{To provide a comprehensive evaluation of computational complexity, Table~\ref{tab:flops} details the floating-point operations (FLOPs) required by each method during both training and inference in Munich scenario. Deep learning baselines demand massive computational resources to train, whereas GENEO is entirely training-free, while being competitive at inference.

\begin{table}[h]
\centering
\caption{Number of parameters and computational cost (FLOPs) of the five reconstruction methods.}
\label{tab:flops}
\begin{tabular}{lccc}
\toprule
Method & Parameters & Inference FLOPs & Training FLOPs \\
\midrule
GENEO   & $1.2\times10^4$            & $8.9\times10^{11}$        & ---               \\
U-Net   & $3.1\times10^7$            & $1.1\times10^{11}$        & $3.9\times10^{15}$ \\
CVAE    & $4.2\times10^7$           & $9.2\times10^{10}$        & $4.5\times10^{15}$ \\
Kriging & ---            & $1.4\times10^{12}$        & ---               \\
1-KNN   & ---            & $1.9\times10^{7}$         & ---               \\
\bottomrule
\end{tabular}
\end{table}
}

\section{CONCLUSION}
\label{sec:conclusion}

In this work, we introduced a novel framework for reconstructing \ac{SINR} maps in urban mobile radio scenarios from extremely sparse measurements, grounded in the theory of GENEOs. First, we develop a GENEO-driven reconstruction method that embeds domain symmetries directly into the operator, yielding compact models with strong inductive biases. Second, we couple traditional statistical metrics (MSE) with a TDA metric (1-Wasserstein distance) to capture not only intensity errors but also the preservation of geometric and spatial signal characteristics. Third, we deliver a thorough empirical evaluation in realistic Munich and Paris scenarios generated via the Sionna RT ray-tracing tool. Our results demonstrate that our approach consistently outperforms state-of-the-art \ac{ML} and spatial interpolation baselines across varying sparsity levels (1\% to 3\%), noise conditions (15\% to 30\% error), and sampling strategies (uniform and non-uniform). Specifically, GENEO reduces statistical error (MSE) by up to 45\% in the Munich scenario and up to 35\% in the Paris generalization scenario. More importantly, it yields massive gains in topological accuracy, decreasing the 1-Wasserstein distance up to 54\% in Munich and up to 27\% in Paris. Notably, the framework exhibits exceptional robustness, maintaining these strong performance margins even under the most severe constraints (1\% sampling rate with 30\% sampling error).

Future work will focus on introducing exact rotational equivariance instead of the current approximate formulation, improving reliability masks and pattern libraries, and addressing dynamic environments to better capture real-world scenarios. \bluee{Specifically, reducing and optimizing the pattern library represents a key avenue to significantly improve the computational efficiency during inference.}

\section*{APPENDIX A}

Here we provide the proof of the theoretical results described in the previous sections. First we present the proof of Proposition~\ref{prophatC}:
\begin{proof}[Proof of Proposition~\ref{prophatC}]
Given a signal $S = (f, \psi_f)$ and a pattern $P = (h, \chi_h)$, we know that $\mathcal{A}_{S,P}$ and $\mathcal{S}_{S,P}$ are nonnegative since the integrating functions are nonnegative.
In order to conclude the proof, it will suffice to show that $\mathcal{S}_{S,P} \le \mathcal{A}_{S,P} \le 1$. We recall that in our model we assume that $\psi_f \le 1$, $f \le 1$, and $\int_{\mathbb{R}^2} \chi_h \ d\mu \le 1$. Since $ \bar f (\xi,\eta) :=\lvert {f}(x+\xi,y+\eta) - h(\xi,\eta) \rvert \le 1$, we have that
\begin{align*}
    \mathcal{S}_{S,P}& = \int_{\mathbb{R}^2}\bar{f}(\xi,\eta) \psi(x+\xi,y+\eta) \chi_h (\xi,\eta) \ d \mu(\xi,\eta)
    \\& \le \int_{\mathbb{R}^2} \psi_f(x+\xi,y+\eta) \chi_h \ d\mu \\
    &= \mathcal{A}_{S,P}(x,y)
    \\& \le \int_{\mathbb{R}^2} \chi_h \ d\mu \le 1.
\end{align*}
\end{proof}
\noindent Then, we can prove Proposition~\ref{propA}:
\begin{proof}[Proof of Proposition~\ref{propA}]
Fixing a point $(x,y) \in \mathbb{R}^2$, we have that
\begin{align*}
\mathcal{A}_{S_1,P}(x,y) & =\int_{\mathbb{R}^2} \psi_{f_1}(x+\xi,y+\eta)\chi_h(\xi,\eta)\ d\xi\ d\eta 
\\&= \int_{\mathbb{R}^2} \psi_{f_2}(x+\xi,y+\eta)\chi_h(\xi,\eta)\ d\xi\ d\eta 
\\&=\mathcal{A}_{S_2,P}(x,y).
\end{align*}
Before proceeding, we set $\mathcal{A}:=\mathcal{A}_{S_1,P}=\mathcal{A}_{S_2,P}$. Denoting $ \bar f_i (\xi,\eta) :=\lvert {f}_i(x+\xi,y+\eta) - h(\xi,\eta) \rvert$ for $i =1,2$, using the reverse triangle inequality we obtain that

\begin{align*}
\lvert \bar f_1 (\xi,\eta) - \bar f_2 (\xi,\eta) \rvert & =
\Big \lvert \lvert {f}_1(x+\xi,y+\eta) - h(\xi,\eta) \rvert 
\\&\quad - \lvert {f}_2(x+\xi,y+\eta) - h(\xi,\eta) \rvert \Big\rvert
\\& \le \lvert f_1(x+\xi,y+\eta) - f_2(x+\xi,y+\eta) \rvert
\\& \le \lVert f_1 - f_2 \rVert_\infty.
\end{align*}
Then, setting $\psi:\equiv\psi_{f_1}\equiv\psi_{f_2}$, we get 
\begin{align*}
    & \lvert \hat c_{S_1,P}(x,y) - \hat c_{S_2,P}(x,y) \rvert = \lvert \mathcal{S}_{S_1,P}(x,y) - \mathcal{S}_{S_2,P}(x,y) \rvert
    \\ & = \bigg\lvert \int_{\mathbb{R}^2} \bar{f}_1(\xi,\eta) \psi(x+\xi,y+\eta) \chi_h (\xi,\eta) \ d\mu (\xi,\eta) \\ 
    &\quad -  \int_{\mathbb{R}^2}\bar{f}_2(\xi,\eta) \psi(x+\xi,y+\eta) \chi_h (\xi,\eta) \ d\mu(\xi,\eta) \bigg\rvert \\
    & \le \int_{\mathbb{R}^2}\left\lvert \bar{f}_1(\xi,\eta) - \bar{f}_2(\xi,\eta)\right\rvert \psi(x+\xi,y+\eta) \chi_h (\xi,\eta) \ d\mu(\xi,\eta)\\
    &\le \int_{\mathbb{R}^2}\lVert f_1 - f_2 \rVert_\infty \psi(x+\xi,y+\eta) \chi_h (\xi,\eta) \ d\mu(\xi,\eta)\\
    &=  \lVert f_1 - f_2 \rVert_\infty \int_{\mathbb{R}^2} \psi(x+\xi,y+\eta) \chi_h (\xi,\eta) \ d\xi\ d\eta\\
    &\le \lVert \mathcal{A} \rVert_\infty \lVert f_1 - f_2 \rVert_\infty.
\end{align*}
\end{proof}

\section*{APPENDIX B}

\blue{The pseudo-code in Alg. \ref{alg:geneo_rec} details the procedure for reconstructing a 2D signal from sparse sampling using the proposed GENEO framework.}

\begin{algorithm}
\footnotesize
\caption{\blue{GENEO-based 2D signal reconstruction}}
\label{alg:geneo_rec}
\begin{algorithmic}[1]
\Statex \textbf{Input:} \\
    Sparse sampling points $\{p_1, \ldots, p_r\} \subset \mathbb{R}^2$;\\ 
    Intensities $\{\hat\varphi(p_i)\}_{i=1}^r$; \\
    Reliabilities $\{\hat\psi(p_i)\}_{i=1}^r$;\\
    Library of $N$ patterns $\{P_i\}_{i=1}^N$ with $P_i = (h_i, \chi_{h_i})$; \\
    Weighting strategy $W$.
\Statex \textbf{Output:} \\
    Reconstructed signal $\varphi_{\mathrm{rec}} \colon \mathbb{R}^2 \to [0, 1]$
\Statex \textbf{Procedure:} 

\For{$i = 1$ to $N$} \textbf{in parallel}
    \For{each spatial coordinate $q \in \mathbb{R}^2$} \textbf{in parallel}
        \State $\mathcal{S}_{S,P_i}(q) \gets \int_{\mathbb{R}^2}\lvert \hat\varphi(q+\lambda) - h_i(\lambda) \rvert \cdot \hat\psi(q+\lambda) \chi_{h_i}(\lambda) \, d\mu(\lambda)$
        \State $\mathcal{A}_{S,P_i}(q) \gets \int_{\mathbb{R}^2} \hat\psi(q+\lambda)\chi_{h_i}(\lambda) \, d\mu(\lambda)$
        \State $\hat c_{S,P_i}(q) \gets \mathcal{A}_{S,P_i}(q) - \mathcal{S}_{S,P_i}(q)$
        \For{each point $p \in \mathbb{R}^2$} \textbf{in parallel}
            \State $\mathrm{sim}(p,P_i,q) \gets \hat c_{\hat S,P_i}(q)\chi_{h_i}(p-q)$
        \EndFor
    \EndFor
\EndFor


\For{each point $p \in \mathbb{R}^2$} \textbf{in parallel}

    \For{$i = 1$ to $N$} \textbf{in parallel}
        \State $w_{i, q}(p) \gets W(\{\mathrm{sim}(p,P_j,q)\}_{q \in \R^2, j = 1, \dots,N})$
    \EndFor
    
    \State $\varphi_{\mathrm{rec}}(p) \gets  \sum_{i=1}^{N} \int_{\R^2} w_{i,q}(p) h_i(p - q) \, dq\,$
\EndFor
    

\State \Return $\varphi_{\mathrm{rec}}$
\end{algorithmic}
\end{algorithm}

\bibliographystyle{IEEEtran}
\bibliography{IEEEabrv,StringDefinitions,BiblioIT}

\begin{IEEEbiography}[{\includegraphics[width=1in,height=1.25in,clip,keepaspectratio]{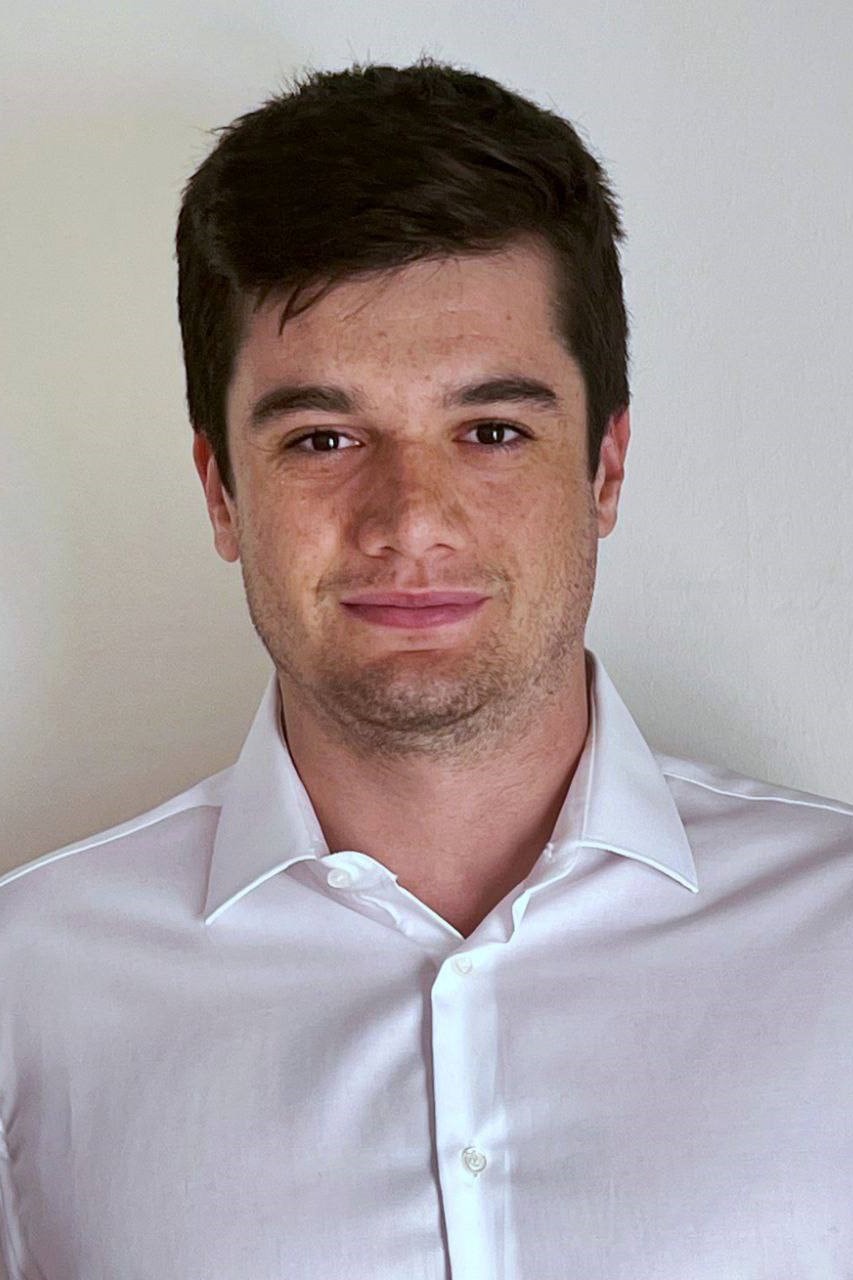}}]{LORENZO MARIO AMOROSA } (Member, IEEE) received the B.S. degree (\textit{summa cum laude}) in Computer Engineering and the M.S. degree (\textit{summa cum laude}) in Artificial Intelligence from the University of Bologna, Italy, in 2019 and 2021, respectively. In 2025, he received the Ph.D. degree in Electronics, Telecommunications and Information Technologies Engineering from the University of Bologna. Currently he is with the Department of Electronic, Information and Electrical Engineering ``Guglielmo Marconi'' as a Postdoctoral Research Fellow at University of Bologna. He is Research Associate at the National Laboratory of Wireless Communications (WiLab) of CNIT (the National, Inter-University Consortium for Telecommunications). His main research interests include Decentralized Artificial Intelligence, Cooperative Multi-Agent Systems, Machine Learning for Industrial IoT.
\end{IEEEbiography}

\begin{IEEEbiography}[{\includegraphics[width=1in,height=1.25in,clip,keepaspectratio]{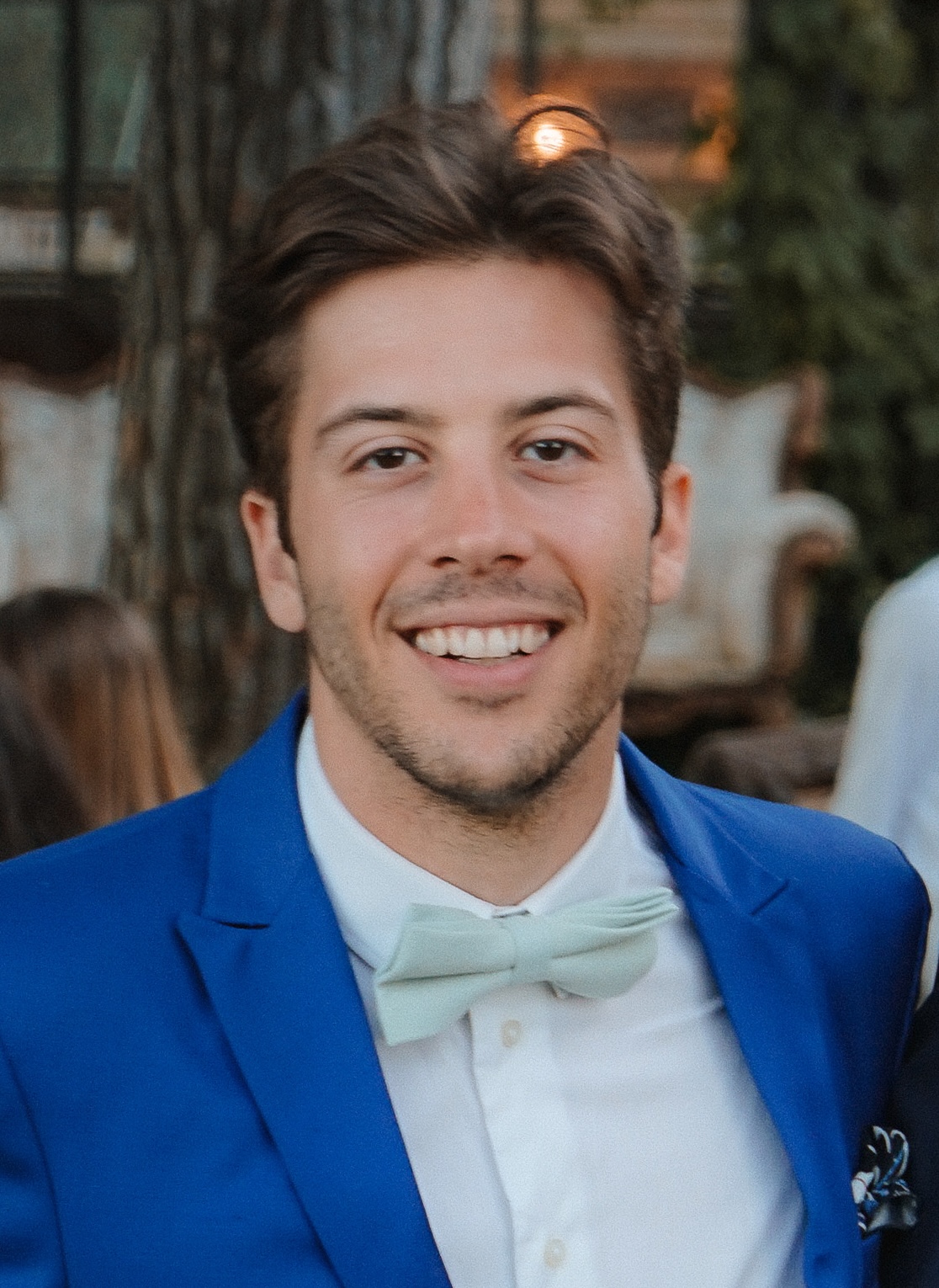}}]{FRANCESCO CONTI } is a PostDoc Researcher at Inria, Sophia-Antipolis, specializing in topological data analysis and geometric machine learning. He holds a PhD in Mathematics from the University of Pisa (2024), where he established theoretical connections between persistent homology and group equivariant non-expansive operators (GENEOs). His current postdoctoral research applies topology-infused generative models to galaxy mapping, while his broader work focuses on the topological and geometrical properties of GENEOs and TDA. Francesco has published in high-impact journals including Nature Scientific Reports and received the Under 35 Best Paper Award from AITA (2023). He has expertise in Python and specialized topological tools.
\end{IEEEbiography}

\begin{IEEEbiography}[{\includegraphics[width=1in,height=1.25in,clip,keepaspectratio]{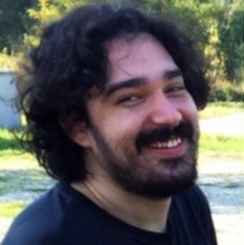}}]{NICOLA QUERCIOLI } is a researcher working at the intersection of mathematics and machine learning, with a focus on symmetry, topology, and data analysis. His work explores how structural properties of data can be formalized and integrated into AI models through advanced mathematical frameworks. He received his Ph.D. in Mathematics from the University of Bologna, where he studied the theory of Group Equivariant Non-Expansive Operators (GENEOs). He has collaborated with WiLab/CNIT on research projects involving topology-driven approaches to machine learning, with applications ranging from signal processing to high-performance computing environments.
\end{IEEEbiography}

\begin{IEEEbiography}[{\includegraphics[width=1in,height=1.25in,clip,keepaspectratio]{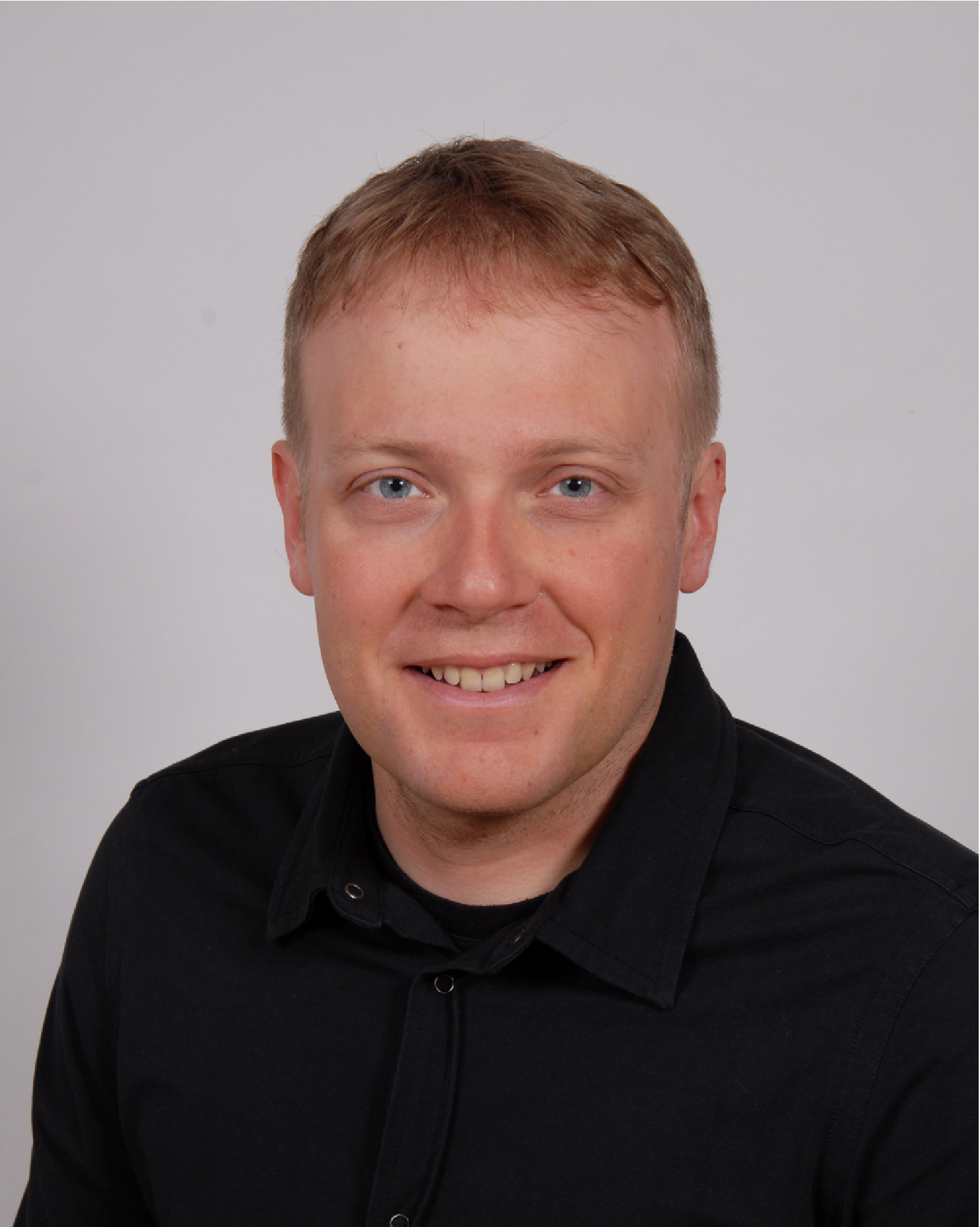}}]{FLAVIO ZABINI } (Member, IEEE) received the Laurea 
({\it summa cum laude}) 
in telecommunications  engineering and the Ph.D. in electronic engineering and computer science from the University of Bologna, Italy, in 2004 and  2010, respectively.
In 2004, he developed his master’s thesis at the University of California San Diego, La Jolla. 
In 2008, he was a Visiting Student with the DoCoMo Eurolabs of Munich, Germany. From 2013 to 2014, he was a Post-Doctoral Fellow with the German Aerospace Center, Cologne, Germany. He was with
the IEIIT-Bo of CNR. He is currently an Associate Professor at the University of Bologna. His current research interests include stochastic sampling, molecular communications, and joint sensing and communications. 
He served as an associate Editor for the IEEE Communication Letters and the KSII Transactions on Internet and Information Systems.
\end{IEEEbiography}

\begin{IEEEbiography}[{\includegraphics[width=1in,height=1.25in,clip,keepaspectratio]{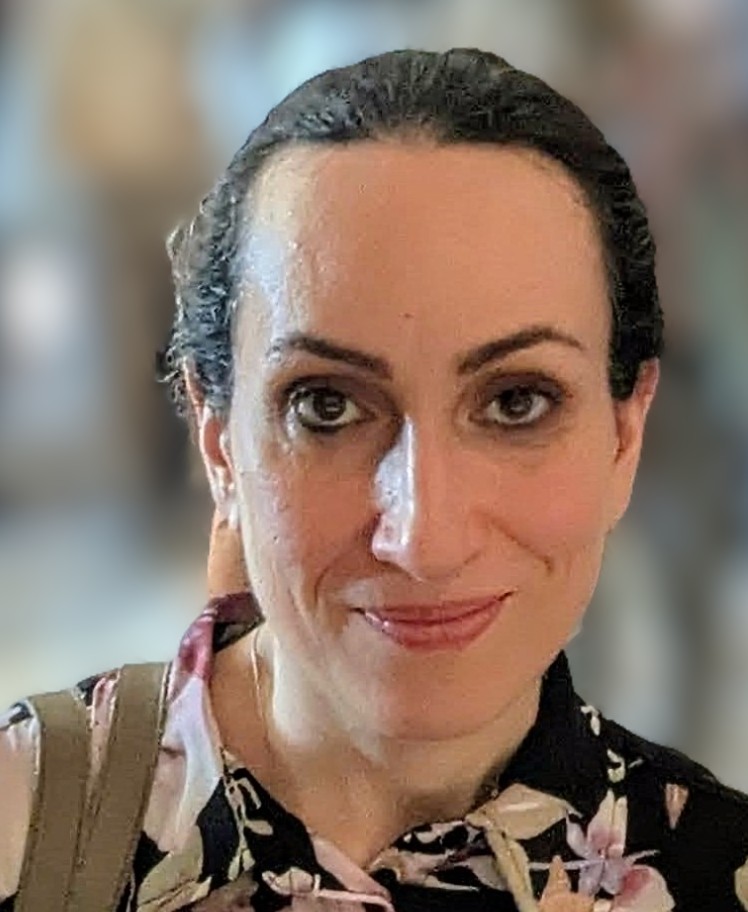}}]{TAYEBEH~LOTFI~MAHYARI } is a Senior Research Engineer at Huawei Technologies in Ottawa, Canada, where she focuses on applying artificial intelligence to wireless communication systems and standardization, with a particular emphasis on massive MIMO technologies for future 6G networks. She holds B.Sc. and M.Sc. degrees in Computer Engineering from Isfahan University of Technology and Sharif University of Technology, respectively, and earned a second M.Sc. and a Ph.D. in Computer Science from Simon Fraser University and Carleton University. Prior to joining Huawei, she taught part-time at Isfahan University of Technology and Islamic Azad University. Her broader research interests include machine learning, statistical modeling, and advanced wireless system design.
\end{IEEEbiography}

\begin{IEEEbiography}[{\includegraphics[width=1in,height=1.25in,clip,keepaspectratio]{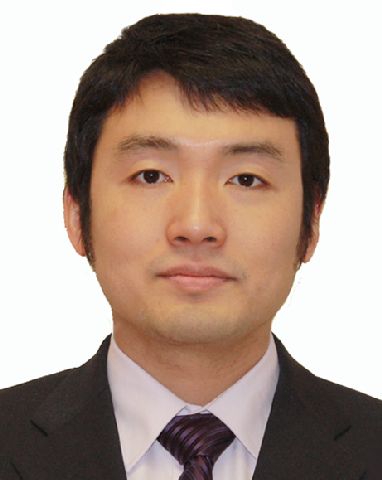}}]{YIQUN GE } is a Distinguished Research Engineer and 6G scientist at Huawei Technologies Co., Ltd. His research focuses on wireless communication systems and standardization, with particular interests in low-power chip design for wireless applications, channel coding, AI, and massive MIMO technologies for future 6G networks. He has over two decades of R\&D experience in wireless systems and has been actively contributing to cutting-edge innovations that bridge theory and industry standards. He received his B.Eng. degree from Shanghai Jiao Tong University in 2000 and his M.Sc. degree from ENST-Bretagne, France, in 2003.
\end{IEEEbiography}

\begin{IEEEbiography}[{\includegraphics[width=1in,height=1.25in,clip,keepaspectratio]{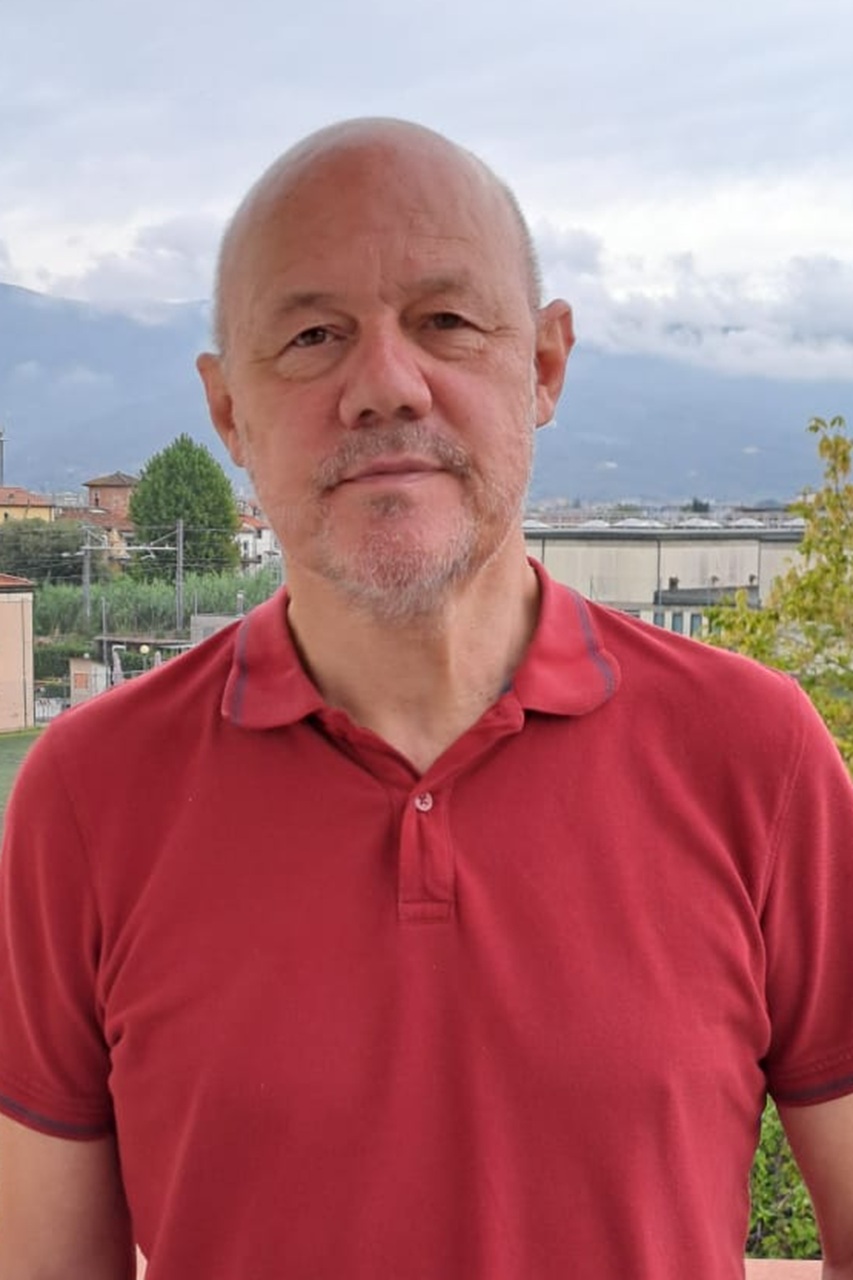}}]{PATRIZIO FROSINI } received his Ph.D. in Mathematics from the University of Florence in 1991. He taught for many years at the University of Bologna and is currently an Associate Professor in the Department of Computer Science at the University of Pisa. His research focuses on the topological and geometric properties of manifolds and topological spaces equipped with filtering functions taking values in $\mathbb{R}^n$, particularly in the presence of symmetry groups. In the early 1990s, he introduced and began studying the theoretical foundations that would later evolve into topological data analysis and persistent homology. Together with his collaborators, he has also introduced and begun developing the topological theory of group-equivariant non-expansive operators (GENEOs), as well as their applications in topological data analysis, machine learning, and artificial intelligence.
\end{IEEEbiography}

\end{document}